\documentclass[preprint,12pt,authoryear]{elsarticle}
\usepackage[colorlinks,linkcolor=blue]{hyperref}
\usepackage{natbib,lineno,hyperref,amsmath,graphicx,epstopdf,epsfig,multirow,booktabs,subfigure,amsfonts,amssymb}
\modulolinenumbers[5]
\usepackage{multicol}
\usepackage{tabularx}
\usepackage{makecell}
\usepackage{color}
\usepackage{diagbox}
\usepackage{algorithm}
\usepackage{algorithmic}
\usepackage[flushleft]{threeparttable}
\journal{ISPRS Journal of Photogrammetry and Remote Sensing}
\usepackage{geometry}
\biboptions{sort}
\begin{document}

\begin{frontmatter}

\title{Refined Equivalent Pinhole Model for Large-scale 3D Reconstruction from Spaceborne CCD Imagery}
%Calibarated Equivelant Pinhole Model for Accurate Spaceborne CCD imagery 3D Reconstruction

%Rational polynomial model Equivalent pinhole camera model for Large-scale Spaceborne CCD imagery 3D Reconstruction 

%Rethinking Spaceborne CCD imagery 3D Reconstruction Problem with Equivelant Pinhole Model

\author[label1]{Danyang Hong}
\ead{hdanyang2022@163.com}
\author[label1]{Anzhu Yu\corref{cor1}}
\ead{anzhu\_yu@126.com}
\author[label1]{Song Ji\corref{cor1}}
\ead{jisong_chxy@163.com}

\author[label1]{Xuefeng Cao}
\ead{CAO\_Xue\_Feng@163.com}
\author[label1]{Yujun Quan}
\ead{qyj5312020@126.com}
\author[label1]{Wenyue Guo}
\ead{ guowyer@163.com}
\author[label1]{Chunping Qiu}
\ead{chunping.qiu@outlook.com}

\cortext[cor1]{Corresponding author}

\address[label1]{PLA Strategic Support Force Information Engineering University,
            {Zhengzhou},
            {450001}, 
            {China}}

\begin{abstract}
In this study, we present a large-scale earth surface reconstruction pipeline for linear-array charge-coupled device (CCD) satellite imagery. While mainstream satellite image-based reconstruction approaches perform exceptionally well, the rational functional model (RFM) is subject to several limitations. For example, the RFM has no rigorous physical interpretation and differs significantly from the pinhole imaging model; hence, it cannot be directly applied to learning-based 3D reconstruction networks and to more novel reconstruction pipelines in computer vision. Hence, in this study, we introduce a method in which the RFM is equivalent to the pinhole camera model (PCM), meaning that the internal and external parameters of the pinhole camera are used instead of the rational polynomial coefficient parameters. We then derive an error formula for this equivalent pinhole model for the first time, demonstrating the influence of the image size on the accuracy of the reconstruction. In addition, we propose a polynomial image refinement model that minimizes equivalent errors via the least squares method. The experiments were conducted using four image datasets: WHU-TLC, DFC2019, ISPRS-ZY3, and GF7. The results demonstrated that the reconstruction accuracy was proportional to the image size. Our polynomial image refinement model significantly enhanced the accuracy and completeness of the reconstruction, and achieved more significant improvements for larger-scale images.

\end{abstract}

% \linenumbers
%% main text
\end{frontmatter}

%\begin{multicols}{2}
\section{Introduction}
Numerous sources are available for producing digital surface models (DSMs), including UAV, aerial, and satellite imagery, and point data can be obtained from laser scanner. Despite there exist the large amount of source utilized to obtain DSMs, extracting DSM results from satellite images is the most cost-effective option given the constant influx of terabyte-scale image data. Since satellites with various high-resolution (VHR) sensors, such as the WorldView-3 and Gaofen-7 series, were first launched, optical satellite images have achieved a resolution of better than 1 m. VHR satellite imagery has the potential to enhance the precision of DSM reconstruction and facilitate the 3D reconstruction of urban areas \cite{urban,urban_2017,city}.

Large-scale reconstruction of the earth’s surface from satellite images for the purpose of obtaining complete and accurate DSM results remains a challenge. Typically, a linear-array pushbroom is used to acquire satellite imagery, and a generalized rational functional model (RFM) \cite{RFM2, RFM1} is used for the imaging equation. Hence, pinhole camera images differ significantly from the imaging methods and imaging equations of linear array charge-coupled device (CCD) images. However, traditional methods for 3D reconstruction \cite{s2p, Marc2016, JM2020} of optical satellite images often rely on RFM. The 3D reconstruction of satellite optical images based on RFM involves several essential steps, such as bundle adjustment \cite{BA-Net, BAornot}, epipolar rectification \cite{pushre2014, NT2019, PL2022}, dense matching \cite{GA-Net, dense2020}, point cloud generation \cite{point2018}, and DSM fusion filtering \cite{Edge2017, MVSpipe2023}. Many studies have focused on improving these steps.

%For instance, integrating deep learning techniques has further developed dense matching. SuperGlue \cite{superglue2020} is a milestone in deep learning-based matching algorithms by combining attention mechanism and graph neural network to solve the optimization problem of feature assignment. Inspired by SuperGlue, LoFTR \cite{LoFTR2021}, a Transformers-based feature matcher without feature extraction, optimizes the traditional algorithm and achieves better results than SuperGlue in weak texture regions.

Additionally, numerous commercial software packages (e.g., ERDAS Imagine LPS  \cite{Erdas}, RSP \cite{RSP2016}, MicMac \cite{minmac2017}, Pixel Factory \cite{PixelFactory}, SURE \cite{SURE}, SOCET SET \cite{SOCETSET}, Agisoft Metashape \cite{agisoft}, and Catalyst professional \cite{catalyst}) and open source pipelines (e.g., S2P \cite{s2p}, CARS \cite{JM2020}, and ASP \cite{ASP2018}) have been developed to photogrammetrically process satellite images using RFM. ERDAS Imagine LPS (Leica Photogrammetry Suite) \cite{Erdas} is a well-established and robust photogrammetric processing package for aerial and orbital imagery. Nearly every orbital sensor is supported by rigorous information describing the camera model. For most other sensors, rational polynomial coefficient (RPC) processing is also supported. Pixel Factory \cite{PixelFactory} generates 3D mesh models from satellite images. By using multiple images for each model, it is able to process very large areas. One feature exclusive to Pixel Factory is that homogeneity and consistency are guaranteed throughout the world. SURE Software \cite{SURE} transforms imagery from classic aerial cameras, multi-head oblique systems, drone cameras, and most consumer-grade terrestrial cameras into 2.5D or 3D data, including point clouds, photorealistic textured meshes, and true orthophotos, via a streamlined fully automatic and integrated image processing technique. MinMac \cite{minmac2017} is free open-source photogrammetric software for 3D reconstruction, and it solved the multi-view fusion with a multi-directional dynamic programming technique for dense matching of VHR satellite images \citep{minmac2017,minmac2018}. S2P \cite{s2p} is a fully automated modular pipeline designed for affine reconstruction of line-array satellite images. Furthermore, the NASA Ames Research Center proposed the NASA Ames Stereo Pipeline \cite{ASP2018}, which is a suite of free and open-source automated geodesy and stereogrammetry tools for processing stereo images captured from satellites, in which rigorous physical sensor models are obtained by querying ephemerides and interpolating camera poses.

%(SoftCopy Exploitation Tool SET)
%RSP provided by \cite{RSP2016} is a RPC stereo processor that implements an entire pipeline of DSM and orthophoto generation based on RPC-modelled satellite imagery (level 1+).

The fully automatic and modular stereo pipeline S2P \cite{s2p} utilizes an affine model in the image space to optimize positioning based on PRC model. \cite{Auto2017} proposed a method that relies on local affine approximation \cite{block2003} and considers multi-date images, making it a multi-modal technique for reconstructing 3D models. \cite{JM2020} designed a new scalable, robust, high-performance stereo pipeline for satellite images called CARS. \cite{Pursue2022} proposed a hierarchical reconstruction framework that consists of an affine dense reconstruction stage and an affine-to-Euclidean upgrading stage based on multiple optical satellite images, which needs only four ground control points (GCPs). To attain a simple and speedy dense matching outcome, the 3D reconstruction pipelines detailed above execute stereo correction before the dense matching stage. However, it is difficult to accurately stereo-correct large-scale satellite stereo image pairs \cite{PL2022, NT2019, pushre2014, epipolar2} because spaceborne optical sensors always follow the linear-array pushbroom imaging process, during which there are differences between the epipolar geometries of different linear-array images. Most importantly, the imaging model for linear-array CCD images is complicated and relies on a generic RPC model fitted with polynomials, which introduces difficulties in constraining the reconstruction results with the original rigorous imaging model. \cite{kai2019} proposed the Adapted COLMAP to fit the RFM to the pinhole camera model (PCM), which fundamentally solves the problem of relying on RFM for linear-array CCD images. After resolving the disparity between the imaging models of linear array CCDs and pinhole cameras, ZhangKai et al. successfully accomplished 3D reconstruction of line-array CCD images utilizing the first-rate open-source computer vision software, COLMAP.

%This method resoundingly apply optical satellite images captured by CCD sensor to the state-of-the-art computer vision COLMAP pipeline \cite{schoenberger2016sfm, schoenberger2016mvs}. COLMAP is a multi-view 3D reconstruction technique that does not necessitate epipolar rectification.
To this end, in this study we propose the RFM is equivalent to PCM (REPM) pipeline, which is based on the Adapted CLOMAP pipeline \cite{kai2019} and relies heavily on the idea of RFM is equivalent to PCM to expand its applicability to linear-array CCD imagery. To enhance the reconstruction accuracy of the REPM pipeline, we construct a image refinement model that minimizes equivalent errors. We also incorporate an image-partitioning module and improve the DSM fusion module by enabling it to process large-scale images. In summary, our main contributions are as follows:

\begin{itemize}
  
  \item We introduce the RFM is equivalent to PCM model and mathematically derive the error formula of the equivalent pinhole model, which enables large-scale 3D reconstruction with linear array imagery using most exisiting 3D reconstruction pipelines. In addition, we further propose the image refinement model to improve the accuracy of 3D reconstruction.
  \item We present a multi-view 3D reconstruction pipeline for large-scale linear-array CCD imagery based on REPM, which encompasses the whole process from image intake to DSM product output.
  \item We have proven through formula derivation and experiments that the error of the equivalent pinhole model is directly proportional to the image size. Additionally, our pipeline shows excellent potential on four datasets. Remarkably, incorporating a polynomial image refinement model highlights a 15\% accuracy advancement on large surface format images.
\end{itemize}

\section{Methods for 3D reconstruction method of satellite images}

In this section, we focus on three satellite image-based 3D reconstruction methods: RFM-based, PCM-based, and REPM-based 3D reconstruction.

\subsection{RFM-based 3D reconstruction}

Satellite images are commonly captured by using linear-array CCD sensors in a pushbroom manner, the projection method of which is significantly different from that of the traditional pinhole camera model. Most satellite optical images reconstruct DSMs through the RPC models, which include 80 polynomial coefficients (78 polynomial coefficients to be solved) and 10 normalized constants (for a total of 90 parameters in the RPC file corresponding to each image), which are defined as 

\begin{large} 
\begin{equation}
 \left\{\begin{matrix}
 u=\mu _{u}+\sigma_{u}g(\frac{lat-\mu_{lat}}{\sigma _{lat}},\frac{lon-\mu_{lon}}{\sigma _{lon}},\frac{alt-\mu_{alt}}{\sigma _{alt}} ) \\
 v=\mu _{v}+\sigma_{v}h(\frac{lat-\mu_{lat}}{\sigma _{lat}},\frac{lon-\mu_{lon}}{\sigma _{lon}},\frac{alt-\mu_{alt}}{\sigma _{alt}} )
\end{matrix}\right. ,
  \label{eq:RPC}
\end{equation}
\end{large}

where $u, v$ represent the row and column pixel coordinates, respectively; $lat, lon, alt$ denote the latitudes, longitudes, and altitudes of the locations within the WGS-84 coordinate system, respectively; $\mu _{i}(i=u,v,lat,lon,alt)$ denotes five translation normalization parameters; $\sigma _{i}(i=u,v,lat,lon,alt)$ denotes five scaling normalization parameters; and the functions $g(\cdot )$ and $h(\cdot )$ are the cubic polynomial functions of the RPC model, each with 40 parameters. Because the numerators and denominators of the $g(\cdot )$ functions are simultaneously divided by a polynomial coefficient, the ratio remains unchanged. The same applies to the $h(\cdot )$ functions. Therefore, out of the 80 polynomial coefficients, there are two constant values of 1. 

The traditional RFM-based 3D reconstruction method \cite{traditionalRFM3d} involves initially acquiring the transformation between image pairs with homologous points using matching techniques. Then a 3D model of the reconstructed ground information is derived by accounting for various coordinate standardization parameters based on the RFM of stereo image pairs. However, the complexity of the RFM renders the entire reconstruction process extremely cumbersome. The S2P pipeline \cite{s2p} offers a solution for decoupling the 3D reconstruction process from the intricacies of satellite imaging. The S2P pipeline utilizes the relative pointing error correction between RPC models to replace the complicated nonlinear bundle adjustment. This process recovers the 3D structure of the paired satellite images using a simple RPC-based elevation iteration. \cite{JM2020} presented a new stable and efficient pipeline for multi-view stereo called CARS. In this pipeline, a colocalization function that employs the epipolar constraint, is fitted with a geometry based on a nonrigid iterative approximation. Then it jointly and recursively estimates two resampling grids mapped from the estimated epipolar geometry to the input images. \cite{Pursue2022} proposed a hierarchical reconstruction framework based on multiple optical satellite images called AE-Rec, which reconstructs the affine and Euclidean scene structures sequentially. In the first stage, an affine dense reconstruction approach is used to obtain the 3D affine structure from the input satellite images, and local small-sized tiles in the satellite images are approximately subject to an affine camera model. This affine approach is performed under an incremental reconstruction strategy and does not use any GCP. In the second stage, the obtained 3D affine structure is upgraded to a Euclidean structure by fitting a global transformation matrix with at least four GCPs.

\subsection{PCM-based 3D reconstruction}

Most 3D reconstructions based on the PCM adopt the idea that the overlapping image first performs feature point matching. The matched feature points are then used to obtain the ground point coordinates via space resection. However, because of the diversity of constraints, it is difficult to unify many 3D reconstruction methods. For example, 3D reconstruction based on local stereo matching \cite{SGM_census, patchmatch} uses the consistency of parallax in a small range for constraints, 3D reconstruction based on global stereo matching \cite{Belief_Propagation} explicitly uses smooth assumption constraints to solve the matching results of all pixels as a whole, and 3D reconstruction based on semiglobal stereo matching \cite{SGM} uses mutual information as the method for computing the similarity measure. Additional priori knowledge can be used as constraints to improve the accuracy of the 3D reconstruction.

%Traditional image-based 3D reconstruction methods include 3D reconstruction based on binocular stereo vision \cite{BSV1, BSV2} and based on multiple-view stereo vision \cite{MSV}. The principle of multiple-view stereo vision is the same as binocular stereo vision, which is the extension of binocular stereo vision, and it can reduce the blind area of the reconstructed site and solve the problem of mismatching in binocular stereo vision. The 3D reconstruction based on binocular stereo vision first finds the matching points between the stereo images through the epipolar rectification and then recovers the 3D information of the scene according to the principle of triangulation. The well-known stereo matching algorithms include Global Stereo Matching (GSM) \cite{Belief_Propagation}, as well as Local Stereo Matching (LSM) \cite{SGM_census, patchmatch}, Semi-global Stereo Matching (SGM) \cite{SGM}, and a series of optimized algorithms of SGM. For instance, the Patchmatch \cite{patchmatch} algorithm is operated in the open-source 3D reconstruction software COLMAP for dense matching throughout the reconstruction procedure.

Furthermore, the combination of structure from motion (SfM) \cite{schoenberger2016sfm} and multi-view stereo (MVS) \cite{schoenberger2016mvs} is considered a favorable vision-based reconstruction framework for 3D scene restoration. SfM estimates the camera position, orientation and reconstructs the sparse point clouds. Subsequently, MVS generates dense point clouds based on the SfM results to reconstruct the 3D scene. However, the linear-array CCD images do not rigorous constraints of camera position and orientation like pinhole camera images because of the significant differences between the imaging models of pinhole cameras and those of satellite linear-array sensors. Thus, this framework primarily focuses on the 3D reconstruction of images captured using pinhole cameras.

\subsection{REPM-based 3D reconstruction}

Statistical analyses in the literature \cite{review_urban} show that traditional 3D reconstruction methods are still significant in the field of  3D reconstruction of satellite images. However, these satellite-based 3D reconstruction methods rely on RFM. Because of the RPC function’s complexity, incompatibility with many novel 3D reconstruction pipelines, and lack of a rigorous epipolar correction model, a recent innovative approach suggested fitting the RFM to the PCM. The Adapted COLMAP \cite{kai2019} method approximates a weak perspective projection model using a well-established RFM and enhances the original COLMAP \cite{schoenberger2016sfm, schoenberger2016mvs} visual reconstruction pipeline for satellite images with a depth reparameterization technique, thereby improving the accuracy of the depth values. However, the Adapted COLMAP method is only applicable to small-scale images. In this pipeline, the equivalent error is larger for large-scale images, the reconstruction effect is poorer, and NOT AVAILABLE (NA) even occurs. To reduce the equivalent error, we corrected image using polynomial function, which is called the image refinement model. In addition, to address the NA phenomenon in large-scale images, we added an image partition module and improved the DSM generation module.

%For large-scale satellite images, stereo matching is usually performed by calculating the matching cost of stereo pairs. Previous work has applied stereo matching algorithms to the 3D reconstruction of satellite images through affine transformations locally approximated as pinhole camera models. For example, S2P pipeline \cite{s2p} uses the stereo matching algorithm of More Global Matching (MGM) \cite{MGM}, which optimizes SGM to solve the streaking effect problem to generate 3D information.

\section{REPM-based 3D reconstruction pipeline}
This section presents the REPM pipeline framework, the algorithmic and error formulas of the equivalent pinhole model, and the principle of the image refinement model. The design of the REPM pipeline includes the RPC model’s approximation of the pinhole camera model, image refinement model and relies heavily on features included in Adapted COLMAP \cite{kai2019}, including the SfM \cite{schoenberger2016sfm} and MVS \cite{schoenberger2016mvs} frameworks.

% \begin{figure}[h]
%   \begin{center}
%     \includegraphics[scale = 0.60 {figs/fig1.png}

%     \caption{Reconstruction overview. The equivalent camera model is created using EPM firstly, and subsequently proceeding with sparse and dense reconstruction to acquire the point cloud and digital surface model.}\label{fig:pipeline}
%   \end{center}
% \end{figure}

\subsection{REPM pipeline}
\begin{figure*}[t]
  \begin{center}
    \includegraphics[scale = 0.7]{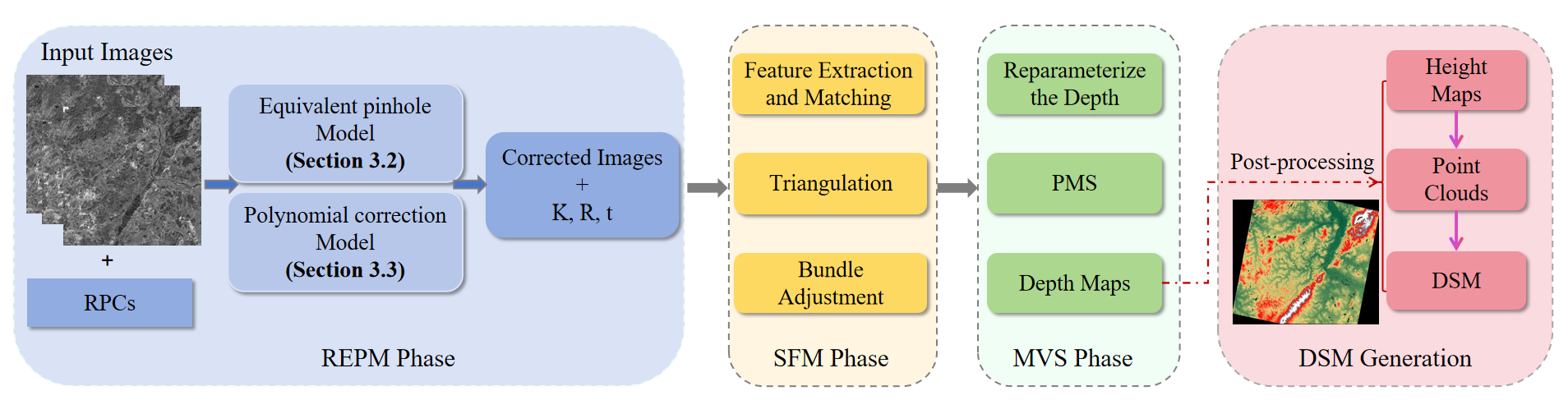}
    
    \caption{Overview of the reconstruction process of the REPM pipeline. The procedure involves four phases: REPM, SfM, MVS, and DSM generation. The inputs consist of several overlapping images and their corresponding RPC models. The outputs include height maps, georeferenced 3D point clouds, and digital surface models.}\label{fig:pipeline}
  \end{center}
\end{figure*}

An overview of the reconstruction process is presented in Fig.\ref{fig:pipeline}. Under the assumption of weak perspective projection, the REPM pipeline performs multi-view image processing by using $n$ source images and their RPC parameters to compute the internal matrix, rotation matrix, and translation vectors $\{\mathbf{K}_i,\mathbf{R}_i, \mathbf{t}_i \}_{i=0}^N$ corresponding to the input views through the equivalent pinhole model. The corrected image is then obtained using a image refinement model that minimizes the error of the equivalent pinhole model. The SfM framework performs sparse reconstruction with the given $\{\mathbf{K}_i,\mathbf{R}_i, \mathbf{t}_i \}_{i=0}^N$ and corrected images. The sparse point cloud and optimized camera poses jointly participate in the MVS phase to estimate the depth map, and then fuse to generate the dense point cloud and DSM.

\textbf{REPM phase.} In the REPM phase, we introduce the RFM is equivalent to PCM algorithm and implement a image refinement model to rectify the image by minimizing the equivalent error. We first perform image partitioning and image enhancement operations, in which the accuracy of the equivalent pinhole model is influenced by the size of the satellite image and the equivalent error is decreased by clipping the image. Image enhancement is necessary because the long-tail distribution of the brightness values of the satellite image is detrimental to image feature matching. We set a brightness threshold and perform the image enhancement technique when the brightness values exceed this threshold. After image partitioning and enhancement, the images and RPC parameters are fed into the equivalent pinhole model described in Section \ref{sec:3.2} and the image refinement model described in Section \ref{sec:3.3}.

%We integrated the image partition module to address the computational challenges posed by large-scale images. Regarding the necessity of image partition and image enhancement. as satellite images are generally in the tens of thousands of pixels, which exceeds the computer's processing capabilities, the preliminary step in the reconstruction involves cropping and resizing it to a suitable size for the computer to run.

\textbf{SfM and MVS phases.} For the SfM and MVS phases, we employ the same approach as that used in Adapted COLMAP \cite{kai2019}. In the field of computer vision, the SfM and MVS frameworks are well-developed 3D reconstruction frameworks, which are used in the excellent reconstruction pipeline COLMAP. During the SfM stage, feature extraction and matching are initially performed to match the homonymous points of the stereo image. Next, triangulation is conducted to calculate the 3D point coordinates from the homonymous points. Finally, global optimization is executed using beam method leveling to generate an optimal camera parameter model. SfM produces the internal parameters, position, sparse point cloud, and co-visual relationship of the point cloud. Using this information, MVS executes pixel-by-pixel dense matching to create a depth map that matches the corresponding source images. In addition, to address the issue of inaccuracy caused by large depth values, the reparameterization approach is adopted and then the depth map is estimated using the classical PatchMatch Stereo (PMS) \cite{patchmatch} algorithm.

%The PMS method is a local stereo-matching technique that uses a window-based approach for calculating the similarity of entire window blocks instead of individual pixels. This approach results in improved search efficiency and increased robustness.

\textbf{DSMs aggregation.} We restructured the DSM generation model to align it with the processing of the image partition module.   Because the image is partitioned into multiple image tiles, the reconstructed DSMs from the same viewpoint are first mosaicked, and then DSMs from different perspectives are aggregated.   Because the image-matching process does not guarantee 100\% accuracy, the reconstructed DSM results contain incorrect height values. Therefore, outlier removal is incorporated into the process of aggregating DSMs, including two methods: 1) the median absolute deviation (MAD), which is used to filter out DSM outliers, and 2) radius point cloud filtering, which removes noise. However, a disadvantage of radius point cloud filtering is the absence of a standard parameter, which necessitates combining the reconstruction results to determine the parameter.

\subsection{Equivalent pinhole model} \label{sec:3.2}
\subsubsection{Introduction to the equivalent pinhole model} 

The equivalent pinhole model is based on the principle of weak perspective projection. First, let the range of the ground altitude variation be denoted by $Z_{range}$ and the distance from the satellite sensor to the ground point be denoted by the depth $D$. For remote sensing images, the satellite sensor is far from the ground point (i.e., $D \gg Z_{range}$). In this case, the average scene depth can be used in the projection calculation instead of the depth; this substitution is the theoretical basis for approximating the perspective camera as a weak perspective camera. Furthermore, \cite{kai2019} proved that a linear pushbroom camera can be reduced to a weak perspective camera under the same conditions. Therefore, a linear pushbroom camera can be approximated as a perspective camera, which we refer to as the equivalent pinhole model. The procedure of the equivalent pinhole model algorithm is presented in \textbf{Algorithm 1}. Given the RPC parameters of the images, the projection matrix of the perspective camera model (i.e., the internal matrix $K$ and the external matrices $R$ and $t$) can be estimated using the equivalent pinhole model algorithm.

\begin{table}[t]
	\centering
	\begin{tabularx}{0.5\textwidth}{X}
		\toprule
        \textbf{Algorithm 1:} Equivalent Pinhole Model Algorithm.
		\\
        \midrule
        \textbf{Input:} RPC parameters
        \\
        \textbf{Output:} K, R, t parameters\\
		\hspace{2em}1: Calculate the latitude, longitude, and altitude ranges based on the RPC parameters, and construct a set of hierarchical virtual 3D target grid points based on the range.
		\\  \\
        \hspace{7em}$X\_min = X_{off} - X_{scale}$ \\   \hspace{7em}$X\_max = X_{off} + X_{scale}$  \\
        \hspace{5em}$(X=Latitude,Longitude,Altitude)$
        \\
         \hspace{9em}$(Lat, Lon, Alt)_{n}$\\
        \\
        \hspace{2em}2: Project the 3D points onto the image utilizing the RPC parameter, and filter out correspondences with pixel coordinates that lie outside the image boundary.
        \\  \\
        \hspace{6em}$(Row, Col) = F(Lat, Lon, Alt);$\\
        $mask = (Row < 0, Row > height, Col < 0, Col > width)$ \\
         \\
        \hspace{2em}3: Convert the 3D points from the WGS-84 coordinate system to the ENU coordinate system.\\ \\
        \hspace{6em}$ (Lat, Lon, Alt)_{n} \longrightarrow (e, n, u)_{n} $\\
         \\
        \hspace{2em}4: Calculate the projection matrix P from 3D point coordinates to 2D pixel coordinates is solved using singular value decomposition (SVD).\\  \\
        \hspace{6em}$ SVD(e, n, u, Row, Col) \longrightarrow P $\\
         \\
        \hspace{2em}5: The factorization projection matrix P forms K, R, t for the camera's intrinsic and extrinsic parameters. \\  \\
        \hspace{9em}$ P \longrightarrow K, R, t $\\
         \\
		\bottomrule
	\end{tabularx}%%
	%\caption{A table with line breaks}
\end{table}%

\subsubsection{Error formula of the equivalent pinhole model}\label{sec:EPM}

The equivalent pinhole model algorithm equates the RPC parameters of the RFM with the internal and external parameters of the PCM via

\begin{equation}
\begin{split}
  &\binom{u}{v}=F(lat,lon,alt) 
    \Rightarrow  \\ &
    Z_{cam}\begin{pmatrix}
     u\\ v\\ 1
    \end{pmatrix} =P_{3\times4}\begin{pmatrix}
     X\\ Y\\ Z\\ 1
    \end{pmatrix} =K_{3\times 3} \left [ R | t \right ]_{3\times 4}\begin{pmatrix}
     X\\ Y\\ Z\\ 1
\end{pmatrix}
  \label{eq:cameraparam}
\end{split},
\end{equation}

where $F(\cdot )$ denotes the cubic polynomial function of the RPC model; $u, v$ denote the pixel coordinates' columns and rows, respectively; $lat, lon, alt$ indicate the latitude, longitude, and altitude, respectively, in the WGS-84 coordinate system; $\begin{pmatrix} u&  v&1 \end{pmatrix}^{T}$ is the homogeneous coordinate in the pixel coordinate system; $\begin{pmatrix} X&  Y& Z &1 \end{pmatrix}^{T}$ is the homogeneous coordinate in the east-north-up (ENU) coordinate system; $Z_{cam}$ denotes the $Z$-coordinate of the object point in the camera coordinate system (which can also be interpreted as the depth value of the object point); $P_{3\times4}$ is the projection matrix that converts the object point coordinates into pixel coordinates; and $K, R, t$ are obtained by factorizing the projection matrix $P$.

Both the RFM and PCM are utilized to model the relationship between 2D pixel coordinates and 3D object point coordinates. The RFM uses a cubic polynomial function, which is computationally complex. In contrast, the PCM employs homogeneous coordinates and matrix multiplication, which significantly simplify operations such as rotation and translation in 3D space. The RFM and PCM use different world coordinate systems, and the equivalent pinhole model algorithm utilizes the ENU coordinate system because of its relative compatibility with the conventional PCM (compared to the WGS-84 coordinate system).

The weak perspective projection principle is utilized to approximate a linear pushbroom camera as a perspective camera. Accordingly, a weak perspective projection formula is used to derive the equivalent error formula. The weak perspective projection formula is

%The weak perspective projection principle is utilized to approximately equate the linear pushbroom camera to a perspective camera. Subsequently, the weak perspective projection formula is used to derive the equivalent error formula. Eq.(\ref{eq:weak}) presents the weak perspective projection formula, which projects image coordinates $x, y$ to object point coordinates within the image space coordinates system. $f_{x}, f_{y}$ represent the mapping of the sensor focal length on the $x$ and $y$ axes, $(X_{cam},Y_{cam},Z_{cam})$ denotes the object point coordinates in the image space coordinates system, and $\bar{Z}$ signifies the average value of all object points within the image area. %The weak perspective projection formula is used instead to obtain Eq.(\ref{eq:weak}).

\begin{normalsize}
\begin{equation}
  \left\{\begin{matrix}
     x=\frac{f_{x}{X_{cam}}}{ \bar{Z} }\\
     \\
     y=\frac{f_{y}{Y_{cam}}}{ \bar{Z} }
  \end{matrix}\right.,
  \label{eq:weak}
\end{equation}
\end{normalsize} 

where $x, y$ are the image coordinates that are projected onto the object point coordinates within the image-space coordinate system; $f_{x}, f_{y}$ represent the mapping of the sensor focal length on the $x$ and $y$ axes, respectively; $(X_{cam},$ $Y_{cam}, Z_{cam})$ denote the object point coordinates in the image space coordinate system; and $\bar{Z}$ denotes the average value of all object points within the image area.

%We take the perspective projection formula for the x-axis as an example for further derivation. For digital images , the multiple order derivatives of the perspective projection formula can be obtained, so the formula satisfies the conditions for the Taylor expansion. Since we approximate the projection model as a weak perspective projection and $Z_{cam}$ is approximated as $\bar{Z}$, we obtain the Taylor expansion at $\bar{Z}$, see Eq.(\ref{eq:Taile}). Next, by comparison with Eq.(\ref{eq:weak}), we can obtain the error formula approximated as a weak perspective projection, see Eq.(\ref{eq:Error}), and the higher-order term of the Taylor formula is not considered.

We consider the perspective projection formula for the x-axis as an example for further derivation. For digital images, the multiple-order derivatives of the perspective projection formula can be obtained; thus, the formula satisfies the conditions for Taylor expansion. Because we approximate the projection model as a weak perspective projection and $Z_{cam}$ is approximated as $\bar{Z}$, we obtain the Taylor expansion at $\bar{Z}$:

\begin{equation}
\begin{split}
  x&=\frac{f_{x}X_{cam} }{Z_{cam}} \\& =  \frac{f_{x}X_{cam} }{\bar{Z}}+\frac{{x}'(\bar{Z}) }{1!}(Z_{cam}-\bar{Z})+o(Z_{cam}-\bar{Z})^{2} 
  \label{eq:Taile}
\end{split} ,
\end{equation}

where $o(\cdot)$ is the Peano remainder term of the Taylor expansion, representing the higher order infinitesimals of (a-b).

Next, by comparison with Eq.(\ref{eq:weak}), we obtain an error formula approximated as a weak perspective projection:

\begin{equation}
  E=\frac{{x}'(\bar{Z} ) }{1!}(Z_{cam}-\bar{Z}) = -f_{x}\frac{X_{cam}}{\bar{Z} ^{2} }(Z_{cam}-\bar{Z} ).
  \label{eq:Error}
\end{equation}

The higher-order term of the Taylor formula was not considered.

%It is notable that when the inequality of arithmetic and geometric means is introduced, the maximum value of the error term E can be estimated as:

%\begin{equation}
%  \left | E \right | =\left | -f_{x}\frac{X_{cam}}{\bar{Z} ^{2} }(Z_{cam}-\bar{Z} ) \right | \le \frac{1}{2}\left ( f_{x}^{2} \frac{X_{cam}^{2}}{\bar{Z} ^{4} }+(Z_{cam}-\bar{Z})^{ 2} \right ) 
%  \label{eq:Error2}
%\end{equation}

\begin{figure}[h]
  \begin{center}
    \includegraphics[scale = 0.45]{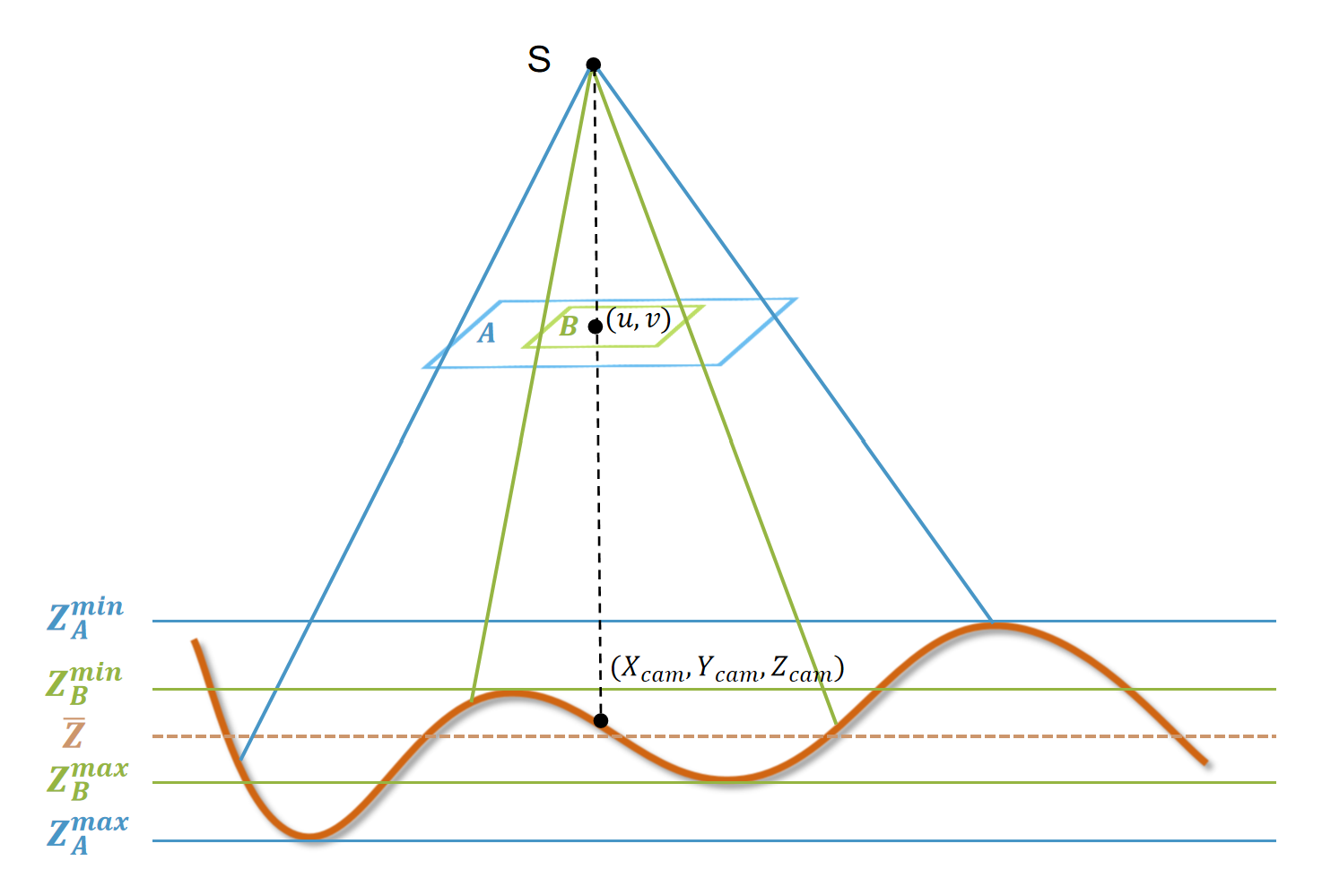}
    
    \caption{Diagram of the error formula of equivalent model. The brown region represents the earth's surface, and the blue and green regions represent the projection of images A and B for two different image sizes, respectively, for the same surface on the earth.}\label{fig:EPM}
  \end{center}
\end{figure}

The factors that influence the equivalent error can be derived from Eq.(\ref{eq:Error}). As $f_{x}\frac{X_{cam}}{\bar{Z}}\approx x $, we can derive $\left | E \right |\approx \left | -x(Z_{cam}-\bar{Z})/ \bar{Z}\right |$. Ideally, when reconstructing the 3D information of a particular area, a schematic such as that shown in Fig.\ref{fig:EPM} should be drawn to assist with the illustration. As $Z_{cam} \gg  Z_{cam}-Z_{min}$ and $\bar{Z} = (Z_{max}+Z_{min})/2$, we propose that $\bar{Z}$ is similar in image regions $A$ and $B$. Consequently, we demonstrate that in identical spatial areas, the equivalent error is determined by the image size and the disparity in ground elevation $Z_{cam}-\bar{Z}$. For images $A$ and $B$, $E_{A} \le \left | -x_{A}(Z_{A}^{max} -\bar{Z}) / \bar{Z}  \right |$, $E_{B} \le \left | -x_{B}(Z_{B}^{max} -\bar{Z})/ \bar{Z}  \right |$. Because $u_{A} > u_{B}$ and $Z_{A}^{max} > Z_{B}^{max}$, we conclude that $E_{A} > E_{B}$. Based on Fig.\ref{fig:EPM}, it can be inferred that, in general, the size of an image can indirectly determine the height difference between the ground and the corresponding object, which is proportional to the size of the image. Therefore, to control the equivalent error in the experiments, suitable image sizes were routinely obtained via cropping.

\subsection{Image refinement model} \label{sec:3.3}

\begin{figure}[h]
  \begin{center}
    \includegraphics[scale = 0.55]{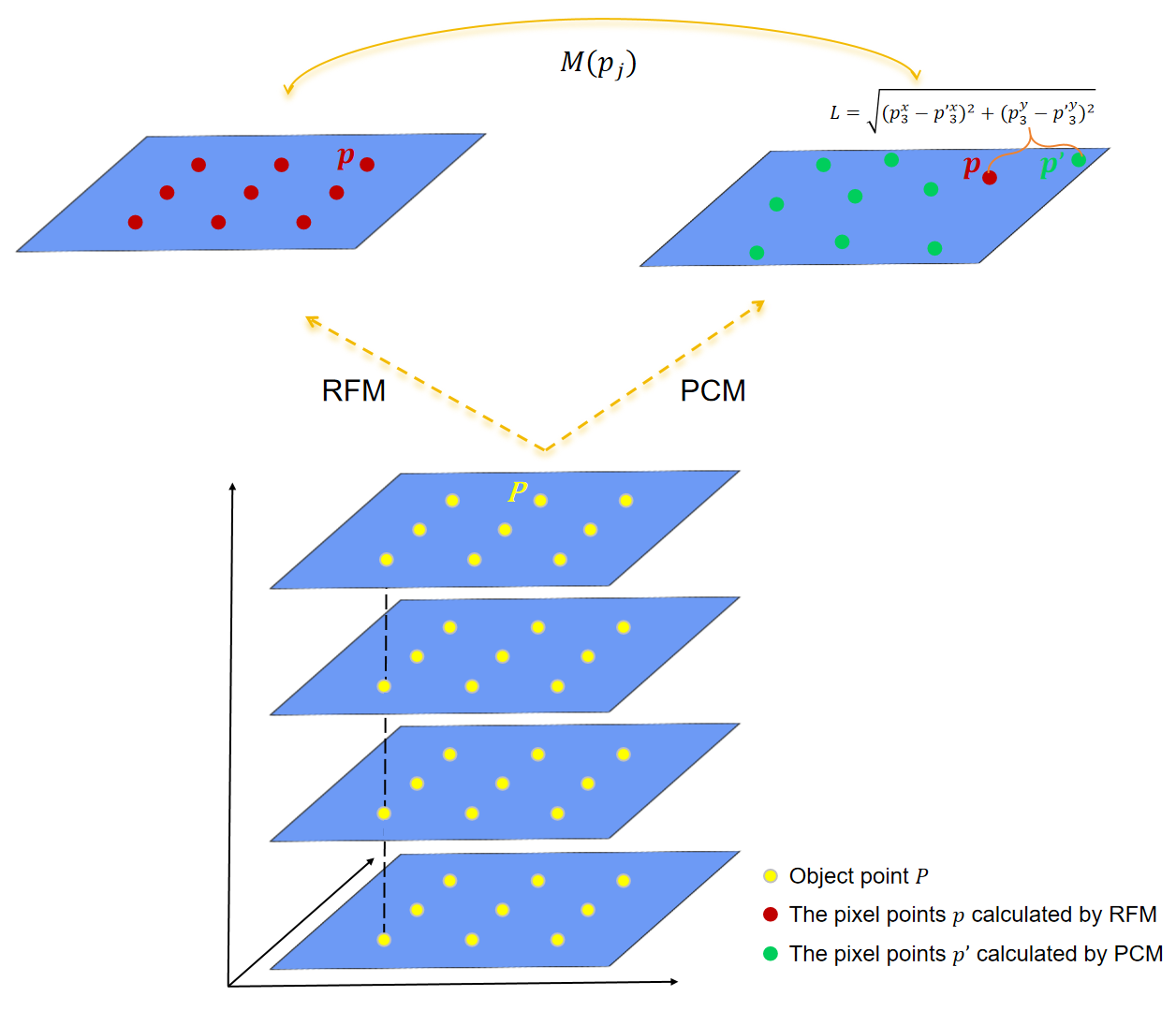}
  \end{center}
  \caption{Image refinement model. An object point $P$ is taken as an example, and $p$ and ${p}'$ denote the positions before and after correction, respectively.}\label{fig:correction}
\end{figure}

In Section \ref{sec:EPM}, we explained that the error in the equivalent pinhole model arises from using RFM and PCM to calculate the positional deviation from the object point to the pixel point. To reduce this bias, we minimize the equivalent error by introducing a polynomial image refinement model, as shown in Fig.\ref{fig:correction}. The pixel points calculated via the RFM and PCM form $n$ sets of corresponding points, which are substituted into the polynomial correction function:

%In Section \ref{sec:EPM}, we discovered that the error in the equivalent model arises from the utilization of RFM and PCM for calculating the positional deviation from the object point to the pixel point. To reduce the bias, we minimise the equivalent error by introducing a correction model, as shown in Fig.\ref{fig:correction}, we use a polynomial correction model. The pixel points calculated by RFM and PCM form n sets of corresponding points, which are substituted into the polynomial correction function Eq.(\ref{eq:Polynomials}), and then the parameters of the polynomial correction model are computed by using the least squares method, i.e. Eq.(\ref{eq:errorequation}). Finally, the polynomial correction model is used to resample the original image to obtain the corrected image.

\begin{equation}
  \begin{cases} x'=m_0+m_1x+m_2y+m_3xy+m_4x^2+m_5y^2\\
  y'=m_6+m_7x+m_8y+m_9xy+m_{10}x^2+m_{11}y^2
  \end{cases},
  \label{eq:Polynomials}
\end{equation}
where $m_{i} (i=0,1 \cdots 11)$ are the coefficients of the polynomial function.

The parameters of the polynomial image refinement model are computed using the least squares method:

\begin{equation}
  \underset{M(p)} {min} \sum_{j}\left \| M(p_{j})-{p}'_{j} \right \|_{2}^{2}.
  \label{eq:errorequation}
\end{equation}

Finally, the polynomial image refinement model is used to resample the original image to obtain the corrected image.

Initially, our aim was to correct the image by minimizing the equivalent error through the homography transformation. However, the results of the homography correction model were unsatisfactory. Instead, we selected a more complex second-order polynomial transformation to correct the images and reduce the equivalent pinhole model error. Through experiments, we found that the polynomial correction was more effective (see Section \ref{sec:correctionerror} for a description of the experiments). Therefore, we added the polynomial image refinement model before starting the downstream reconstruction task. The algorithm used for correcting the model is presented in \textbf{Algorithm 2}.

%Initially, our aim was to correct the image by minimising the equivalent error through the homography transformation. However, the results of the homography correction model were not satisfactory. We then chose a more complex second-order polynomial transformation to correct the images and reduce the equivalent model error. Through experiments we found that the polynomial correction was more effective (see Section \ref{sec:correctionerror} for experiments), so we added the polynomial correction model before starting the downstream reconstruction task. The algorithmic process of correcting the model is shown in \textbf{Algorithm 2}.

\begin{table}[!htbp]
	\centering
	\begin{tabularx}{0.5\textwidth}{X}
		\toprule
        \textbf{Algorithm 2:} Image refinement Model Algorithm.
		\\
        \midrule
        \textbf{Input:} Original images; RPC parameters; and $K, R, t$ parameters 
        \\
        \textbf{Output:} Corrected images and the polynomial image refinement function $M(\cdot)$ \\
		\hspace{2em}1: From Algorithm 1, hierarchical virtual 3D target grid points are projected through the RFM imaging model to produce $n$ sets of 3D-2D ($P$, $p$) correspondences.
		\\   \\
        \hspace{7em}$P(X, Y, Z)_{n} \longrightarrow  p(x, y)_{n}$  
        \\
        \\
		\hspace{2em}2: The hierarchical virtual 3D target grid points are projected to obtain pixels $p'$ based on the equivalent PCM imaging model, where $p$ and $p'$ constitute the $n$ sets of corresponding points.
        \\  \\
         \hspace{7em}$P(X, Y, Z)_{n} \longrightarrow  {p}'({x}', {y}')_{n}$\\
        \hspace{10em}$(x, y, {x}', {y}')_{n}$\\
        \\
        \hspace{2em}3: We back-calculate the polynomial parameters to correct the image. The relationship between $p$ and ${p}'$ is used to calculate the polynomial iamge refinement parameters via the least squares method.
        \\ \\
        \hspace{8em}$\underset{M(p')} {min} \sum_{j}\left \| M(p'_{j})-{p}_{j} \right \|_{2}^{2}$\\ \\
        
        \hspace{2em}4: After polynomial correction, the corrected image is generated. The polynomial correction function $M(\cdot)$ is used to compute the pixel positions of the original image, and the bilinear interpolation step in the resampling process provides the digital number (DN) values of the corrected image.
                \\  \\
        \hspace{8em}$DN({p_{j}}') = DN(M(p'_{j}))$\\ \\
		\bottomrule
	\end{tabularx}%
	%\label{tab:addlabel}%
	%\caption{A table with line breaks}
\end{table}%

%\begin{equation}
%  \begin{pmatrix} {x}' \\ {y}'\\ 1 \end{pmatrix}=H\begin{pmatrix} x \\ y\\ 1 \end{pmatrix}=\begin{bmatrix}
%  h_{0} & h_{1} & h_{2}\\
%  h_{3} & h_{4} & h_{5}\\
%  h_{6} & h_{7} & h_{8} \end{bmatrix}\begin{pmatrix} x \\ y\\ 1 \end{pmatrix}
%  \label{eq:homologous}
%\end{equation}

%\begin{equation}
%  \begin{bmatrix}
%  -x&  -y&  -1&  0&  0&  0&  x{x}' &  y{x}'& {x}'\\
%  0&  0&  0&  -x&  -y&  -1&  x{y}'&  y{y}'&{y}'
%  \end{bmatrix}h=0
%  \label{eq:errorequation}
%\end{equation}

\subsection{Reconstruction of the DSM}

A major factor that distinguishes satellite stereo pipelines from typical vision pipelines is the camera model (RPC vs. pinhole). Once we have obtained the corrected images and the internal and external parameter matrices of the equivalent pinhole camera, we can feed them into the SfM and MVS frameworks in the Adapted COLMAP pipeline \cite{kai2019} to reconstruct the 3D information.  The goal of SfM is to recover accurate camera parameters for use in subsequent MVS steps. In addition, the authors identified and resolved key issues in the MVS framework that prevented the direct application of standard MVS pipelines tailored for ground-level images to the satellite domain.

\section{Experiments}
In this section, we describe the datasets we used and the experiments we conducted on them. We also present an analysis of the experimental results.

\subsection{Experimental setup}

\subsubsection{Datasets}
We assessed our pipeline using three publicly accessible image datasets: the WHU-TLC test set, the DFC2019 dataset, and the ISPRS-ZY3 image data. We also present the reconstruction outcomes of applying the REPM pipeline to GF7 image data.

\begin{itemize}

  \item \textbf{WHU-TLC test set.} The WHU-TLC test set, including three-view images and RPC parameters that have been refined in advance to achieve sub-pixel reprojection accuracy, is provided by \cite{Gao_2021_ICCV}. The ground-truth DSMs were prepared using both high-accuracy LiDAR observations and GCP-supported photogrammetric software. The DSM is stored as a regular grid with 5-m resolution using the WGS-84 geodetic coordinate system and the UTM projection coordinate system.

  \item \textbf{DFC2019 dataset.} 
  The four sites of the DFC2019 dataset \cite{DFC} from the 2019 IEEE Geoscience and Remote Sensing Society (GRSS) Data Fusion Competition were used in this study. The dataset was obtained using the WorldView3 satellites. The dataset is comprised of satellite images featuring multiple views captured on multiple dates between 2014 and 2016. Each RGB image measures $2048 \times  2048$ pixels, and the ground truth DSM is $512 \times  512$ pixels.
  
  \item \textbf{ISPRS-ZY3 data.} 
  China’s first high-resolution stereo mapping civilian satellite, ZY3, has provided reliable high-resolution stereo image data. The experimental data from the International Society for Photogrammetry and Remote Sensing (ISPRS-ZY3) \cite{ISPRS_zy} covers Sainte-Maxime, France, including three line-array stereo panchromatic images of 2.1 m for nadir and 2.5 m for forward and backward. The ISPRS-ZY3 data consists of 12 ground control points for checking the absolute accuracy of the DSM.
  
  \item \textbf{GF7 data.} 
  The Gaofen-7 (GF7) satellite has a dual-line array stereo camera that delivers high-precision remote sensing imagery with a panchromatic stereo resolution superior to 0.8 m, with 0.8 m for forward and 0.64 m for backward. Our experimental findings revealed an area in Zhengzhou, China with an image size measuring $35864 \times 40000$ pixels. Furthermore, a local reference DSM measuring $5717 \times 6043$ pixels is available as a benchmark in comparison experiments.
  
\end{itemize}

\subsubsection{Implementation details}

The experiments were conducted in a computing environment featuring an NVIDIA A100-PCI graphics card with a video memory of 40 GB. Python was used as the programming language, and VCcode served as the compiler. Four distinct 3D reconstruction methods were employed for experimental testing, and the implementation details are provided below.

\textbf{S2P.} Satellite Stereo Pipeline (S2P) is an automatic and modular stereo pipeline for pushbroom images. The images are divided into small tiles and processed in parallel using multiple processes to improve efficiency. In our experiment, the tile size was set to 500-1000 pixels, and the dense matching method in S2P was the default More Global Matching (MGM) algorithm \cite{MGM}. In the fusion step, an outlier removal threshold of 25 m was chosen (the same operation as in \cite{WhuISPRS}), and the height map outlier cleaning was set to false (otherwise, the completeness rate was meager). The other settings were left at their default values.

\textbf{LPS.} The Leica Photogrammetry Suite (LPS) \cite{Erdas} is a collection of digital software for processing photogrammetry and remote sensing data. The software performs binocular stereo reconstruction, thereby displaying the outcomes with maximum precision when using multi-view images as input.

%The platform boasts highly accurate and effective production tools for image processing and photogrammetry. Among other capabilities, it can process a range of aerospace and aeronautical sensors, offering functionality in image orientation and aerial triangulation, as well as digital terrain generation. Additionally, it is the first system to bring together remote sensing and photogrammetry in a single.

\textbf{Adapted COLMAP.} Adapted COLMAP (AC) \cite{kai2019} fits the pinhole camera model to the image-based RPC model, and then employs a computer vision reconstruction pipeline for 3D reconstruction. However, Adapted COLMAP cannot handle large-scale images, and the direct output of large-scale images is displayed as NA. Therefore, the pipeline was only run on the WHU TLC and DFC2019 datasets.

\textbf{Sat-MVSF.} 

\textbf{Ours.} 
We used our REPM pipeline to reconstruct the DSM for two datasets, the WHU-TLC and DFC2019 dataset, which are referred to as “Ours” in the experiments. In addition, the REPM pipeline was constructed based on the Adapted COLMAP, which we improved by incorporating the image partition module. We also improved the DSM generation module to reconstruct the DSM, which is referred to as “REPM” in the experiments. When the image refinement model is introduced, it is called “REPM+Ref.” in the experiments. The crop size represents both length and width in all experiments. 

\subsubsection{Accuracy metrics}\label{sec: ExperimentsSettings}

We utilized the assessment metric codes established in the literature \cite{kai2019}, and the equations employed to calculate the assessment metrics are as follows.

(1) The root-mean-square error (RMSE), which is the standard deviation of the residuals between the ground truth and the estimation, is defined as 

\begin{equation}
  RMSE=\sqrt{\frac{(\hat{h}_{i}-h_{i})^{2}}{N_{i} }} (i\in (\hat{h}\cap h)),
  \label{eq:RMSE}
\end{equation}

where $\hat{h},h$ denote the predicted and true values, respectively, and $N_{x}$ denotes the number of computations required. For example, when calculating the RMSE accuracy of the reconstructed DSM, $\hat{h},h$ denote the heights of the generated DSM and true DSM, respectively, and $N_{i}$ denotes the number of pixels.

(2) The median error (ME), which is the median of the absolute values of the residuals between the ground truth and the estimation, is defined as 

\begin{equation}
  ME=median\left|\hat{h}_{i}-h_{i} \right | (i\in (\hat{h}\cap h)).
  \label{eq:ME}
\end{equation}

(3) The mean absolute error (MAE), which is the mean of the absolute values of the residuals between the ground truth and the estimation, is defined as 

\begin{equation}
  MAE=\frac{1}{n}\left ( \sum_{i=1}^{n} \left|\hat{h}_{i}-h_{i} \right | \right ) (i\in (\hat{h}\cap h)).
  \label{eq:MAE}
\end{equation}

(4) Completeness, which is the percentage of points with a height error less than a certain threshold. In this study, the completeness is denoted as $Comp_{threshold}$ and defined as

\begin{equation}
  Comp_{threshold} = \frac{N_{\left | \hat{h}_{i}-h_{i} \right | < threshold}}{N_{i}} (i\in h).
  \label{eq:comp1}
\end{equation}

(5) Time consumption, which is the time that elapses between the input of an image and the generation of a DSM product. The unit of time is min.

\subsection{Comparative results on benchmark datasets}

We evaluated the proposed pipeline on four datasets, and compared its performance to the experimental results obtained from the image reconstruction pipeline for classic linear-array satellite CCDs \cite{s2p} and from commercial software \cite{Erdas}.

%The WHU-TLC and DFC2019 datasets are standard datasets that have undergone processing and are directly applicable as pipeline input. Both ISPRS-ZY3 and GF7 data are raw satellite CCD images, and we select the image size with the maximum equivalent error not exceeding 1 (see Table\ref{tab:error}) for cropping.

\subsubsection{Results for the WHU-TLC test set}

To evaluate the accuracy and completeness of our proposed reconstruction pipeline, we compared its performance to that of other reconstruction methods using the WHU-TLC test set. To ensure fairness, we used more straightforward point-cloud filtering for postprocessing. Table \ref{tab:WHU_back} compares the experimental results for the WHU-TLC test set contained in the literature \cite{WhuISPRS} to our results. The table shows that our method outperformed other methods in terms of accuracy and completeness and that there was no significant decrease in the running time of our method. Furthermore, compared to the Adapted COLMAP approach, the addition of our image refinement model led to a 5.52\% increase in the RMSE accuracy and a 2.57\% improvement in the completeness for a threshold of 2.5 m.

Table \ref{tab:WHU_back} shows that our method improves both the accuracy and completeness. The poor RMSE accuracy of LPS can be attributed to the WHU-TLC test set, which is comprised of 46 image sets, some of which are occluded by clouds, whereas others have weakly textured regions such as water bodies. Because of LPS's inability to densely match such regions, a triangulation mesh is used to fill in the gaps, and the DSM outliers cannot be removed by point-cloud filtering, eventually resulting in an overall poor RMSE accuracy.

%Methods	&\makecell{ME\\(m)\textcolor{red}{$\downarrow$}}	&\makecell{RMSE\\(m)\textcolor{red}{$\downarrow$}}	&\makecell{Comp$_{2.5}$\\(\%)\textcolor{red}{$\uparrow$}}	&\makecell{Comp$_{5}$\\(\%)\textcolor{red}{$\uparrow$}}	&\makecell{Time\\(min.)\textcolor{red}{$\downarrow$}} 
%S2P &\textbf{0.996}	&8.710	&66.738	 &77.820	 & 7.487 \\
% LPS &1.970	&13.934	&62.242	&83.578	& 8.062 \\
% Adapted COLMAP & 1.146 &3.334	&73.926	& 88.244 & \textbf{7.171}\\
% Ours & 1.004	&\textbf{3.150}	& \textbf{75.826}	& \textbf{88.423} & 7.317\\

\begin{table*}[htbp]
%\scriptsize
\footnotesize
  \caption{\label{tab:WHU_back}  Quantitative results of DSM reconstruction quality on the WHU-TLC test set. (Bold works best)}
  \centering
\begin{threeparttable}
  \begin{tabular}{lccccc}
    \toprule
    \textbf{Methods}	&\textbf{MAE(m)\textcolor{red}{$\downarrow$}} &\textbf{RMSE(m)\textcolor{red}{$\downarrow$}}	&\textbf{Comp$_{2.5}$(\%)\textcolor{red}{$\uparrow$}}	&\textbf{Comp$_{5}$(\%)\textcolor{red}{$\uparrow$}}	&\textbf{Time(min.)\textcolor{red}{$\downarrow$}}	\\
    \midrule
    CATALYST\tnote{*} \cite{WhuISPRS} &  3.454  &7.939 &52.31 &82.52 & \textbf{3.80}\\
    Metashape\tnote{*} \cite{WhuISPRS}& 2.693 & 13.047 & 56.59 & 75.46 & 24.51\\
    SDRDIS \tnote{*} \cite{WhuISPRS} & 4.496 & 15.012 & 47.58 & 73.57 &9.41  \\
    Sat-MVSF \tnote{*} \cite{WhuISPRS} & 1.895 & 3.654 &64.82 & 80.05 & 5.87 \\
    \midrule
    S2P &1.692	&8.710	& 66.738 &94.652 & 7.487 \\
    LPS &3.581	&13.934	&62.242	&91.729	& 8.062 \\
    AC & 1.766 &3.334	&73.926	& \textbf{97.428} & 7.171\\
    Ours & \textbf{1.666}	&\textbf{3.150}	& \textbf{75.826}	& 97.308 & 7.317\\
    \bottomrule
  \end{tabular}
  \begin{tablenotes}
        \footnotesize
        \item[*] This is from the results of qualitative experiments on the WHU-TLC test set in \cite{WhuISPRS}. The time of Sat-MVSF represents inference time and does not include training time. 
      \end{tablenotes}
\end{threeparttable}
\end{table*}

\begin{figure*}[htbp]
	\centering
    %第一行图片展示  %左标题1
     \rotatebox{90}{\scriptsize{~~~~~~~~image1}}
    \subfigure{
		\begin{minipage}[b]{0.185\linewidth}
			\centering		\includegraphics[width=1\linewidth]{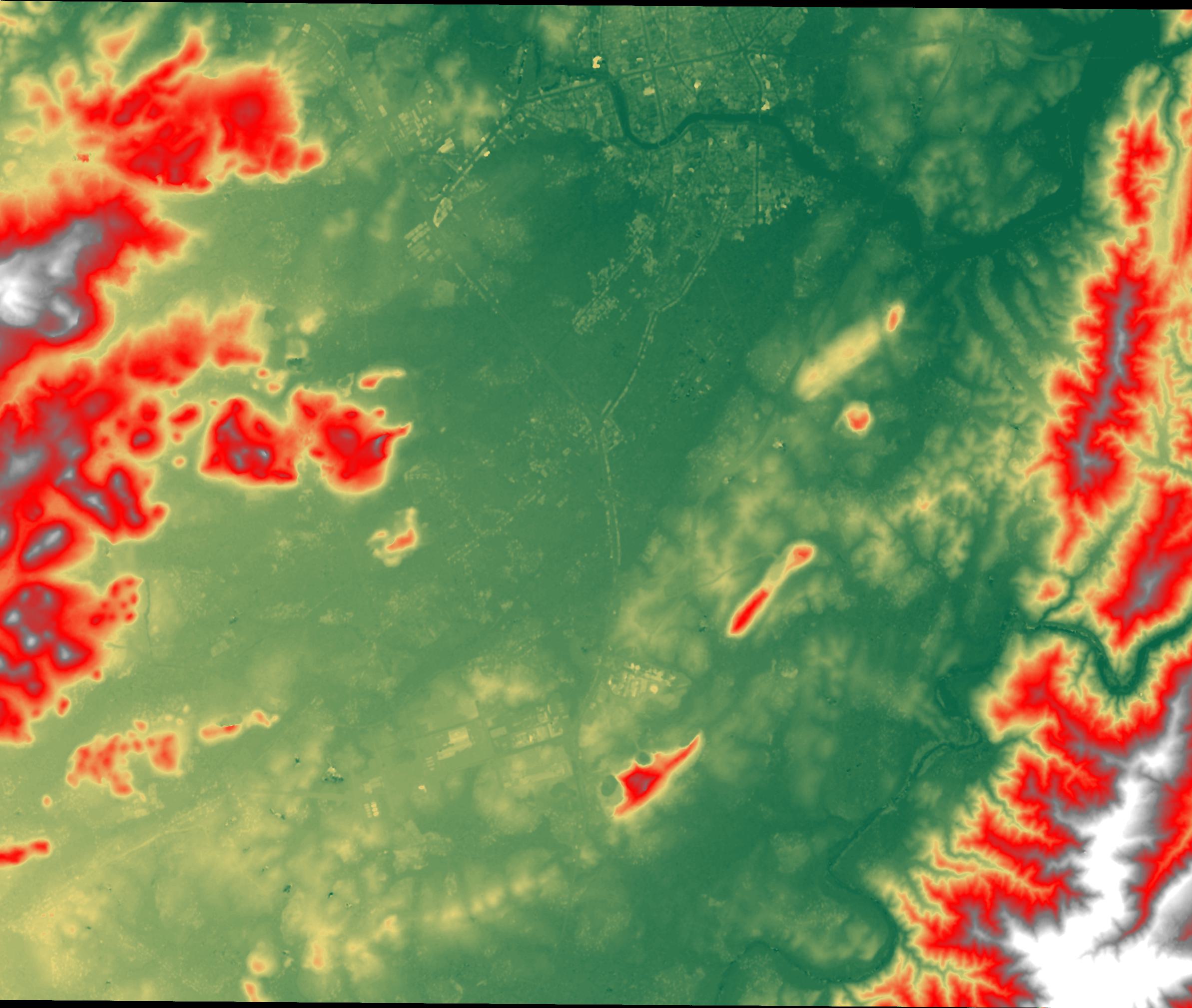}
		\end{minipage}
        \begin{minipage}[b]{0.185\linewidth}
			\centering		\includegraphics[width=1\linewidth]{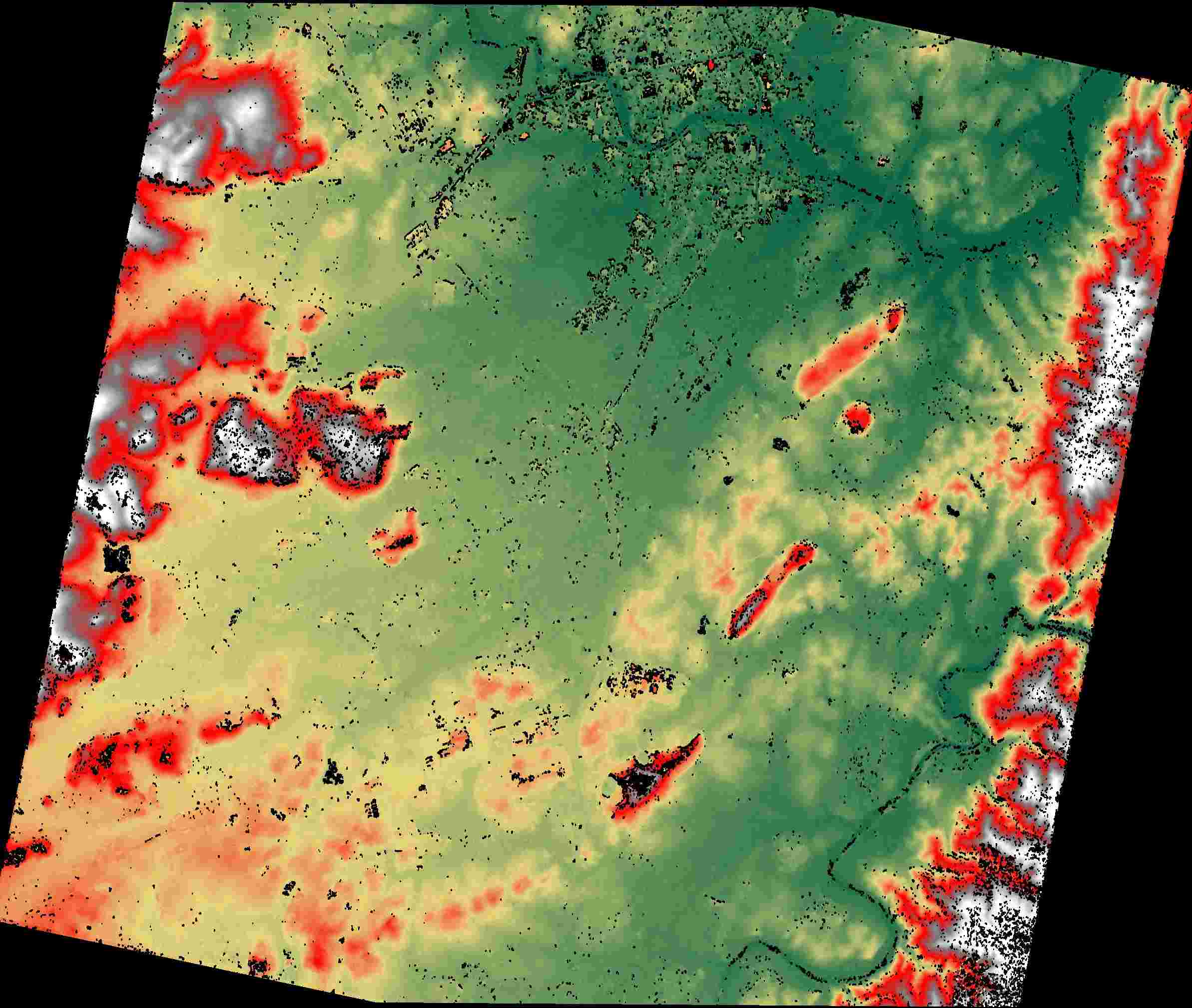}
		\end{minipage}
        \begin{minipage}[b]{0.185\linewidth}
			\centering	\includegraphics[width=1\linewidth]{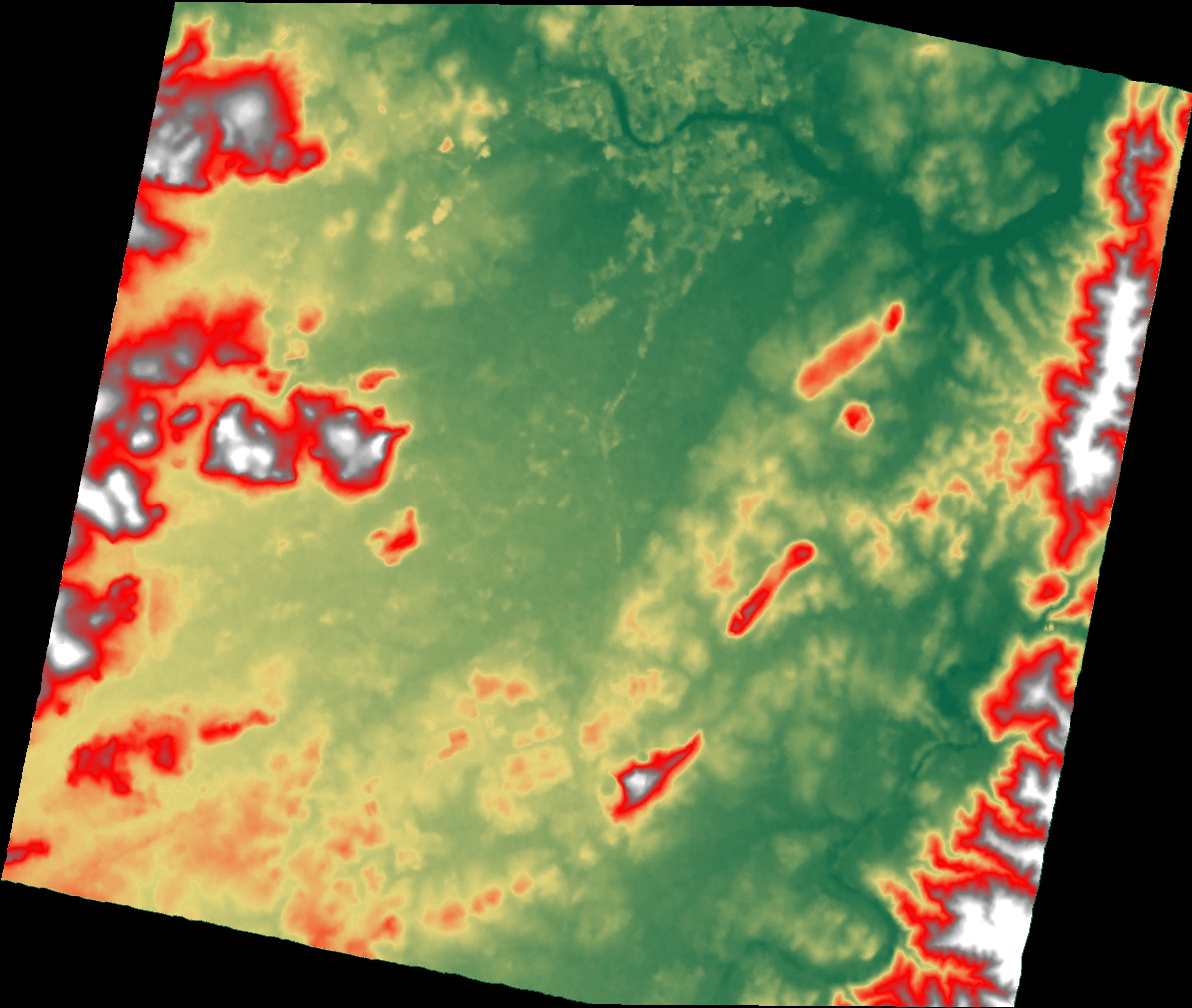}
		\end{minipage}
		\begin{minipage}[b]{0.185\linewidth}
			\centering	\includegraphics[width=1\linewidth]{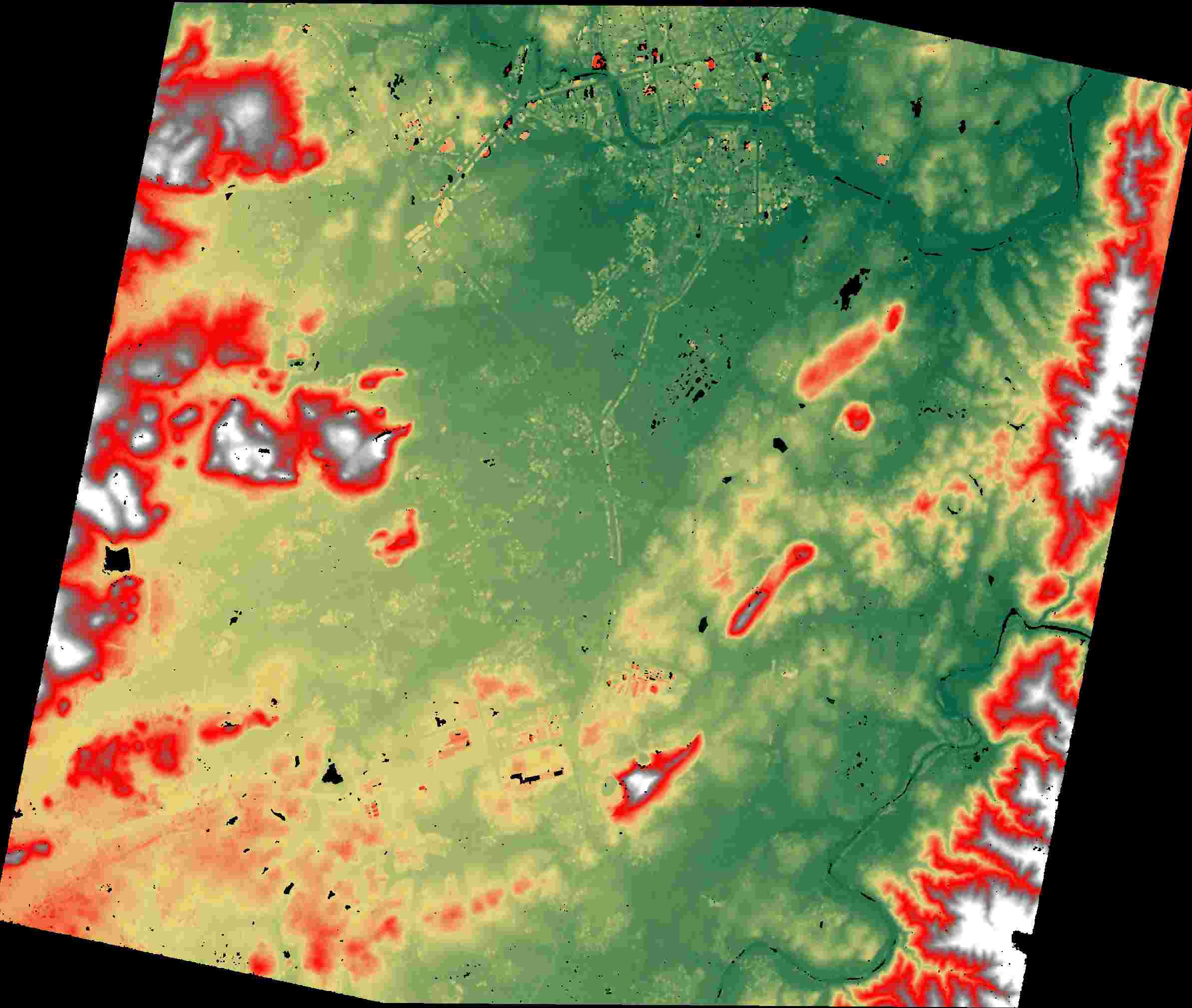}
		\end{minipage}
		\begin{minipage}[b]{0.185\linewidth}
			\centering	\includegraphics[width=1\linewidth]{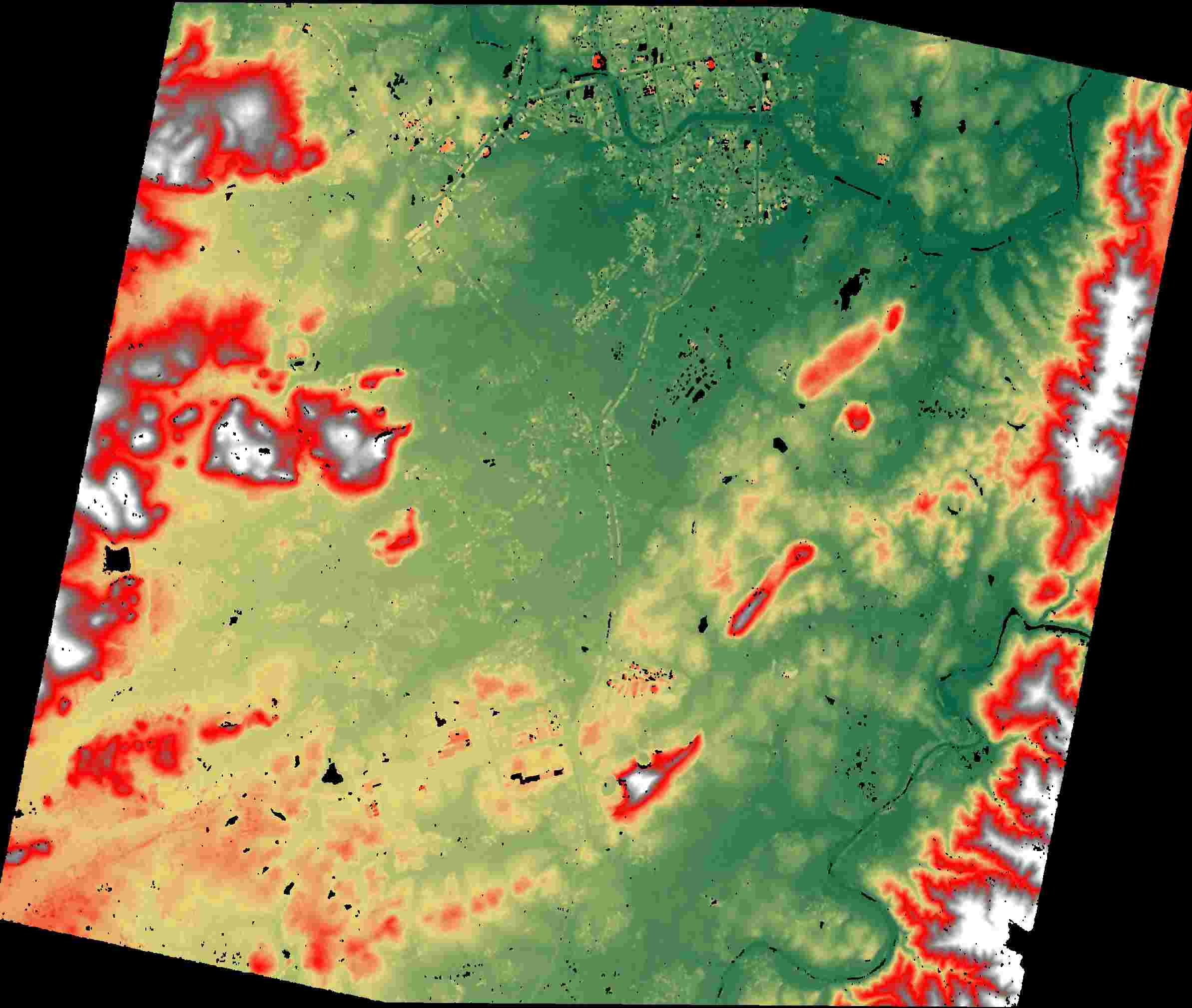}
		\end{minipage}
	}
    % 两行图片的间隙有点大，通过vspace进行微调
	\vspace{-3mm}
    % 由于上面已经用了subfigure，下面我们希望从 a 重新编号，而不是从 d 开始，清零。
	\setcounter{subfigure}{0}
 
    % 第二行图片展示
    \subfigure[GT]{
        % 左标题2
		\rotatebox{90}{\scriptsize{~~~~~~~~image2}}
		\begin{minipage}[b]{0.188\linewidth}
			\centering	\includegraphics[width=1\linewidth]{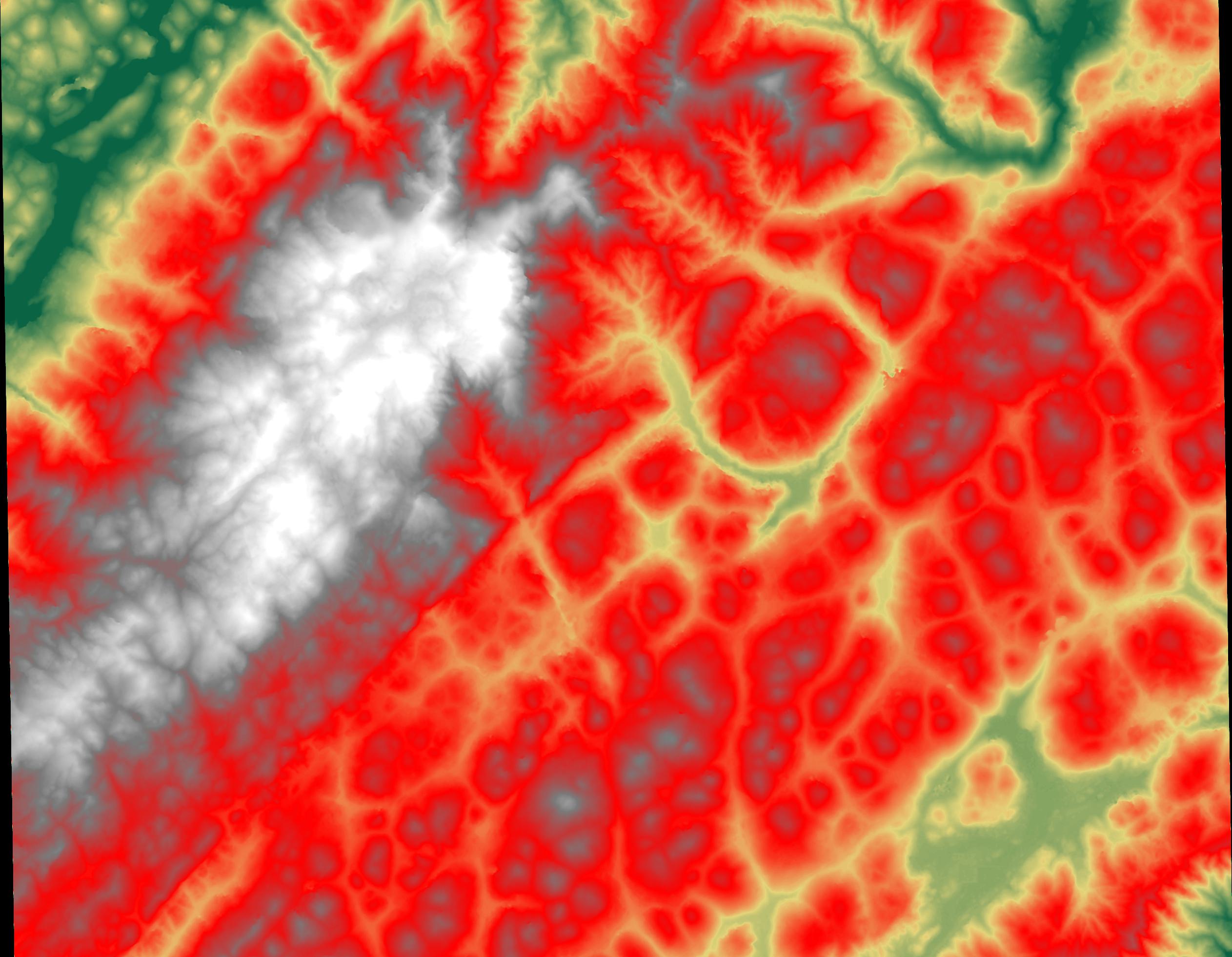}
		\end{minipage}
	}\hspace{-2.6mm}
	\subfigure[S2P]{
		\begin{minipage}[b]{0.188\linewidth}	\centering\includegraphics[width=1\linewidth]{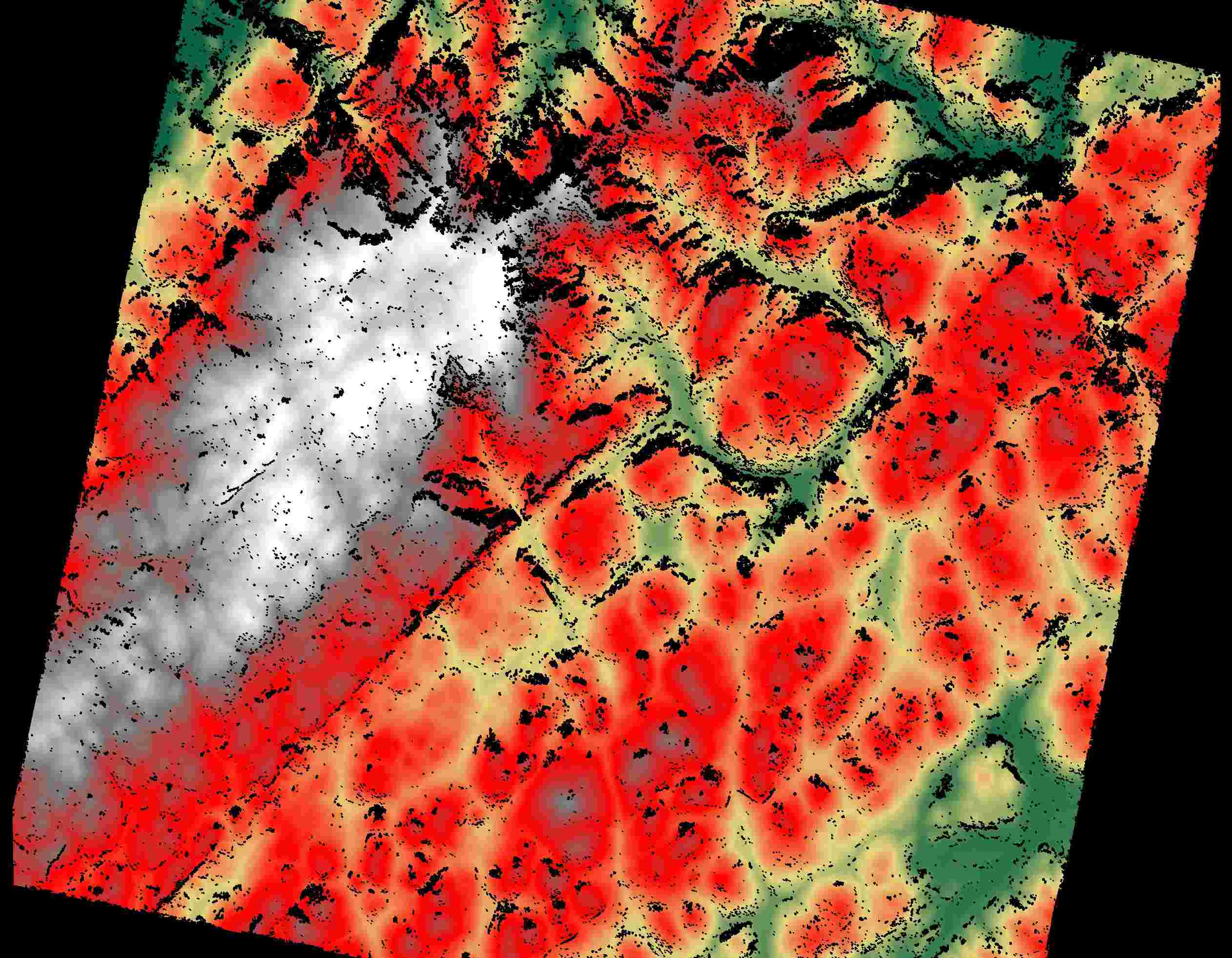}
		\end{minipage}
	}\hspace{-2.6mm}
	\subfigure[LPS]{
		\begin{minipage}[b]{0.188\linewidth}
			\centering	\includegraphics[width=1\linewidth]{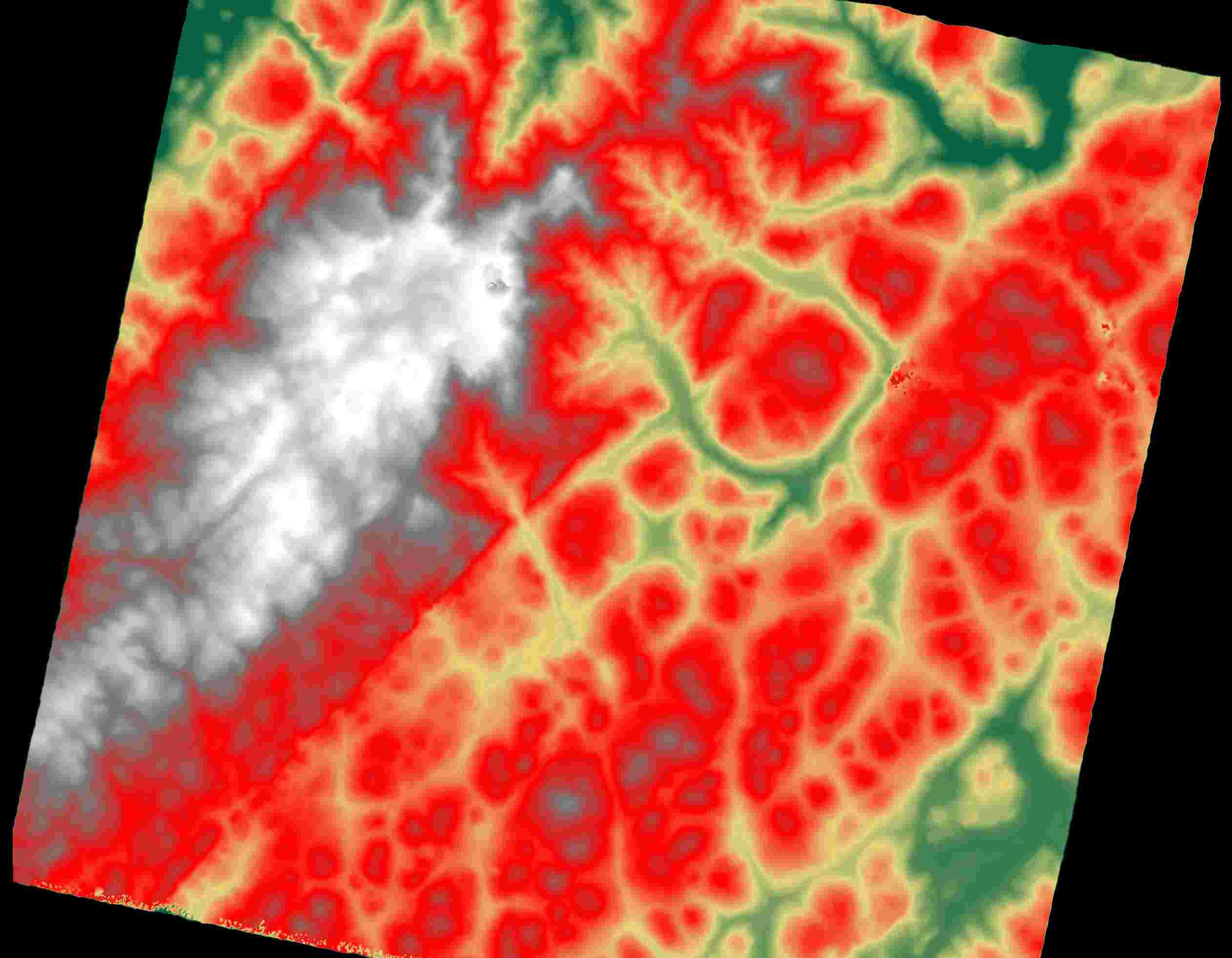}
		\end{minipage}}\hspace{-2mm}
    \subfigure[AC]{
		\begin{minipage}[b]{0.188\linewidth}
			\centering	\includegraphics[width=1\linewidth]{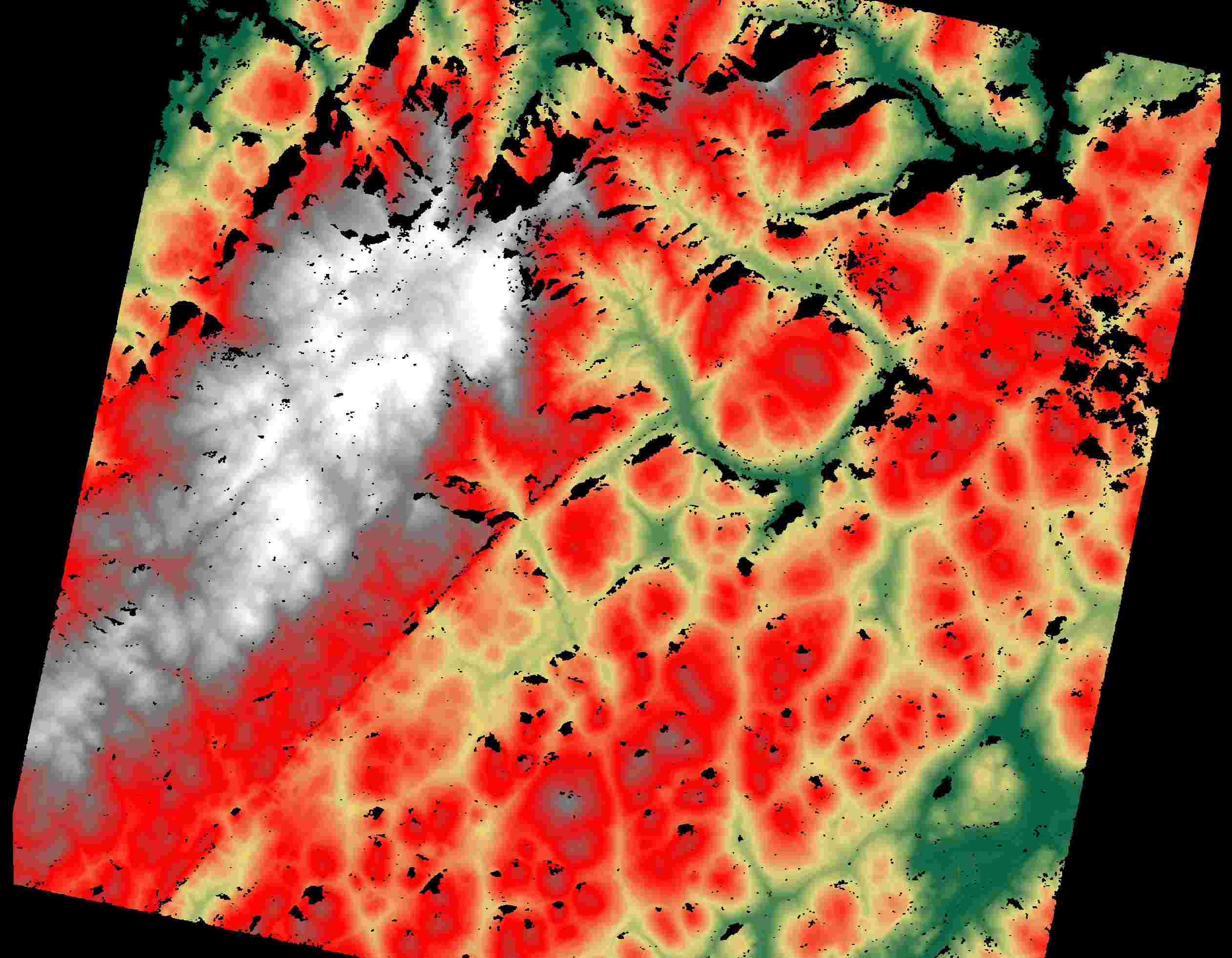}
		\end{minipage}}\hspace{-2mm}
    \subfigure[Ours]{
		\begin{minipage}[b]{0.188\linewidth}
			\centering		\includegraphics[width=1\linewidth]{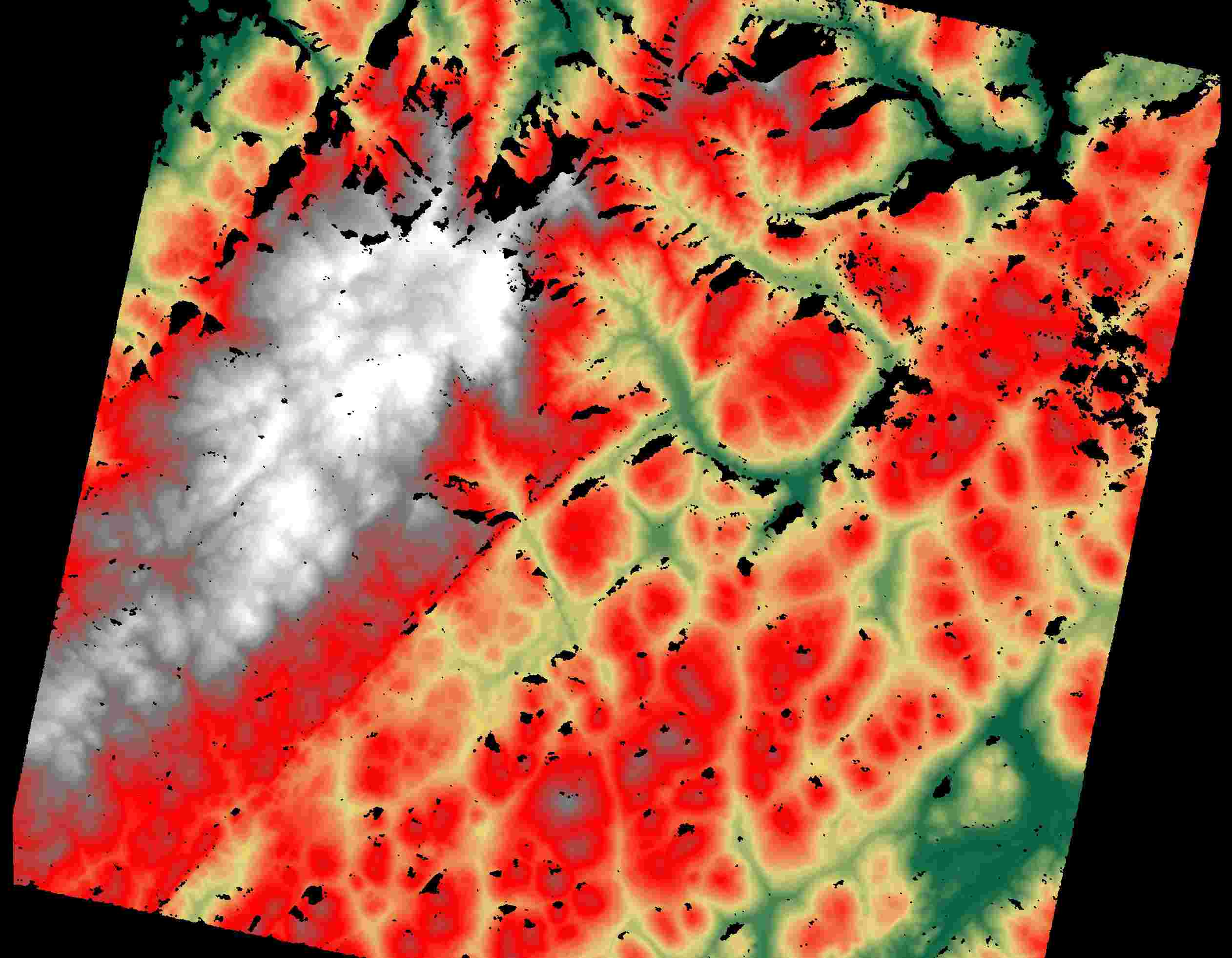}
		\end{minipage}
	}
 \vspace{-3mm}
 \rotatebox{90}{\scriptsize{~~~~~~~~image1}}
    \subfigure{
        \begin{minipage}[b]{0.2\linewidth}
			\centering		\includegraphics[width=1\linewidth]{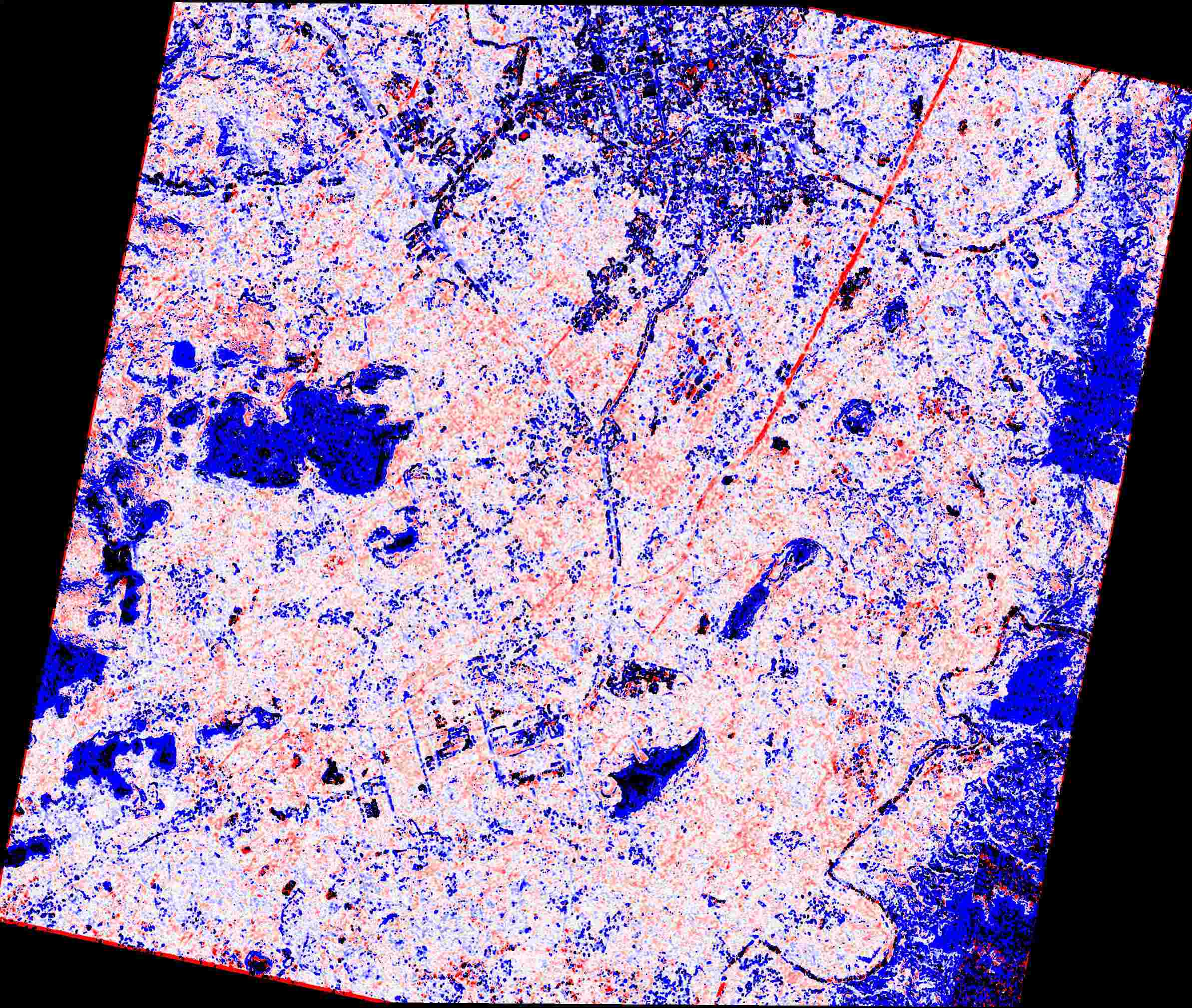}
		\end{minipage}
        \begin{minipage}[b]{0.2\linewidth}
			\centering	\includegraphics[width=1\linewidth]{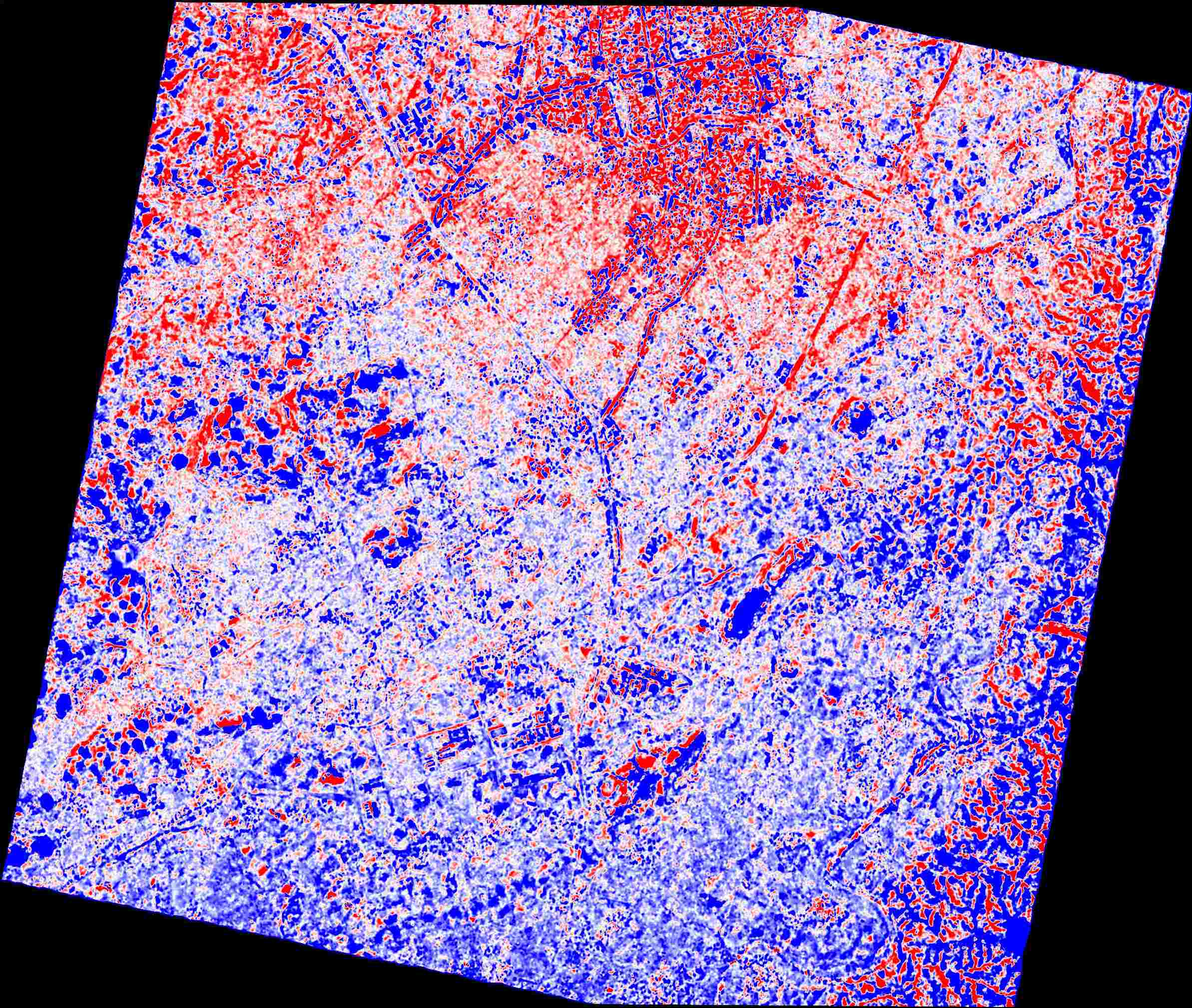}
		\end{minipage}
		\begin{minipage}[b]{0.2\linewidth}
			\centering	\includegraphics[width=1\linewidth]{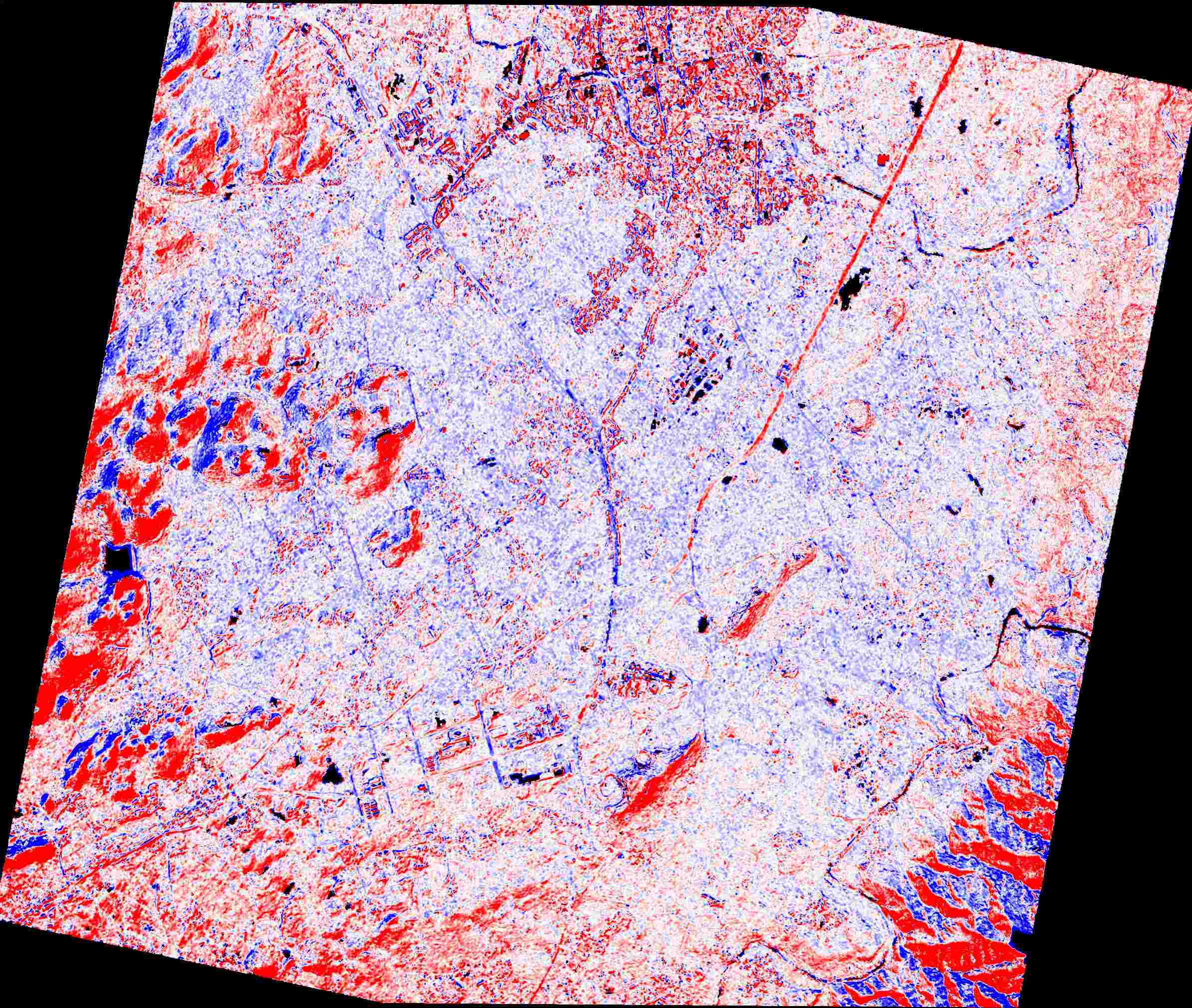}
		\end{minipage}
		\begin{minipage}[b]{0.2\linewidth}
			\centering	\includegraphics[width=1\linewidth]{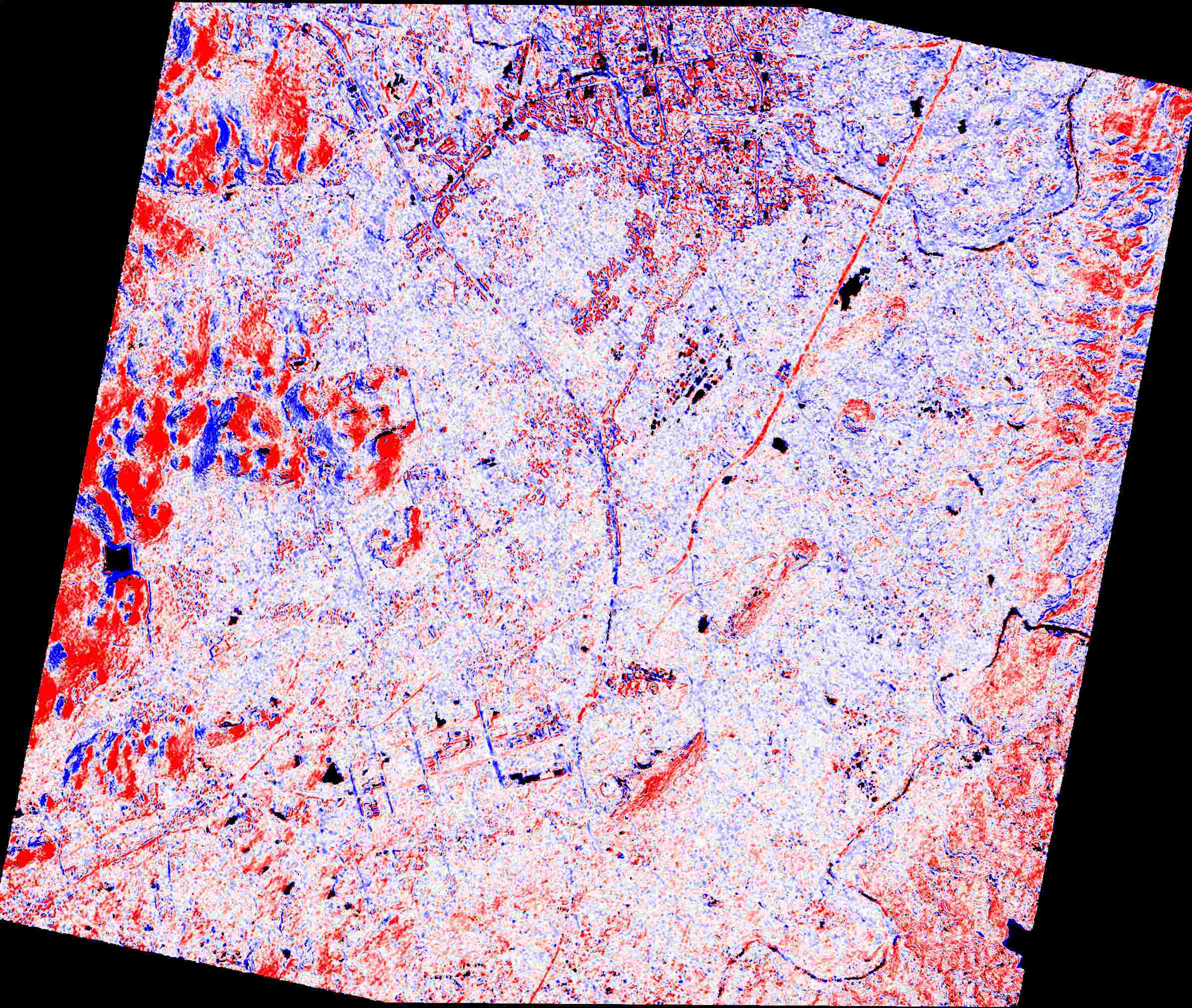}
		\end{minipage}
	}
    % 两行图片的间隙有点大，通过vspace进行微调
	%\vspace{-3mm}
    % 由于上面已经用了subfigure，下面我们希望从 a 重新编号，而不是从 d 开始，清零。
	\setcounter{subfigure}{5}
    % 第二行图片展示% 左标题2
    
	\subfigure[S2P]{
    \rotatebox{90}{\scriptsize{~~~~~~~~image2}}
		\begin{minipage}[b]{0.2\linewidth}	\centering\includegraphics[width=1\linewidth]{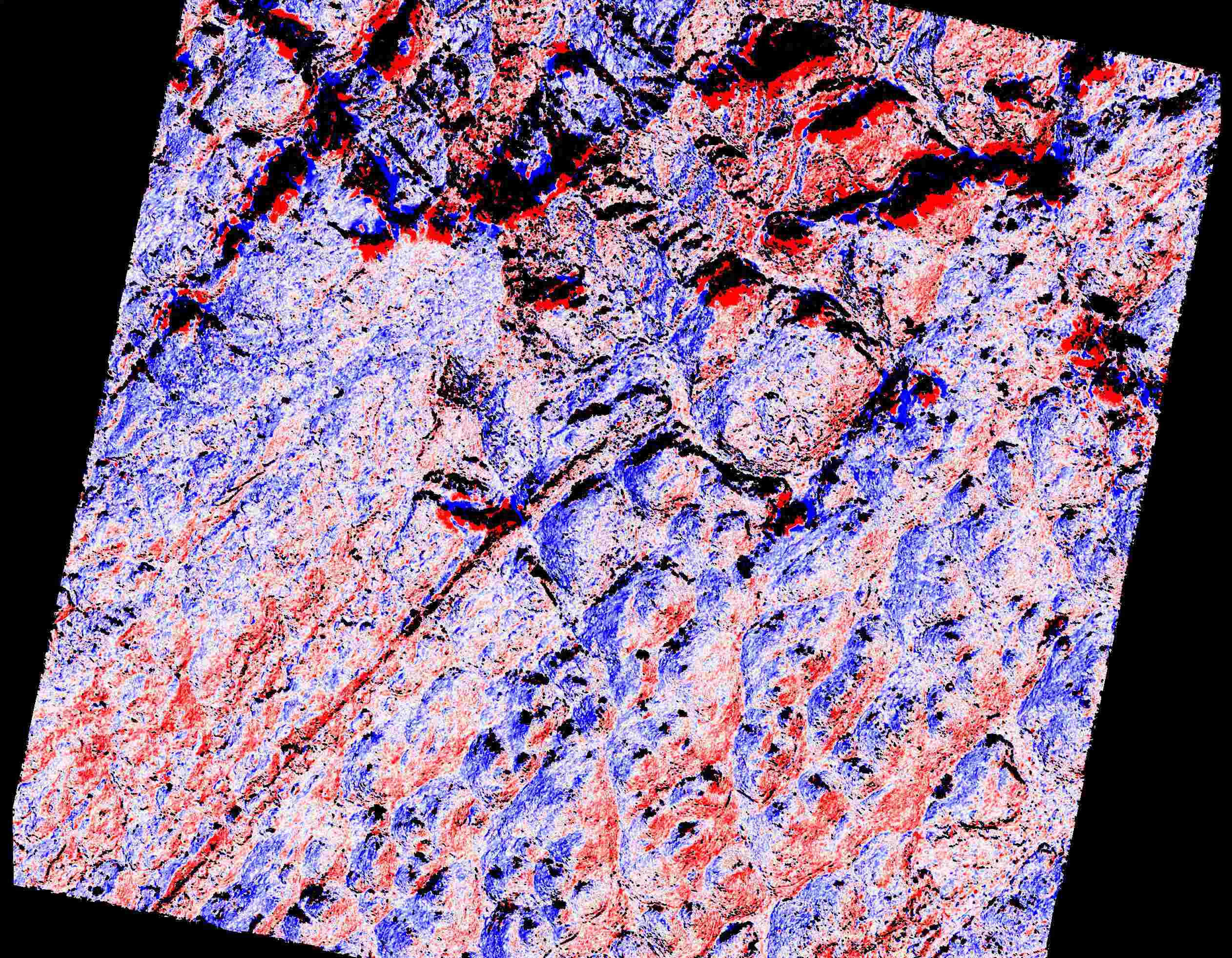}
		\end{minipage}
	}\hspace{-2.7mm}
	\subfigure[LPS]{
		\begin{minipage}[b]{0.2\linewidth}
			\centering	\includegraphics[width=1\linewidth]{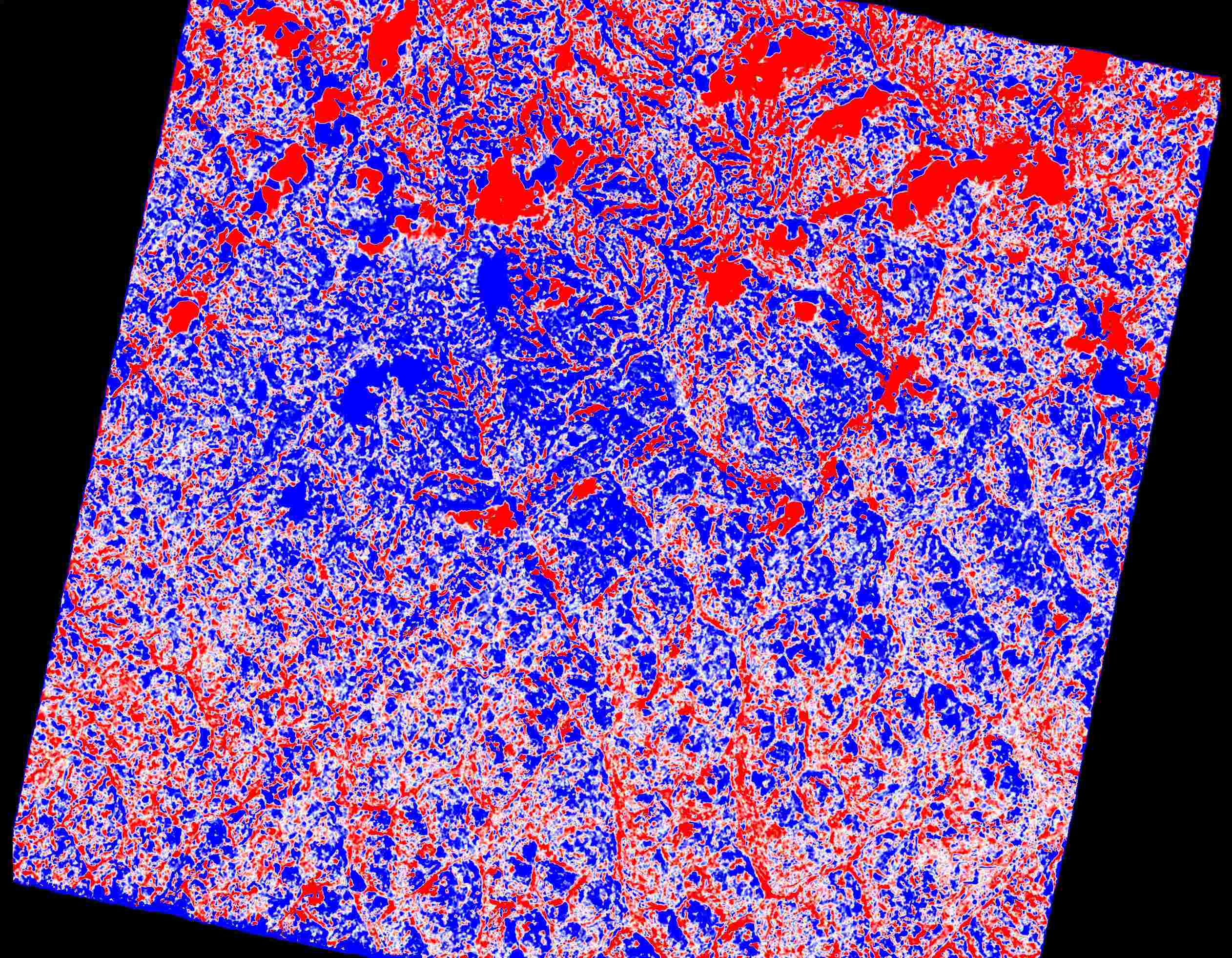}
		\end{minipage}}\hspace{-2mm}
    \subfigure[AC]{
		\begin{minipage}[b]{0.2\linewidth}
			\centering	\includegraphics[width=1\linewidth]{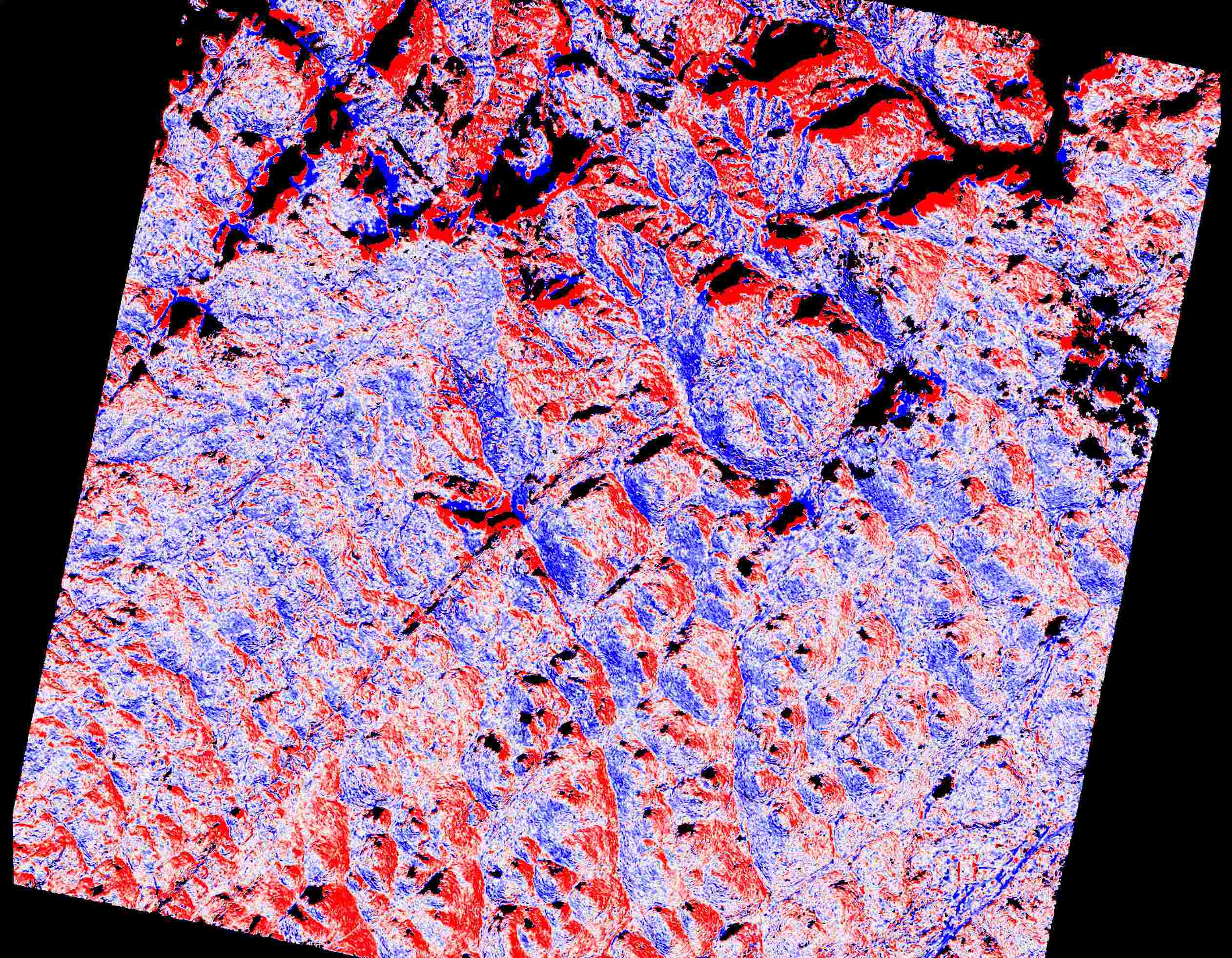}
		\end{minipage}}\hspace{-2mm}
    \subfigure[Ours]{
		\begin{minipage}[b]{0.2\linewidth}
			\centering		\includegraphics[width=1\linewidth]{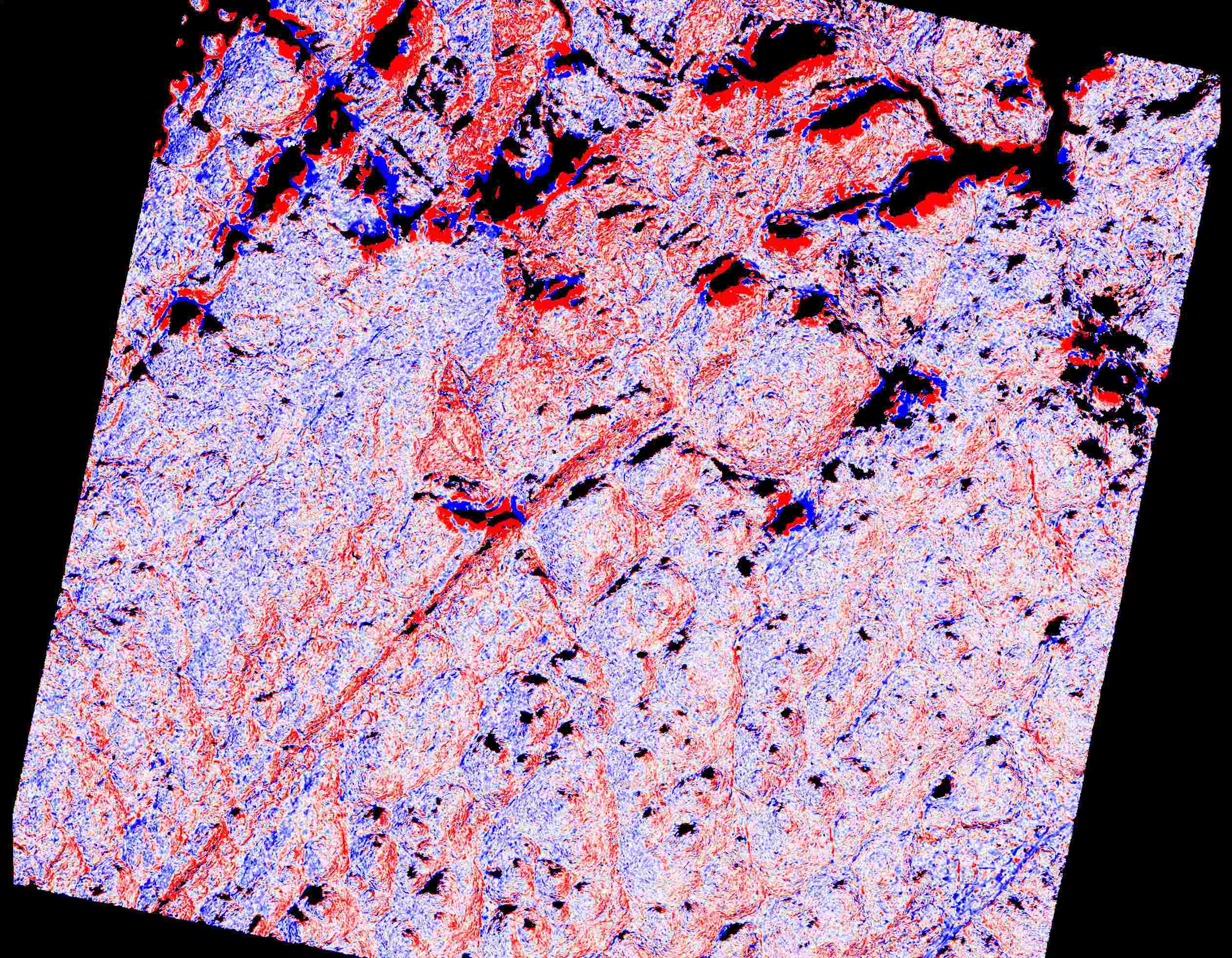}
		\end{minipage}
	}
 
    \subfigure[Image1 bar]{
    \begin{minipage}[b]{0.3\textwidth}
    \includegraphics[width=1\linewidth]{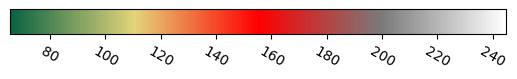}\vspace{4pt}
    \end{minipage}}
    \subfigure[Image2 bar]{
    \begin{minipage}[b]{0.3\textwidth}
    \includegraphics[width=1\linewidth]{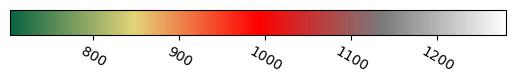}\vspace{4pt}
    \end{minipage}}
    \subfigure[Error bar]{
    \begin{minipage}[b]{0.3\textwidth}
    \includegraphics[width=1\linewidth]{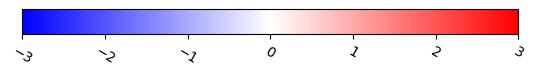}\vspace{4pt}
    \end{minipage}
    }
 
	% 添加题注，即对这个图片的说明
	\caption{Examples of DSMs and error plots for different solutions of the WHU-TLC test set. The color bar is expressed in units of meters.} \label{fig:WHU_benchmark}
\end{figure*}

A comparison between the DSM reconstruction output and the error maps for the WHU test set is shown in Fig. \ref{fig:WHU_benchmark}. S2P had more pixels with reconstruction failures than LPS, and all the pixels from LPS were successfully reconstructed. This discrepancy may be attributed to the provision of a low-resolution digital elevation model (DEM) by the ERDAS LPS software. A comparison of the error maps demonstrates that the iamge refinement model significantly enhanced the reconstruction accuracy and that our method had the highest reconstruction accuracy.

\subsubsection{Results for the DFC2019 dataset}

We conducted a comparative study using four prevalent sites in the DFC2019 dataset. For each of the four sites, we performed simple artificial masking of water bodies based on pixel values. Each site includes various satellite images obtained from different dates, and S2P and our pipeline could perform multi-view 3D reconstruction. LPS generated a DSM for every two images, and we selected the one with the highest accuracy, as listed in Table \ref{tab:DFC}. In the DFC2019 dataset, the accuracy of our method and the Adapted COLMAP were essentially the same, but the completeness of our method was significantly higher than that of the other methods.

%S2P only supports two-view and three-view, here we arbitrarily select three images per site for our experiments. In the DFC2019 dataset, the precision and completeness of our method is almost unchanged after the addition of the correction module. This is due to the equivalent error of this dataset in Table \ref{tab:error} achieving the desired accuracy. Furthermore, This aligns with the conjecture that the correction model is better suited for large scale images.

\begin{table}[htbp]
\footnotesize
  \caption{\label{tab:DFC}  Quantitative results of DSM reconstruction quality on the WoildView3 image data. (Bold works best).}
  \centering
  \begin{tabularx}{0.5\textwidth}{XXXXXX}
    \toprule
    \textbf{Site} &\textbf{Methods}	&\textbf{\makecell{ME\\(m)\textcolor{red}{$\downarrow$}}}	& \textbf{\makecell{RMSE \\ (m)\textcolor{red}{$\downarrow$}}}	& \textbf{\makecell{Comp$_{1}$ \\(\%)\textcolor{red}{$\uparrow$}}} &\textbf{\makecell{Time \\ (min.)\textcolor{red}{$\downarrow$}}}	\\
    
    \midrule
    \multirow{4}*{JAX\_004} 
    &S2P &\textbf{0.293}	& \textbf{2.822}	&60.318	 &40.192    \\~
    & LPS & 0.858	& 3.136	&53.763	& 12.440 \\~
    & AC & 0.381	&3.453	&63.737	&  \textbf{6.740}  \\~
    & Ours &0.386	&3.326	&\textbf{63.770}	  & 7.317        \\
    \midrule
    \multirow{4}*{JAX\_068} 
    &S2P & 0.202  &\textbf{2.467}	&69.430		 &11.281   \\~
    & LPS &0.993	&4.358	&50.193	&\textbf{10.143}      \\~
    & AC & \textbf{0.239}	&3.883	&\textbf{72.849}	& 13.279    \\~
    & Ours &0.260	&3.753	&72.450	&  14.406    \\
    \midrule
    \multirow{4}*{JAX\_214} 
    &S2P &\textbf{0.234} &\textbf{3.630}	&54.459	&39.185    \\~
    & LPS &1.463	&8.629	&42.943	&\textbf{14.305}         \\~
    & AC &0.326	&4.868	&65.957&  17.072      \\~
    &Ours & 0.353	& 4.893	&\textbf{66.014}	&  18.287   \\
    \midrule
    \multirow{4}*{JAX\_260} 
    &S2P &\textbf{0.577} &9.452	&36.045	 &30.047  \\~
    & LPS &1.600	&12.831	&38.122	&13.297           \\~
    & AC &0.592	&\textbf{3.812}	&59.938 &  \textbf{12.551}    \\~
    & Ours &0.599	&3.865	&\textbf{60.451}	&  13.390   \\
    	
    \bottomrule
  \end{tabularx}
\end{table}

\begin{figure}[htbp]
	\centering
    %第一行图片展示  %左标题1
    \rotatebox{90}{\tiny{~~~~~~~~JAX\_004}}
	\subfigure{
    \centering
		\begin{minipage}[b]{0.185\linewidth}
			\centering	\includegraphics[width=1\linewidth]{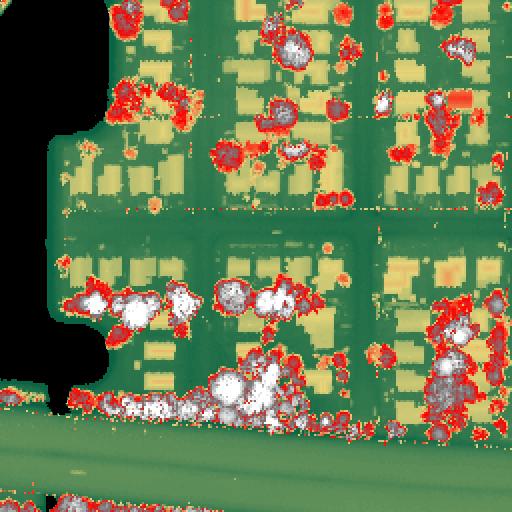}
		\end{minipage} \hspace{-2mm}
        \begin{minipage}[b]{0.185\linewidth}
			\centering	\includegraphics[width=1\linewidth]{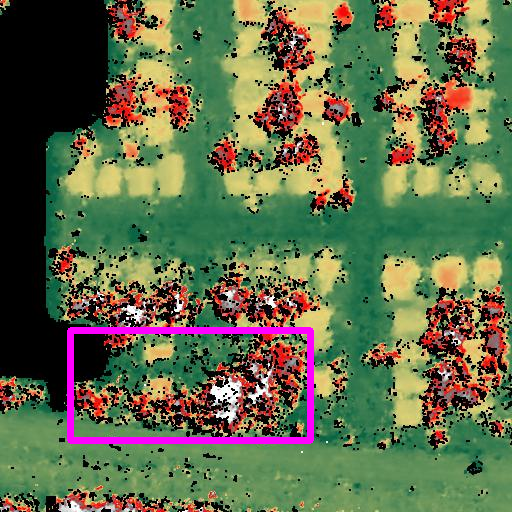}
		\end{minipage} \hspace{-2mm}
        \begin{minipage}[b]{0.185\linewidth}
			\centering	\includegraphics[width=1\linewidth]{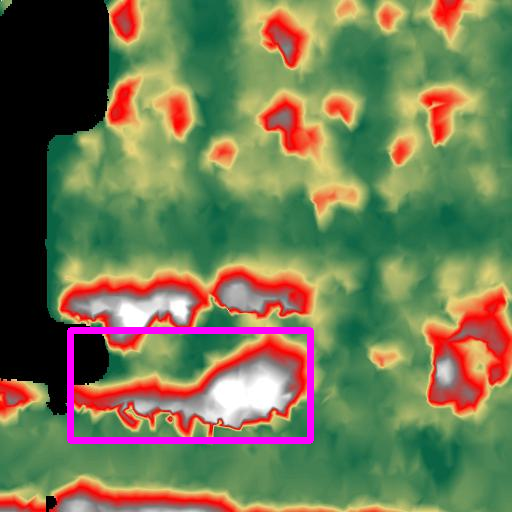}
		\end{minipage} \hspace{-2mm}
		\begin{minipage}[b]{0.185\linewidth}
			\centering	\includegraphics[width=1\linewidth]{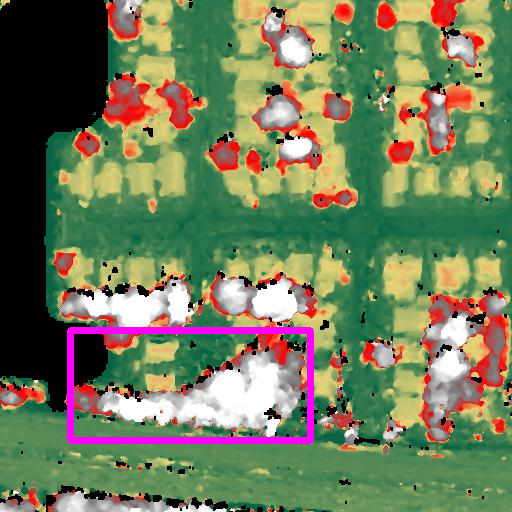}
		\end{minipage} \hspace{-2mm}
		\begin{minipage}[b]{0.185\linewidth}
			\centering	\includegraphics[width=1\linewidth]{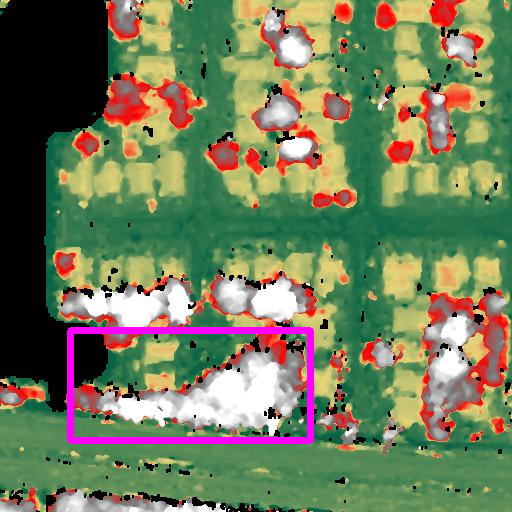}
		\end{minipage}
	}
    \vspace{-2mm}

    \rotatebox{90}{\tiny{~~~~~~~~JAX\_068}}
    \subfigure{
		\begin{minipage}[b]{0.185\linewidth}
		\centering		\includegraphics[width=1\linewidth]{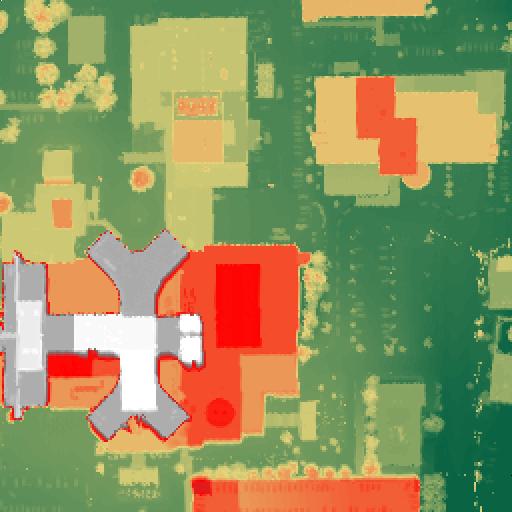}
		\end{minipage} \hspace{-2mm}
        \begin{minipage}[b]{0.185\linewidth}
			\centering	\includegraphics[width=1\linewidth]{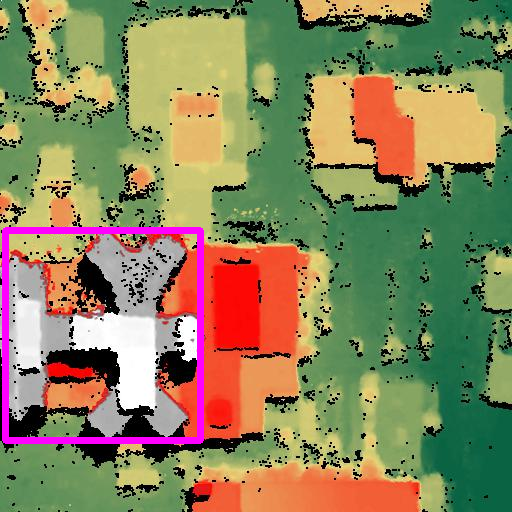}
		\end{minipage} \hspace{-2mm}
        \begin{minipage}[b]{0.185\linewidth}
			\centering	\includegraphics[width=1\linewidth]{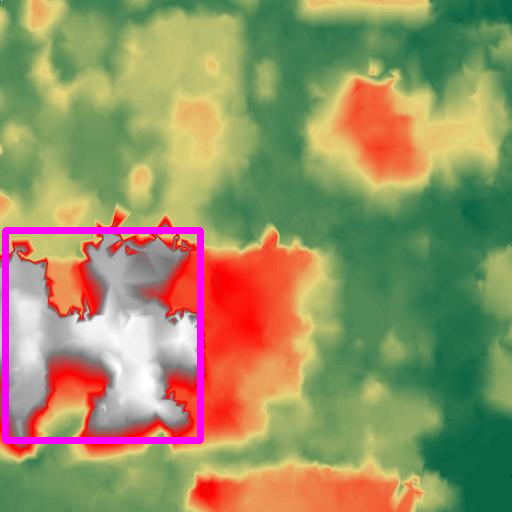}
		\end{minipage} \hspace{-2mm}
		\begin{minipage}[b]{0.185\linewidth}
			\centering	\includegraphics[width=1\linewidth]{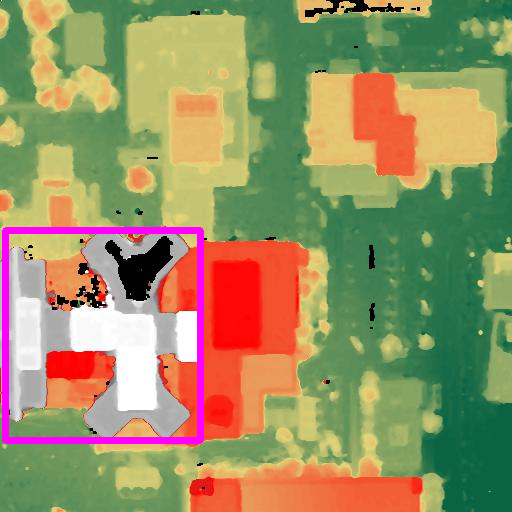}
		\end{minipage} \hspace{-2mm}
		\begin{minipage}[b]{0.185\linewidth}
			\centering	\includegraphics[width=1\linewidth]{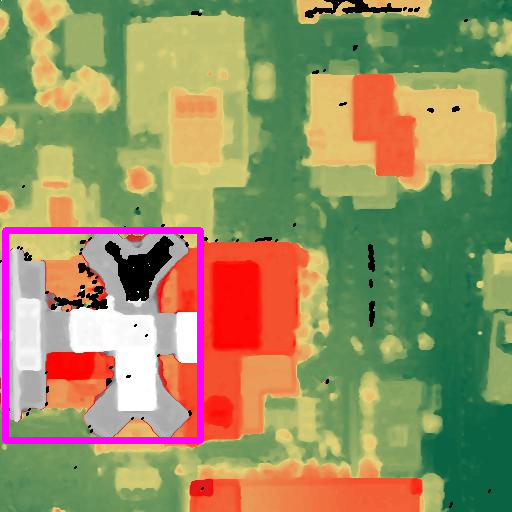}
		\end{minipage}
	}
    
     \rotatebox{90}{\tiny{~~~~~~~~JAX\_214}}
    \subfigure{
		\begin{minipage}[b]{0.185\linewidth}
			\centering		\includegraphics[width=1\linewidth]{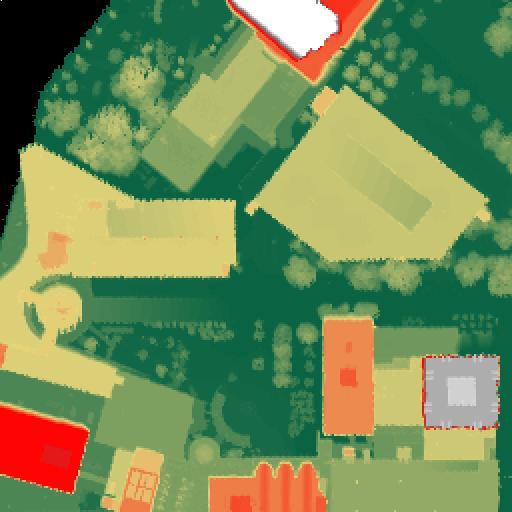}
		\end{minipage} \hspace{-2mm}
        \begin{minipage}[b]{0.185\linewidth}
			\centering		\includegraphics[width=1\linewidth]{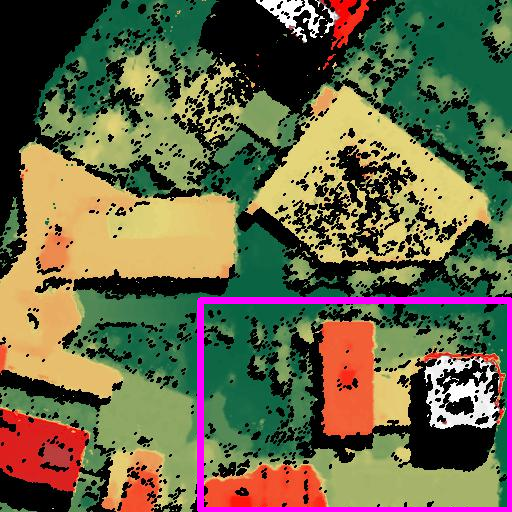}
		\end{minipage} \hspace{-2mm}
        \begin{minipage}[b]{0.185\linewidth}
			\centering	\includegraphics[width=1\linewidth]{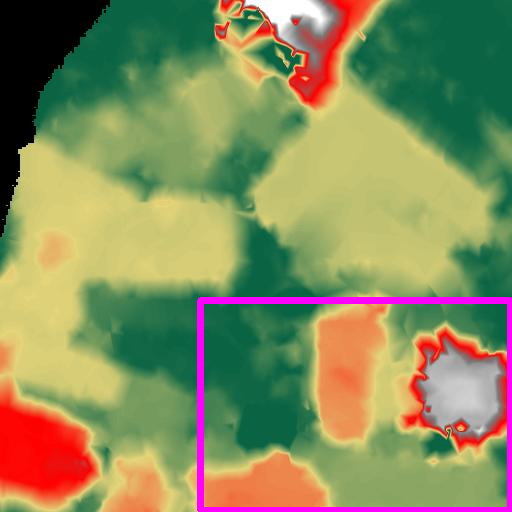}
		\end{minipage} \hspace{-2mm}
		\begin{minipage}[b]{0.185\linewidth}
			\centering	\includegraphics[width=1\linewidth]{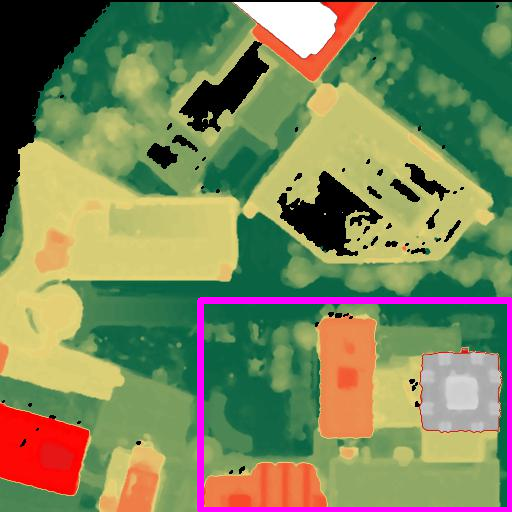}
		\end{minipage} \hspace{-2mm}
		\begin{minipage}[b]{0.185\linewidth}
			\centering	\includegraphics[width=1\linewidth]{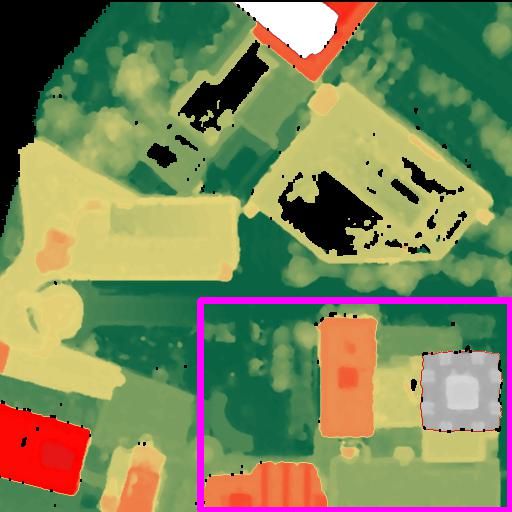}
		\end{minipage}
	}
    % 两行图片的间隙有点大，通过vspace进行微调
	%\vspace{-3mm}
    % 由于上面已经用了subfigure，下面我们希望从 a 重新编号，而不是从 d 开始，清零。
	\setcounter{subfigure}{0}
 
    % 第二行图片展示
    \subfigure[GT]{
        % 左标题2
		\rotatebox{90}{\tiny{~~~~~~~~JAX\_260}}
		\begin{minipage}[b]{0.185\linewidth}
			\centering	\includegraphics[width=1\linewidth]{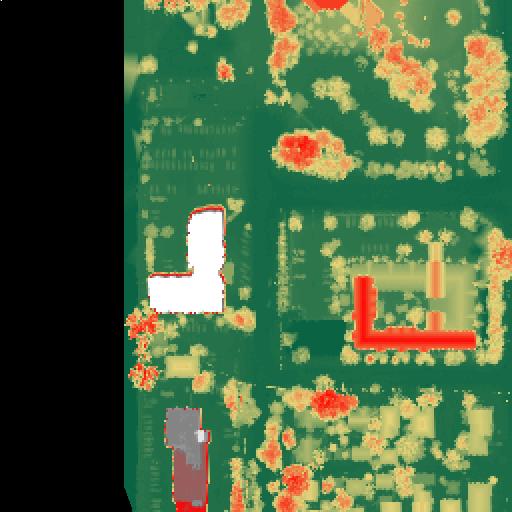}
		\end{minipage}
	}\hspace{-3mm}
	\subfigure[S2P]{
		\begin{minipage}[b]{0.185\linewidth}	\centering\includegraphics[width=1\linewidth]{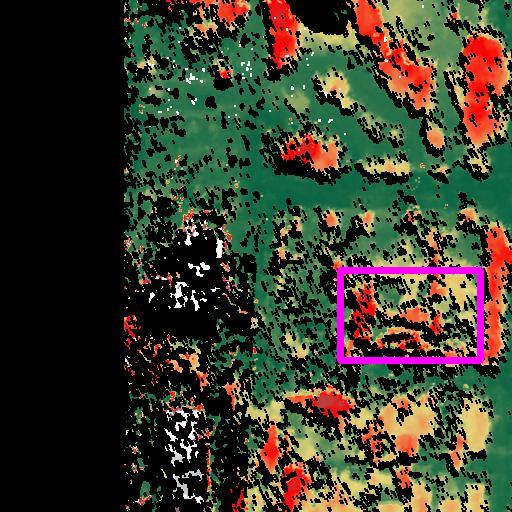}
		\end{minipage}
	}\hspace{-3mm}
	\subfigure[LPS]{
		\begin{minipage}[b]{0.185\linewidth}
			\centering	\includegraphics[width=1\linewidth]{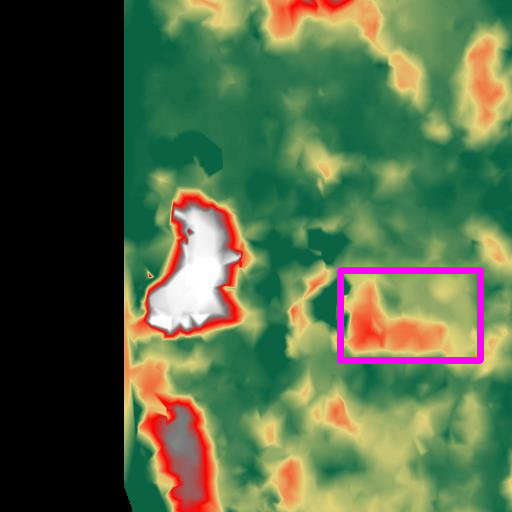}
		\end{minipage}} \hspace{-3mm}
    \subfigure[AC]{
		\begin{minipage}[b]{0.185\linewidth}
			\centering	\includegraphics[width=1\linewidth]{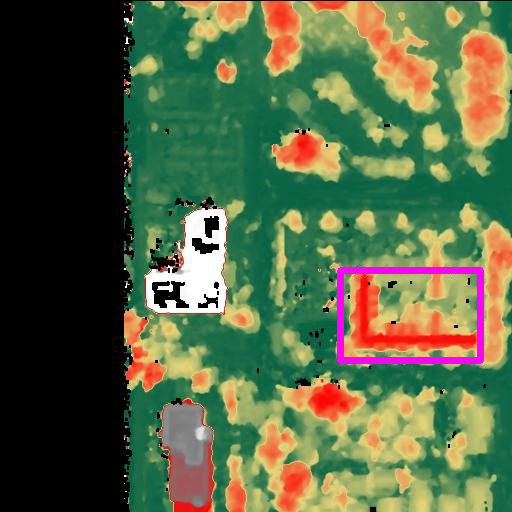}
		\end{minipage}} \hspace{-3mm}
    \subfigure[Ours]{
		\begin{minipage}[b]{0.185\linewidth}
			\centering		\includegraphics[width=1\linewidth]{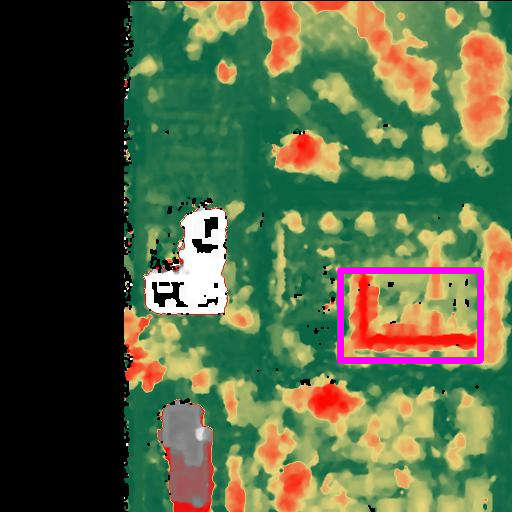}
		\end{minipage}
	}
    \subfigure[]{
    \begin{minipage}[b]{0.1\textwidth}
    \includegraphics[width=1\linewidth]{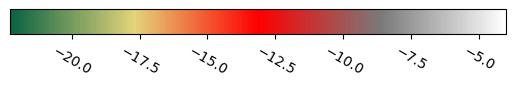}\vspace{4pt}
    \end{minipage}}
    \subfigure[]{
    \begin{minipage}[b]{0.1\textwidth}
    \includegraphics[width=1\linewidth]{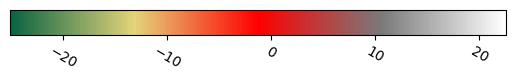}\vspace{4pt}
    \end{minipage}}
    \subfigure[]{
    \begin{minipage}[b]{0.1\textwidth}
    \includegraphics[width=1\linewidth]{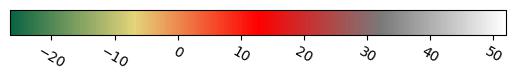}\vspace{4pt}
    \end{minipage}
    }
    \subfigure[]{
    \begin{minipage}[b]{0.1\textwidth}
    \includegraphics[width=1\linewidth]{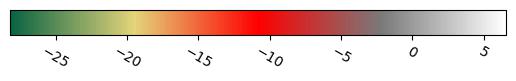}\vspace{4pt}
    \end{minipage}
    }
 
	% 添加题注，即对这个图片的说明
	\caption{DSM results for the DFC dataset. The color bars of JAX\_004, JAX\_068, JAX\_214, and JAX\_260 are represented by (f), (g), (h), and (i), respectively. The color bar is expressed in units of meters.}
    \label{fig:DFC_benchmark}
\end{figure}

\begin{figure}[htbp]
\centering
\subfigure[S2P]{
\begin{minipage}[b]{0.115\textwidth}
\includegraphics[width=1\linewidth]{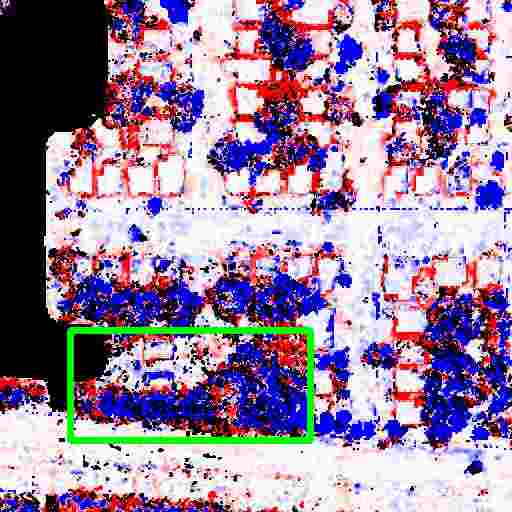}\vspace{2pt}
\includegraphics[width=1\linewidth]{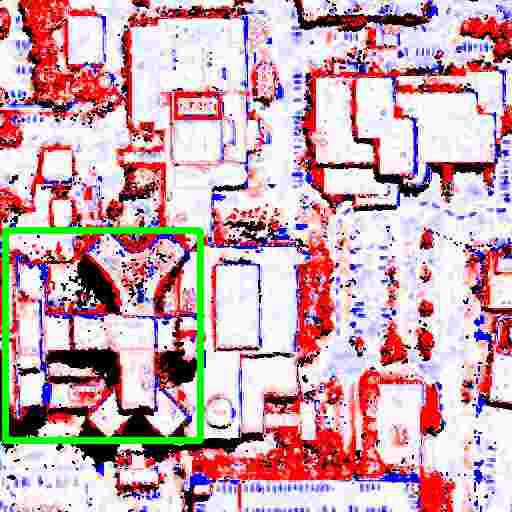}\vspace{2pt}
\includegraphics[width=1\linewidth]{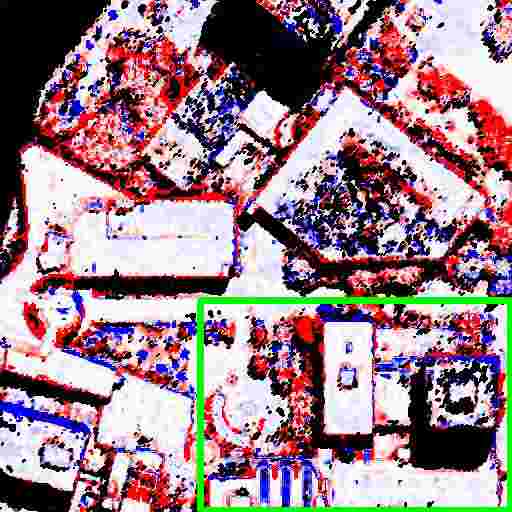}\vspace{2pt}
\includegraphics[width=1\linewidth]{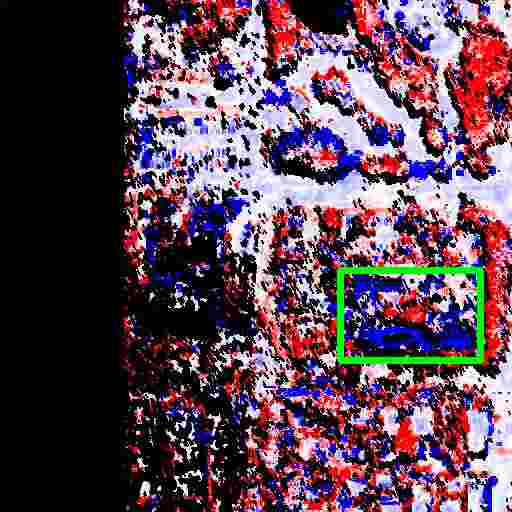}\vspace{2pt}
\end{minipage}}  \hspace{-3mm}
\subfigure[LPS]{
\begin{minipage}[b]{0.115\textwidth}
\includegraphics[width=1\linewidth]{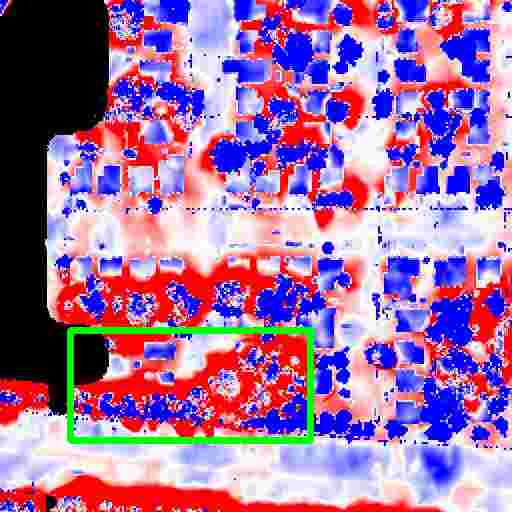}\vspace{2pt}
\includegraphics[width=1\linewidth]{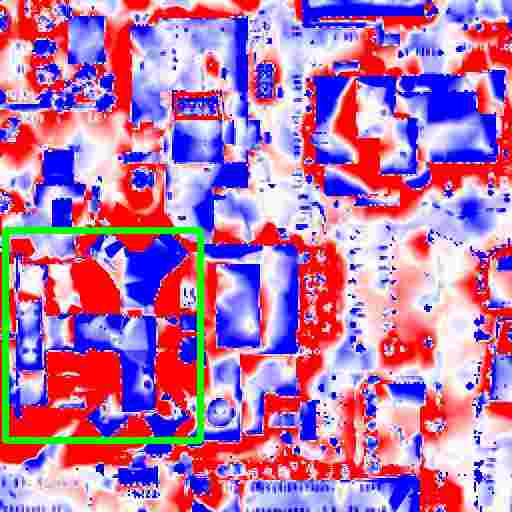}\vspace{2pt}
\includegraphics[width=1\linewidth]{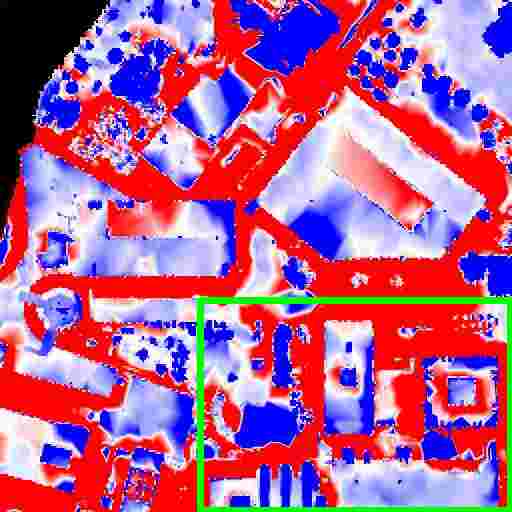}\vspace{2pt}
\includegraphics[width=1\linewidth]{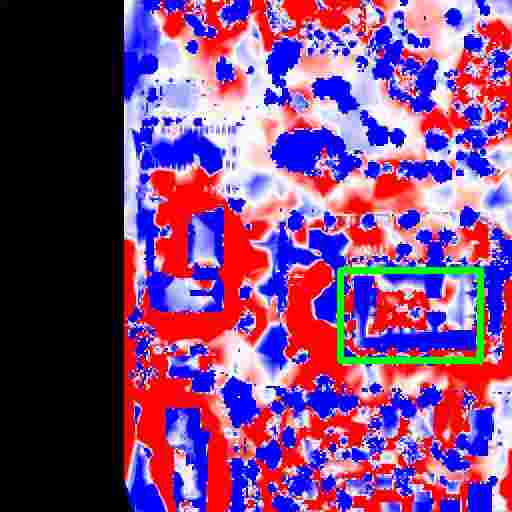}\vspace{2pt}
\end{minipage}}  \hspace{-3mm}
\subfigure[AC]{
\begin{minipage}[b]{0.115\textwidth}
\includegraphics[width=1\linewidth]{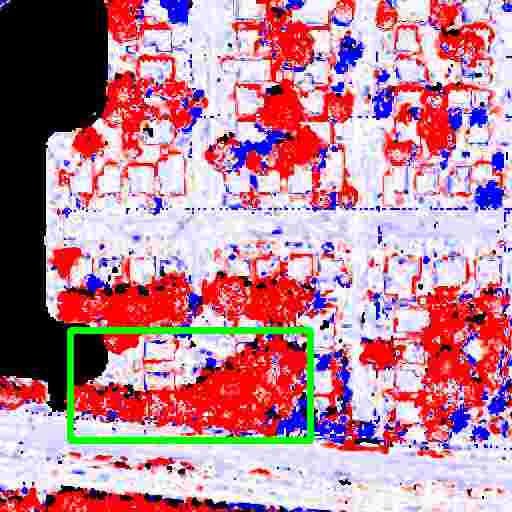}\vspace{2pt}
\includegraphics[width=1\linewidth]{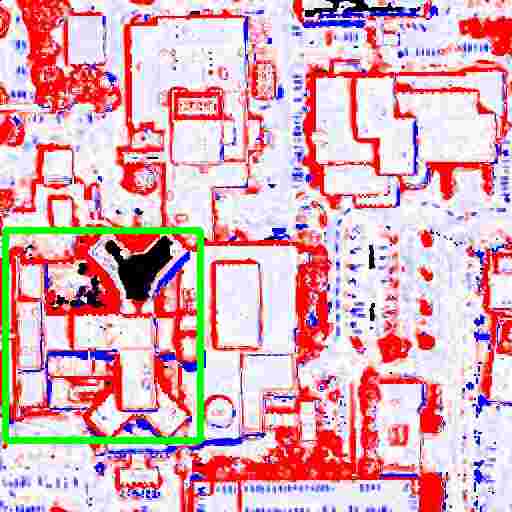}\vspace{2pt}
\includegraphics[width=1\linewidth]{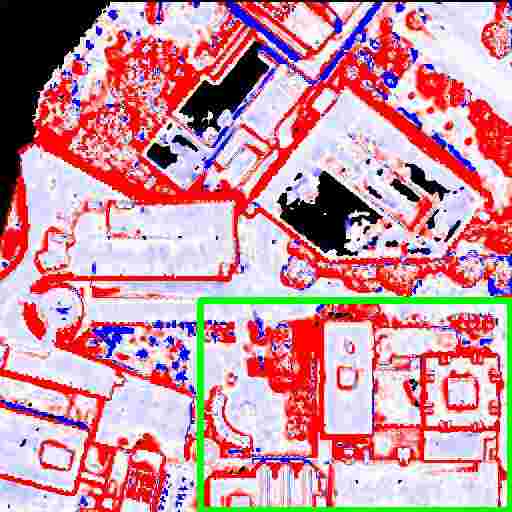}\vspace{2pt}
\includegraphics[width=1\linewidth]{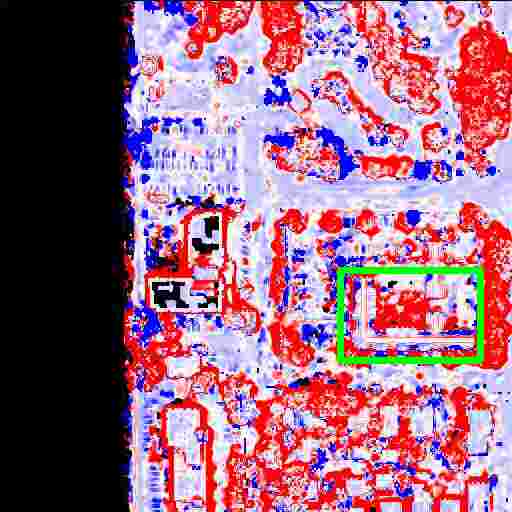}\vspace{2pt}
\end{minipage}}   \hspace{-3mm}
\subfigure[Ours]{
\begin{minipage}[b]{0.115\textwidth}
\includegraphics[width=1\linewidth]{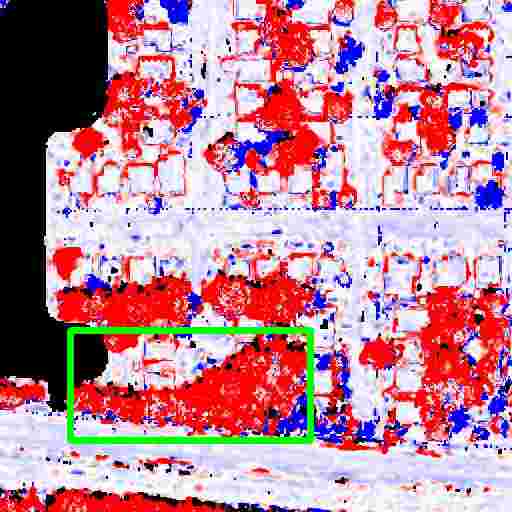}\vspace{2pt}
\includegraphics[width=1\linewidth]{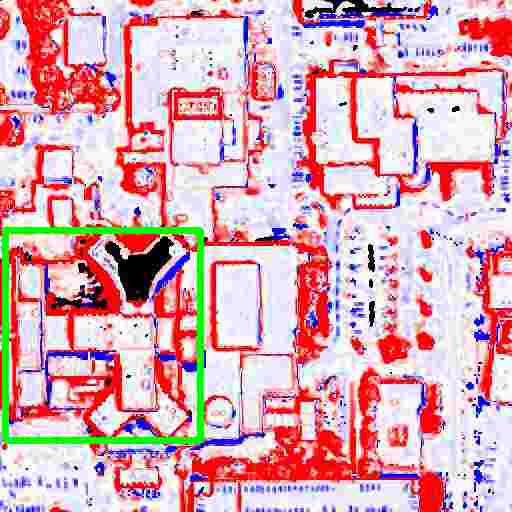}\vspace{2pt}
\includegraphics[width=1\linewidth]{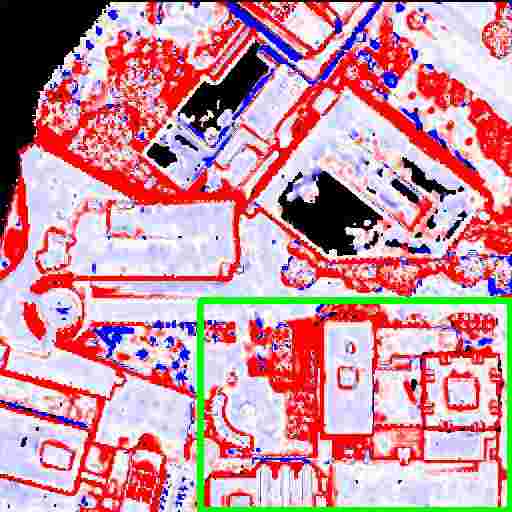}\vspace{2pt}
\includegraphics[width=1\linewidth]{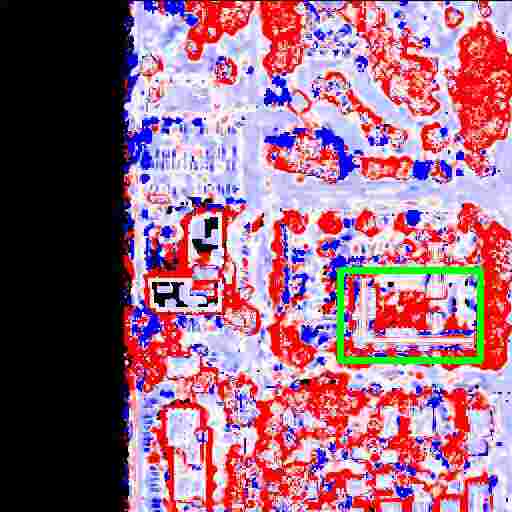}\vspace{2pt}
\end{minipage}}

\subfigure{
\begin{minipage}[b]{0.2\textwidth}
\includegraphics[width=1\linewidth]{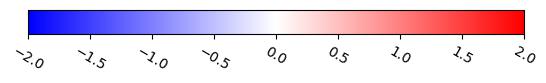}
\end{minipage}}

\caption{Error maps of on the DFC dataset. The color bar is expressed in units of meters.}
\label{fig:DFC_errmap}
\end{figure}

Table \ref{tab:DFC} shows that S2P achieved a higher accuracy. However, compared to our method, the S2P reconstruction failed in more areas, and the completeness at a threshold of 1 m was worse, causing a maximum drop of approximately 40\% and a minimum drop of approximately 4\%. By combining the DSM reconstruction results shown in Fig. \ref{fig:DFC_benchmark} and the error maps presented in Fig. \ref{fig:DFC_errmap}, our method performed exceptionally well for vegetation, roads, and the edges of buildings. Fig. \ref{fig:DFC_errmap} displays the error maps of the DSM against the true DSM. Our method exhibited superior accuracy in estimating the heights of buildings and roads, but its accuracy was relatively low for vegetation and building shadows. The DSMs reconstructed by LPS exhibited a lower accuracy overall, possibly because the LPS method’s 3D reconstruction of multi-date VHR imagery without GCPs is poor.

\subsubsection{Results for the ISPRS-ZY3 dataset}

\begin{figure}[htbp]
\centering
\subfigure[The true color image \cite{ISPRS_zy}]{
\includegraphics[width=7cm]{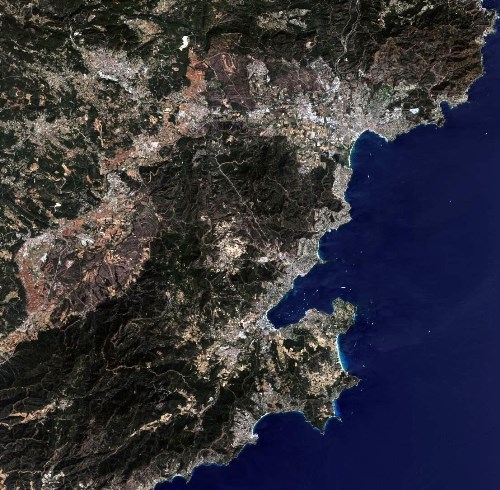}
}
\quad
\subfigure[DSM]{
\includegraphics[width=7.5cm]{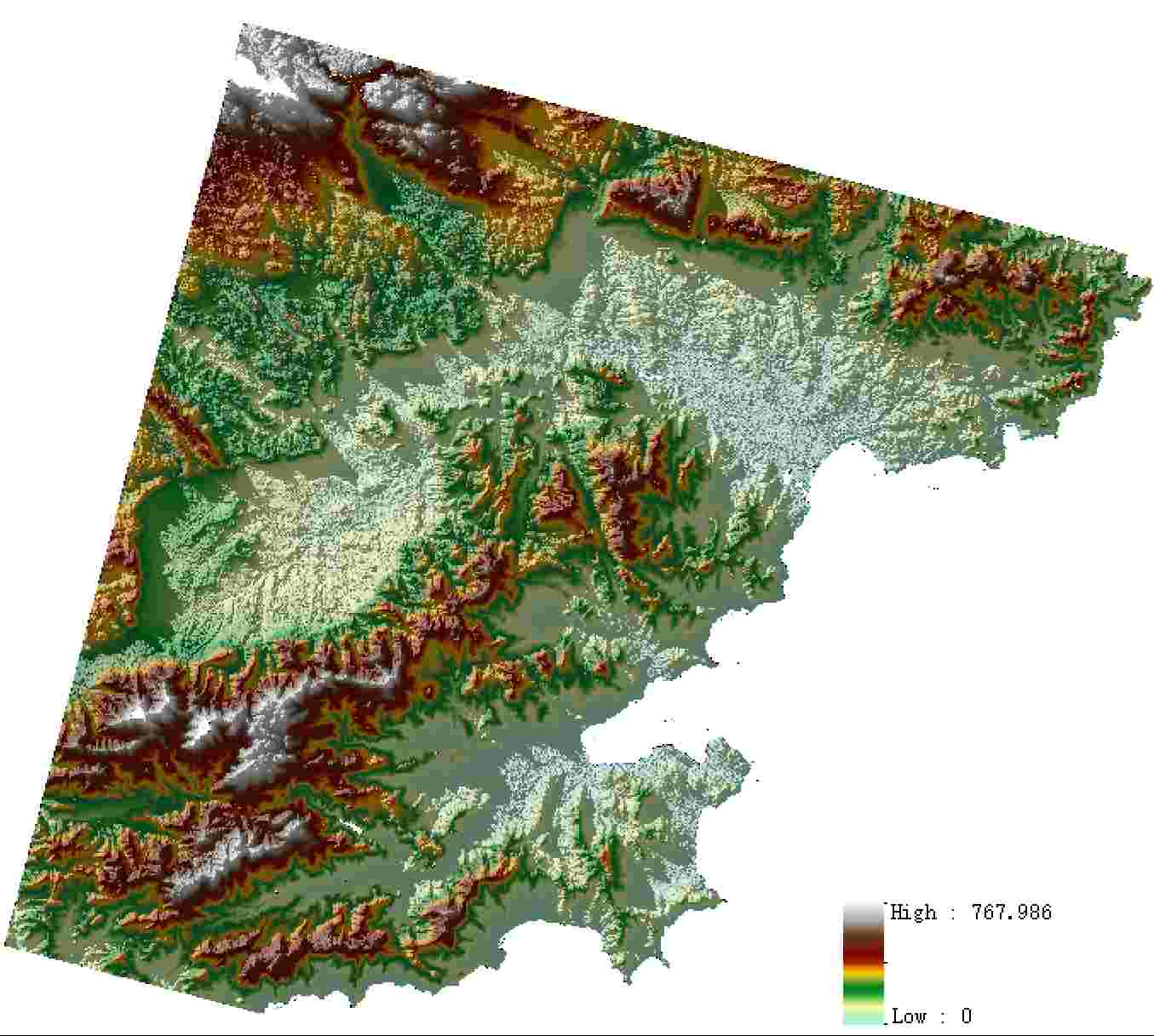}
}
\quad

\caption{Results produced by our REPM pipeline for the Sainte-Maxime dataset: (a) true color image and (b) DSM product.}
\label{fig:ZY3}
\end{figure}

Because the data for the Sainte-Maxime region in the ISPRS ZY3 satellite imagery contained GCPs, we assessed the absolute positioning accuracy of these data. We first calculated the altitude accuracy of the stereo positioning of the checkpoints for the RPC model’s image space compensation scheme, as shown in Table \ref{tab:ZY3}. We then examined the corresponding altitudes of the GCPs in the reconstructed DSM and calculated the median, RMSE, and maximum of the altitude errors. For this large-scale satellite image, the Adapted COLMAP generated NA. Table \ref{tab:ZY3} demonstrates that the addition of the image refinement model dramatically improved the precision of our method, which exhibited an accuracy comparable to S2P. In addition, our method attained optimality regarding both the median and maximum errors. The final results of our REPM pipeline are presented in Fig. \ref{fig:ZY3} (water is filtered out in the pipeline).

\begin{table}[htbp]
\footnotesize
  \caption{\label{tab:ZY3} Absolute accuracy comparison of reconstruction outcomes for Sainte-Maxime, France in the ISPRS-ZY3 image data is undertaken employing GCP. (Bolded is the best, underlined is the second best.)}
  \centering
  \begin{tabularx}{0.5\textwidth}{lXXX}
    \toprule
    \textbf{Methods}	&\textbf{\makecell{ME\\(m)\textcolor{red}{$\downarrow$}}} &\textbf{\makecell{RMSE\\(m)\textcolor{red}{$\downarrow$}}} &\textbf{\makecell{Max\\(m)\textcolor{red}{$\downarrow$}}}  \\
    \midrule
    RPC model localization & \textbf{2.668}& 4.053 &8.508	\\
    S2P& 3.464 & \textbf{3.679} & \underline{6.308} \\
    LPS &6.779 & 7.629 &13.276	 \\
    REPM &12.122 & 12.353 & 18.960	 \\
    REPM+Ref. &\underline{3.442} & \underline{3.812} &\textbf{5.862} \\
    \bottomrule
  \end{tabularx}
\end{table}

\subsubsection{Results for the GF7 dataset}

\begin{figure}[!htbp]
\centering
\subfigure[Reference]{
\begin{minipage}[b]{0.16\textwidth}
\includegraphics[width=1\linewidth]{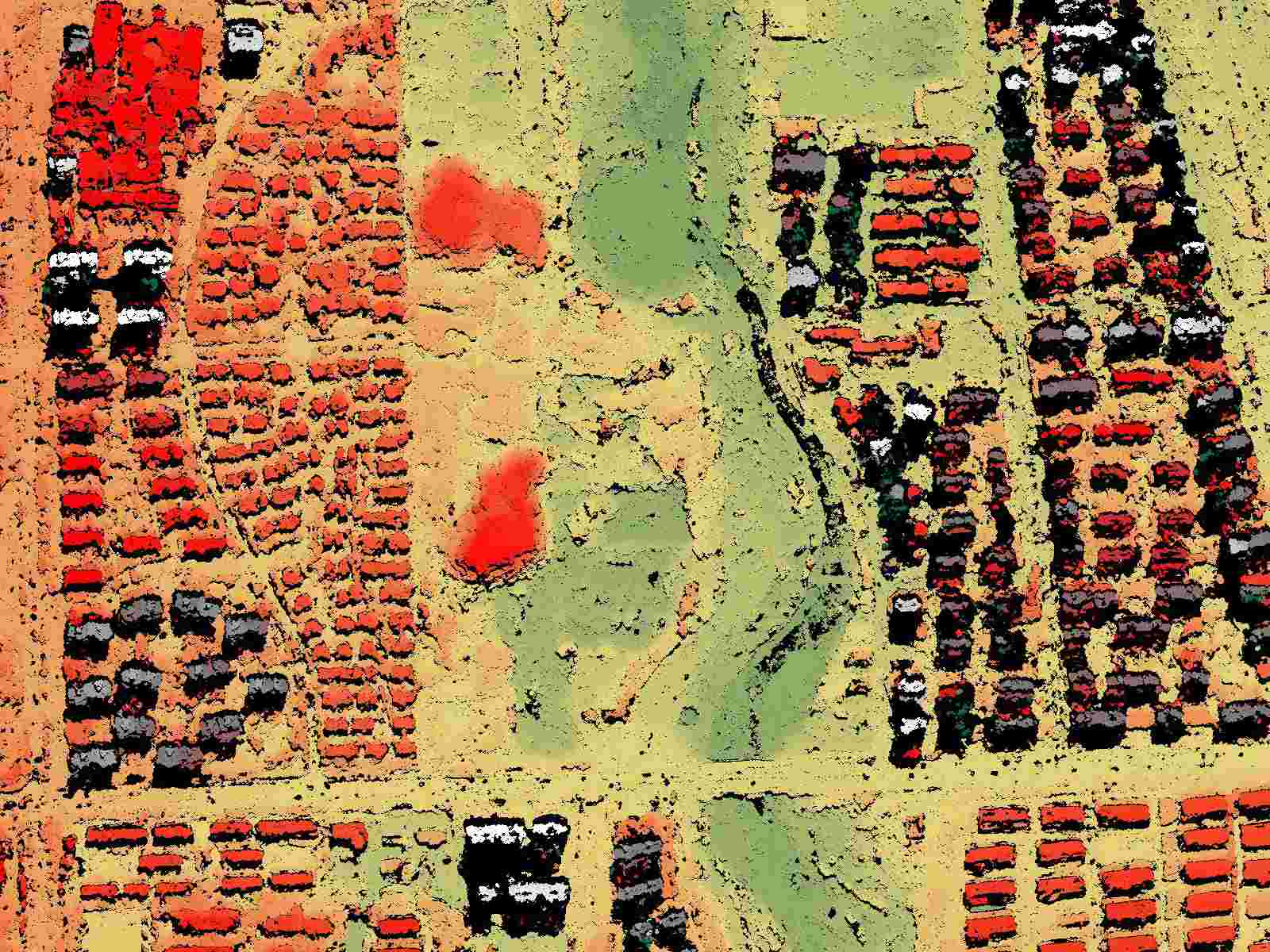}\vspace{4pt}
\end{minipage}
\begin{minipage}[b]{0.15\textwidth}
\includegraphics[width=1\linewidth]{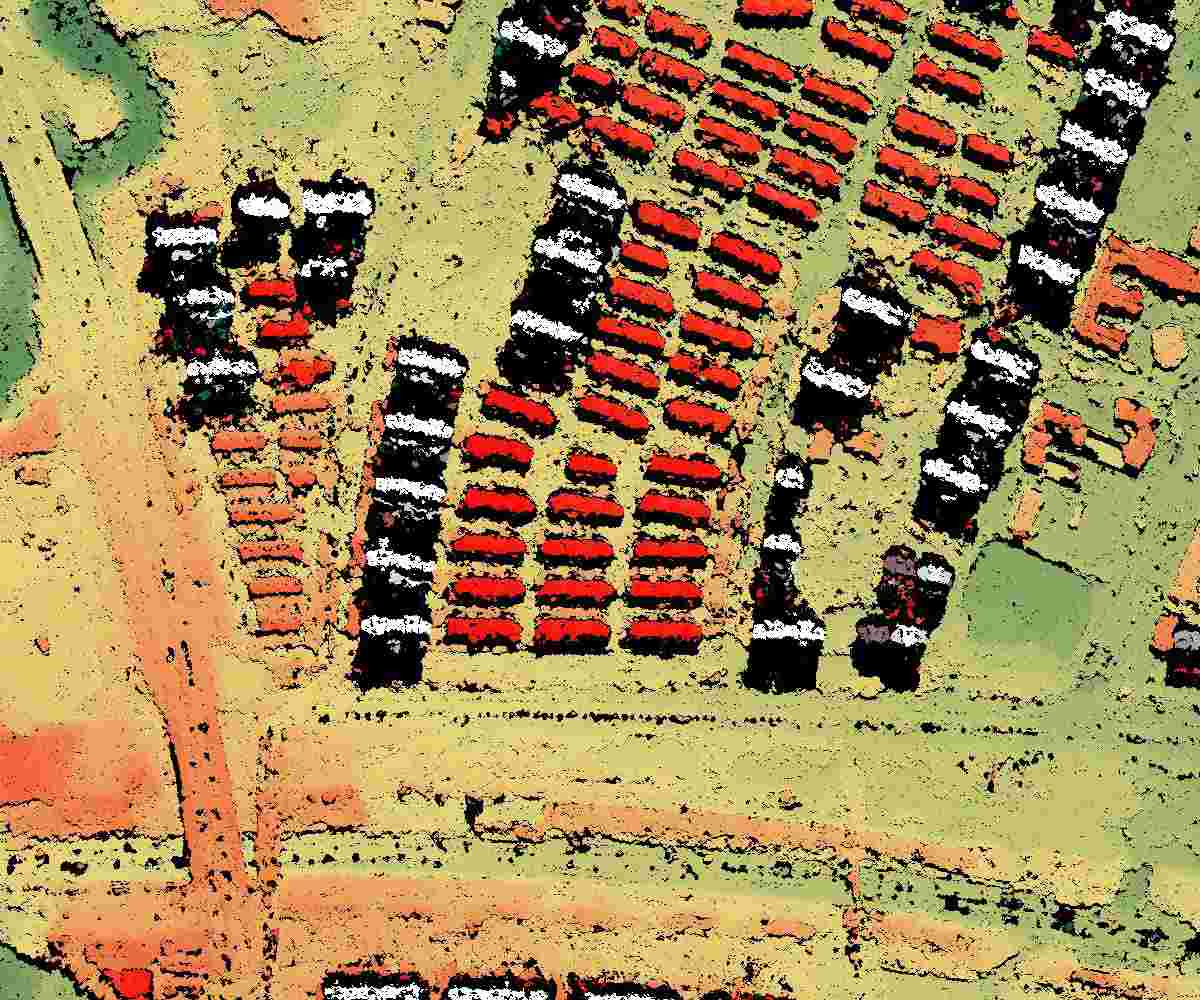}\vspace{4pt}
\end{minipage}
}

\subfigure[S2P]{
\begin{minipage}[b]{0.118\textwidth}
\includegraphics[width=1\linewidth]{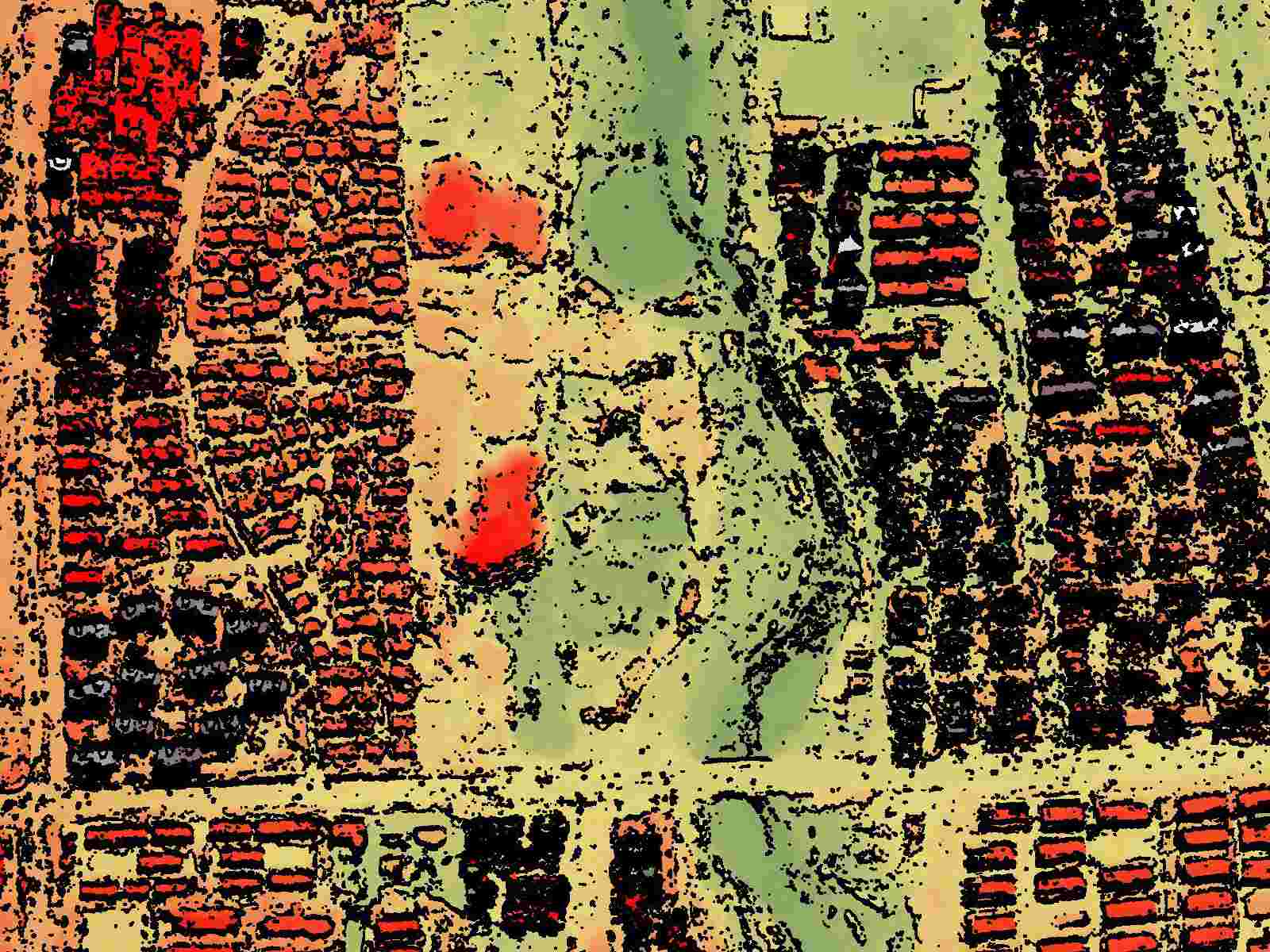}\vspace{1pt}
\includegraphics[width=1\linewidth]{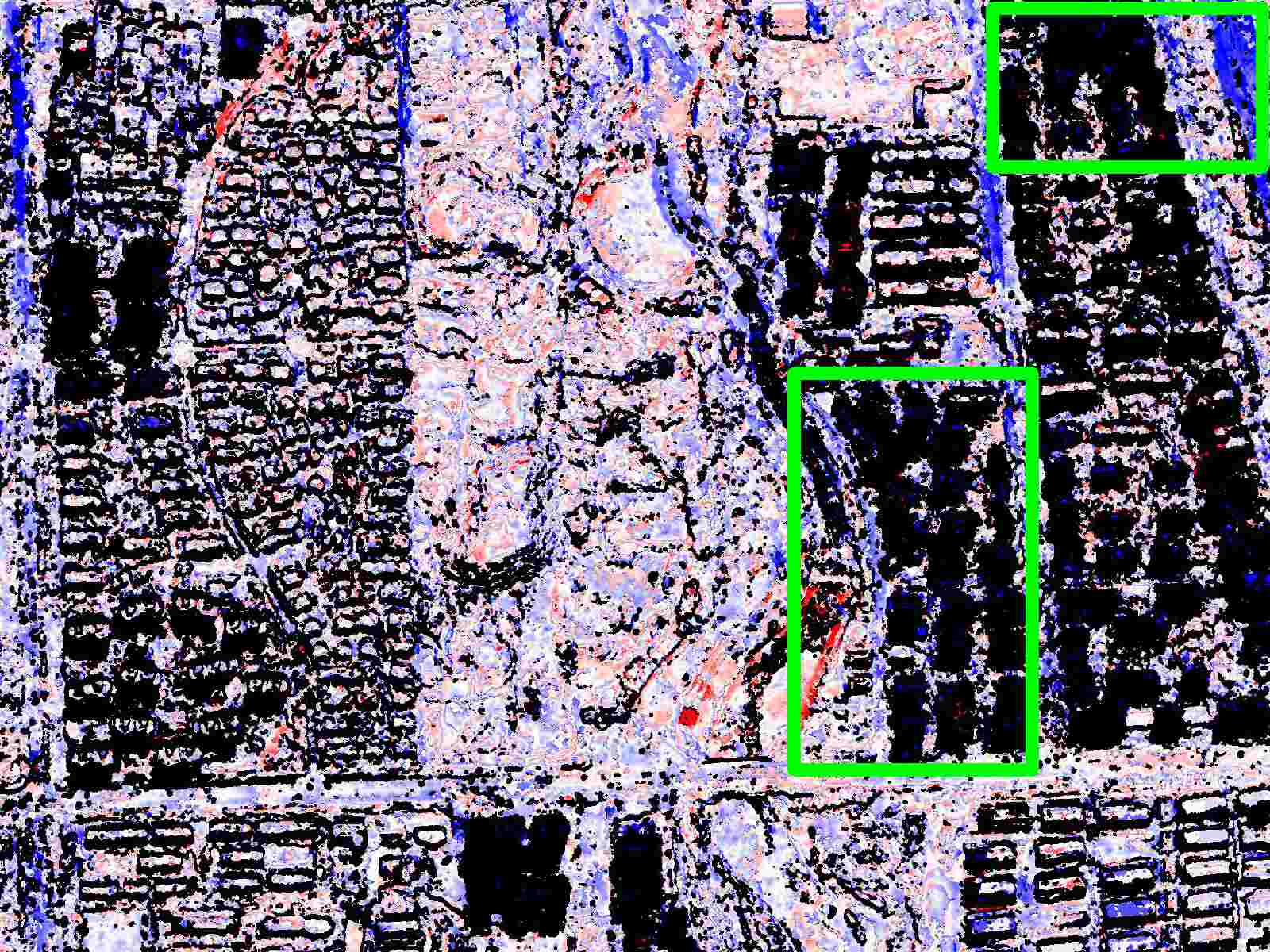}\vspace{1pt}
\includegraphics[width=1\linewidth]{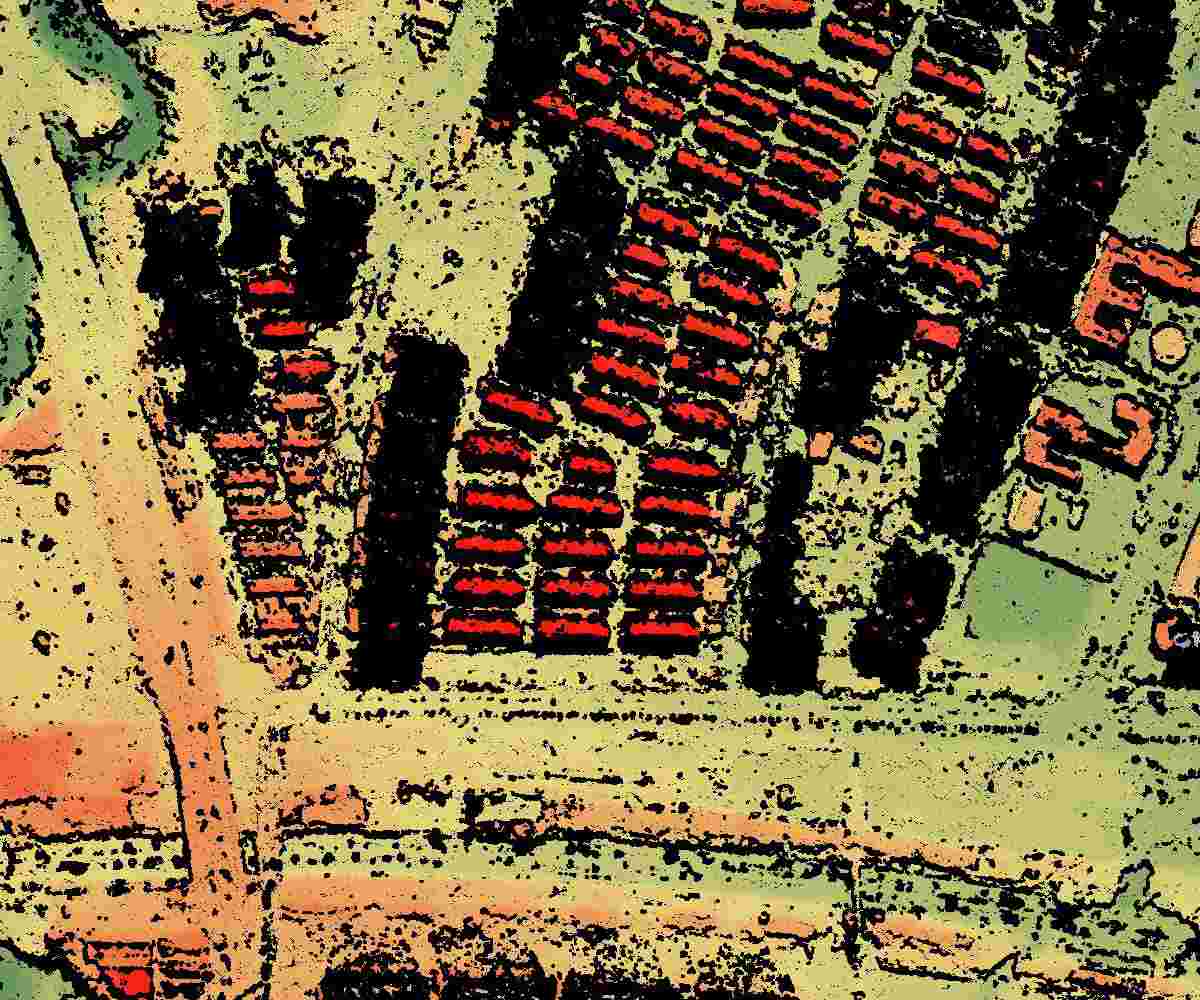}\vspace{1pt}
\includegraphics[width=1\linewidth]{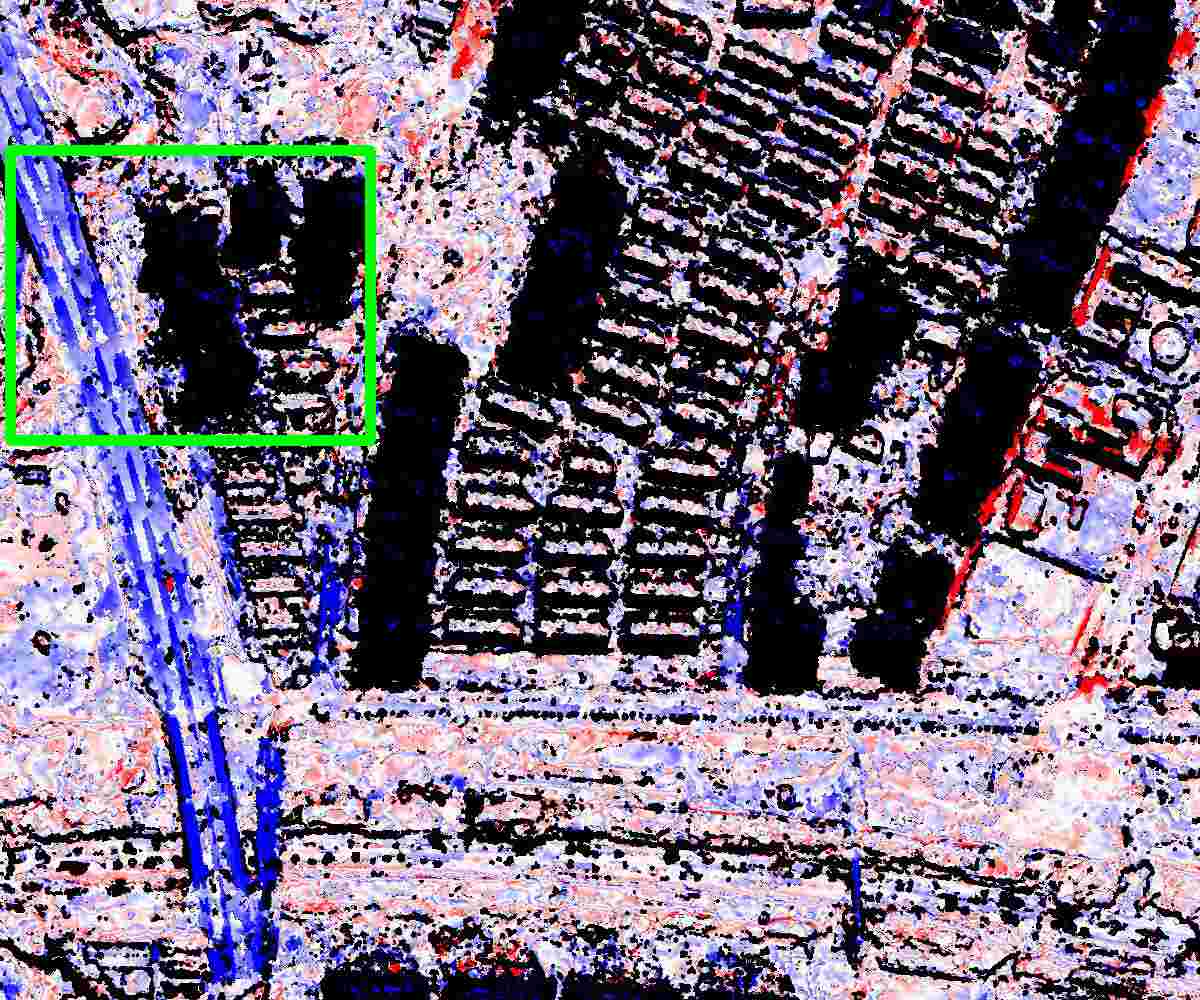}\vspace{1pt}
\end{minipage}} \hspace{-3mm}
\subfigure[LPS]{
\begin{minipage}[b]{0.118\textwidth}
\includegraphics[width=1\linewidth]{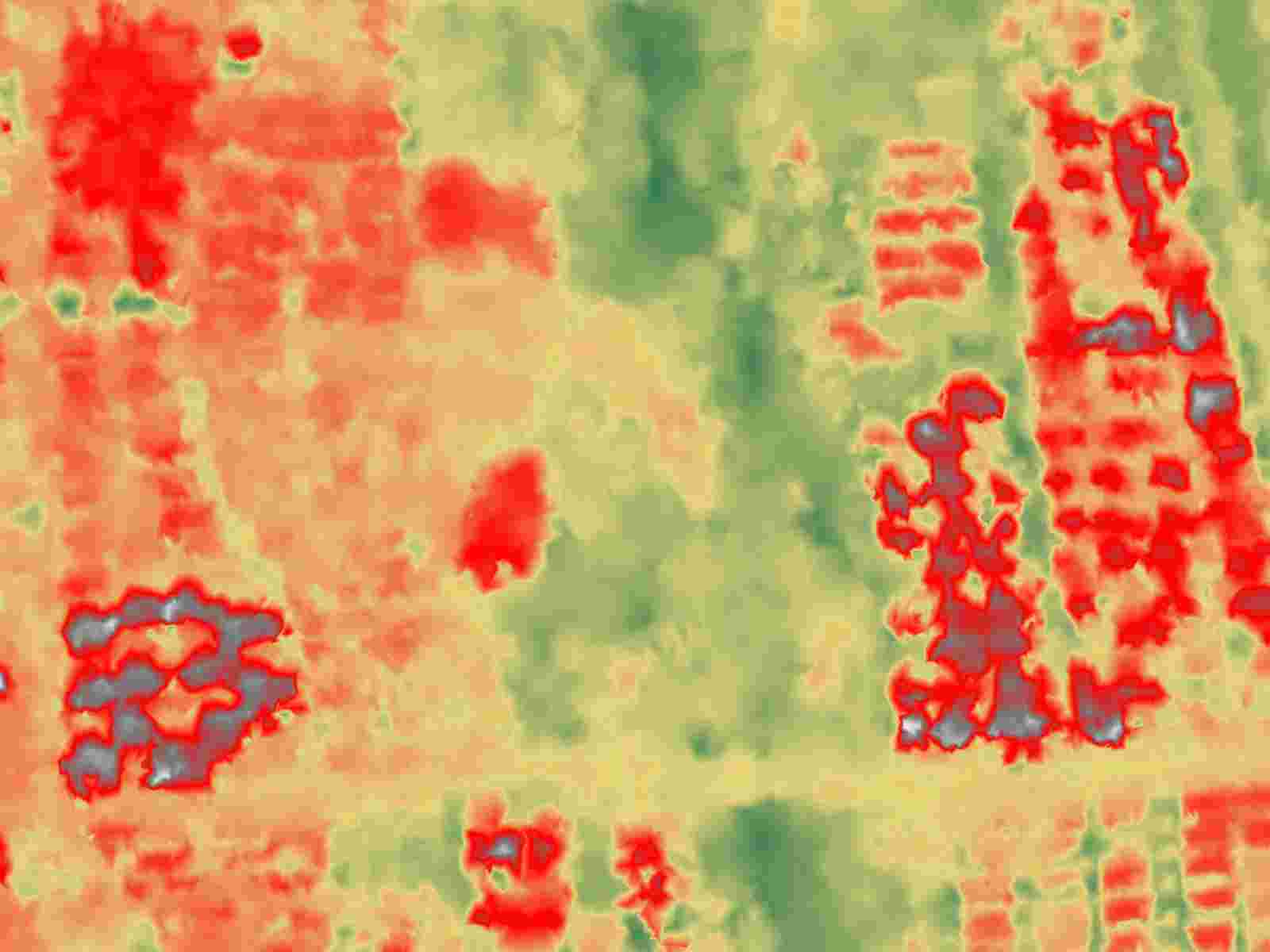}\vspace{1pt}
\includegraphics[width=1\linewidth]{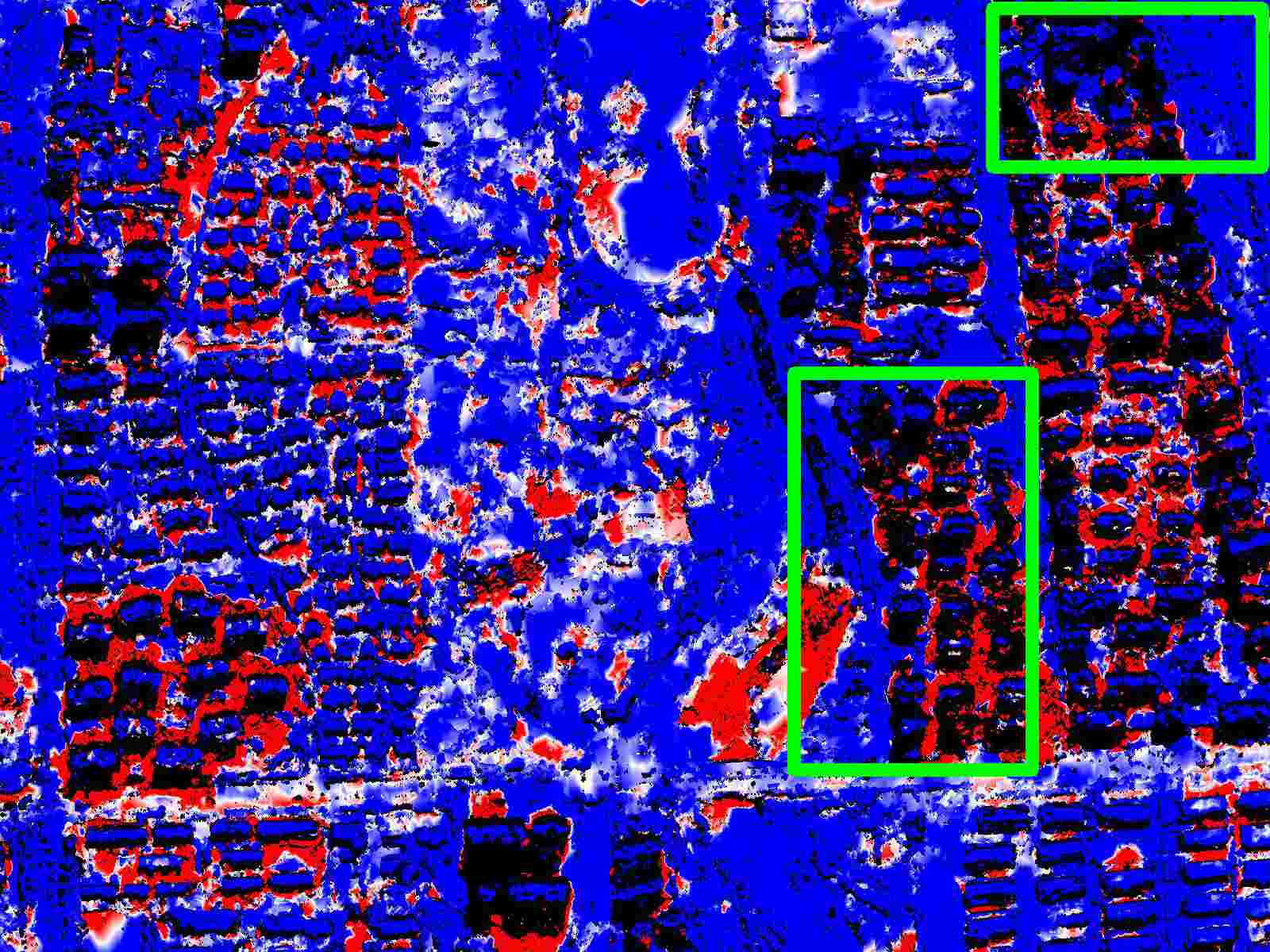}\vspace{1pt}
\includegraphics[width=1\linewidth]{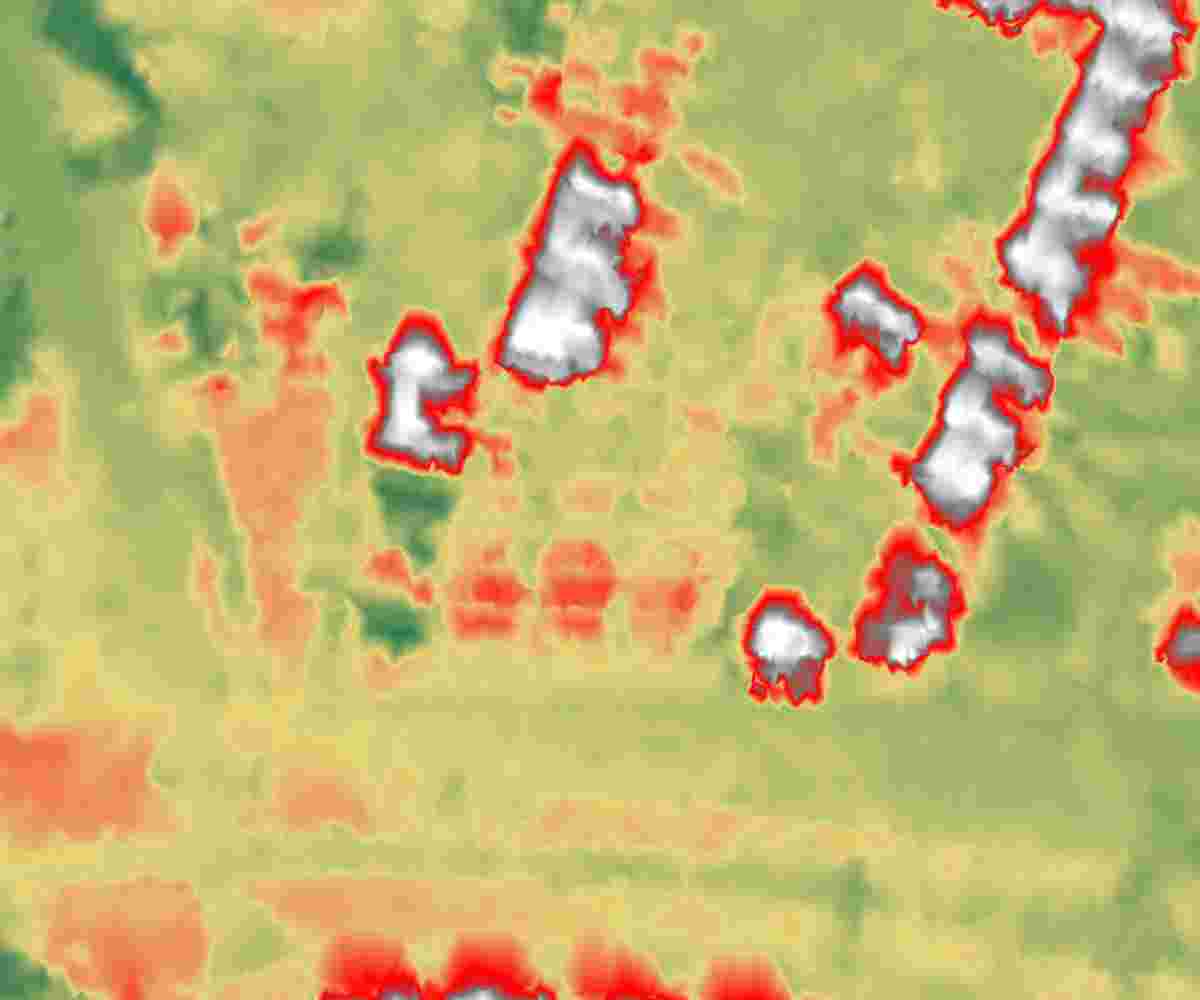}\vspace{1pt}
\includegraphics[width=1\linewidth]{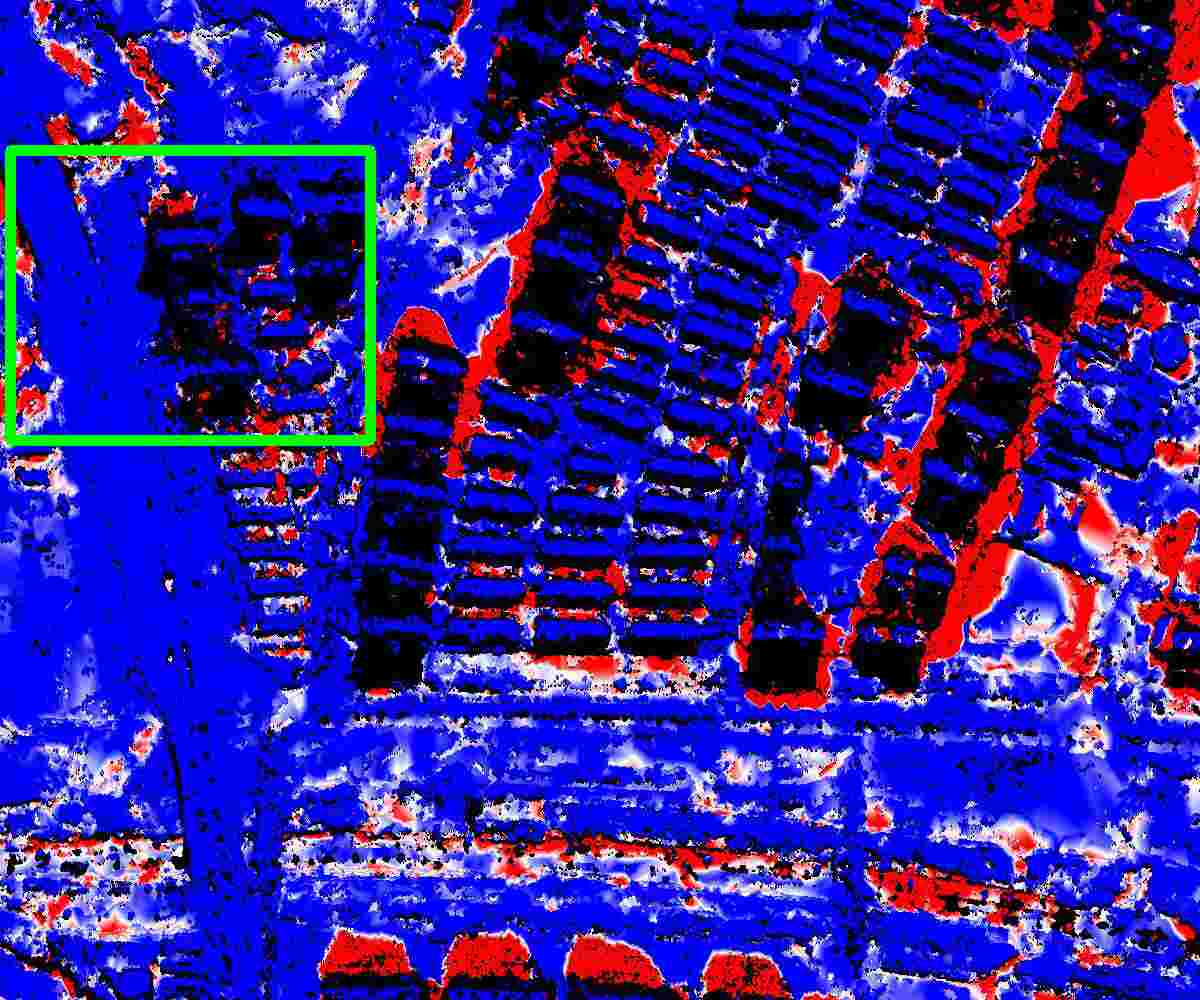}\vspace{1pt}
\end{minipage}}\hspace{-2mm}
\subfigure[REPM]{
\begin{minipage}[b]{0.118\textwidth}
\includegraphics[width=1\linewidth]{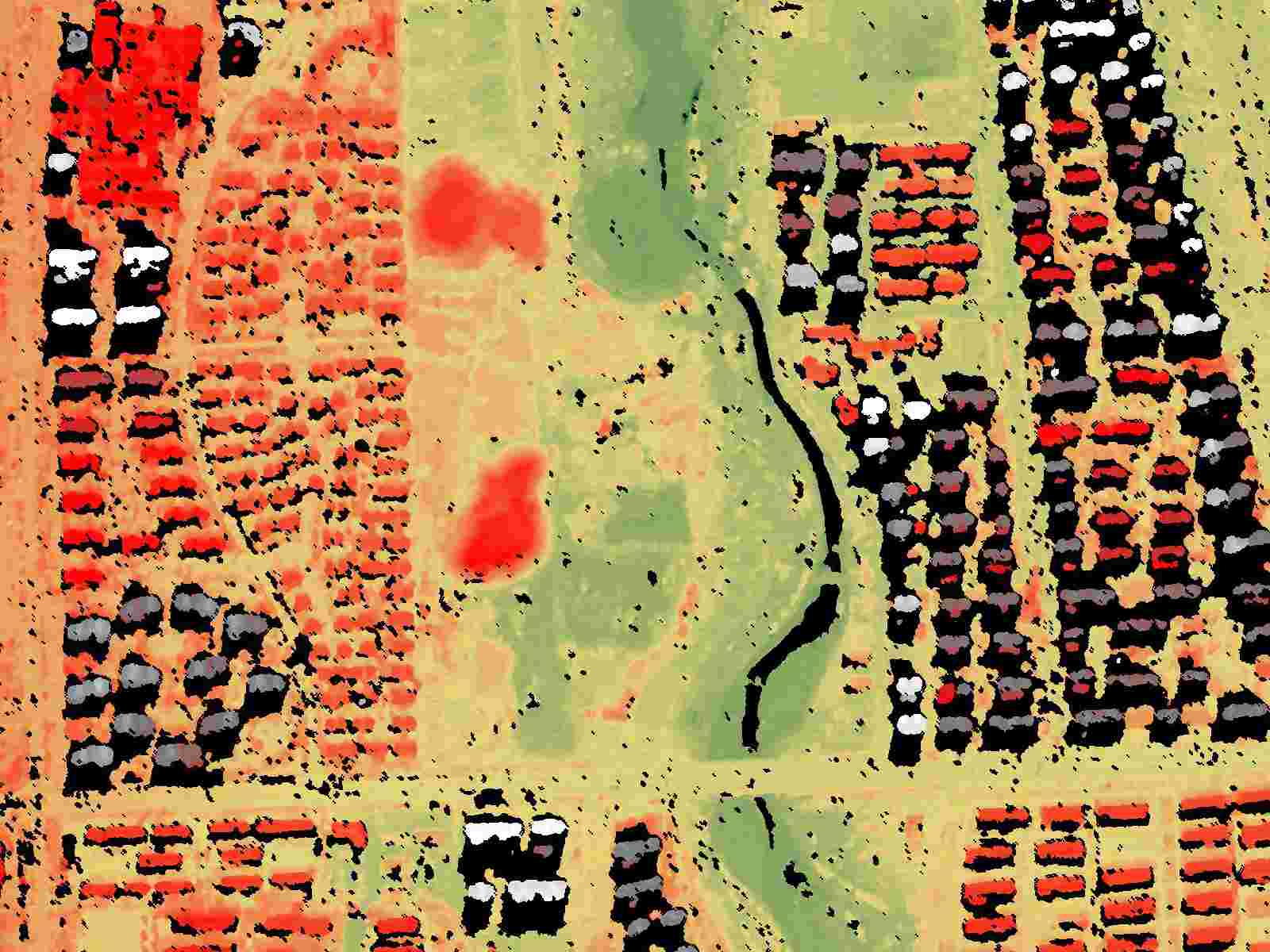}\vspace{1pt}
\includegraphics[width=1\linewidth]{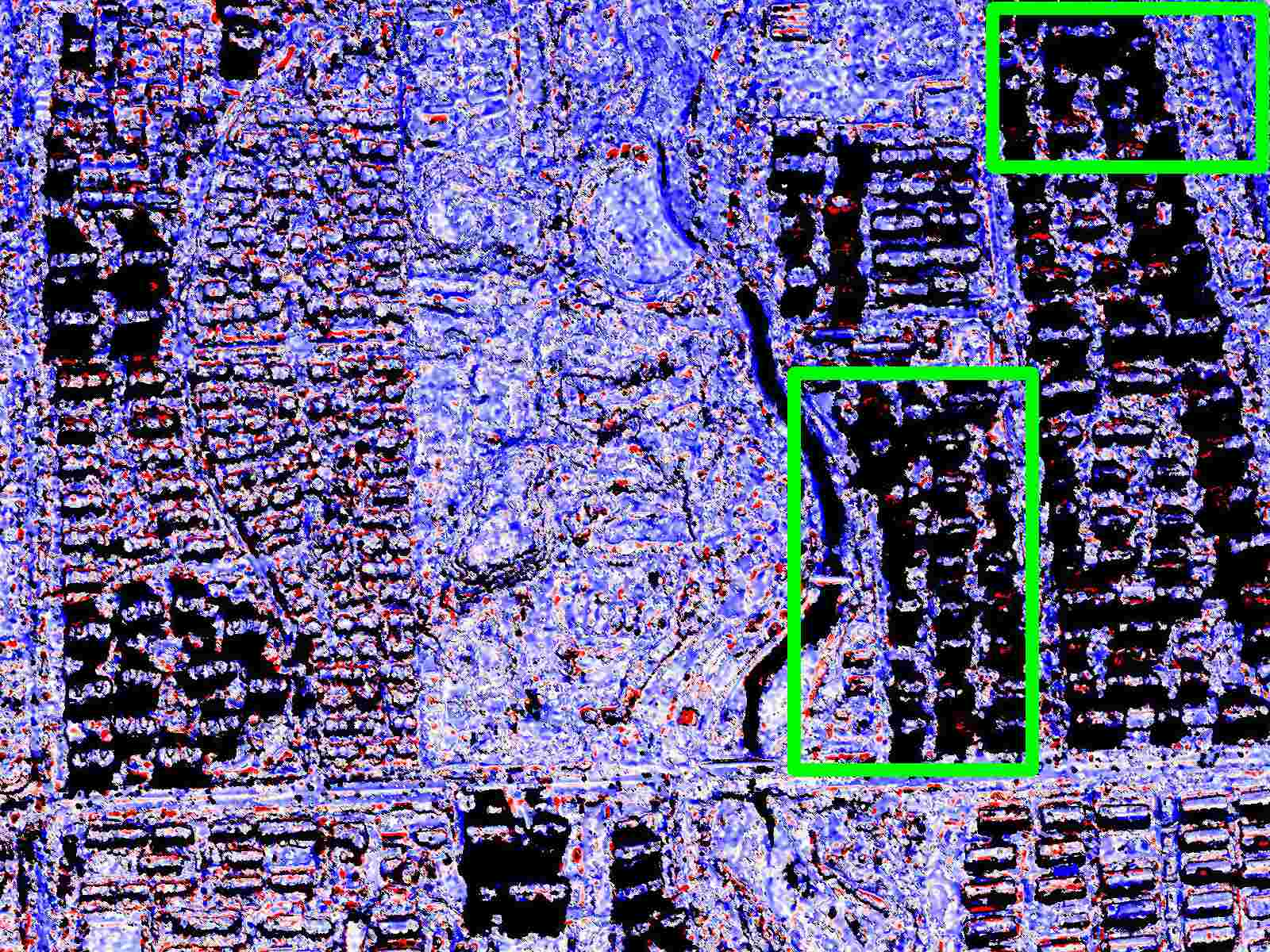}\vspace{1pt}
\includegraphics[width=1\linewidth]{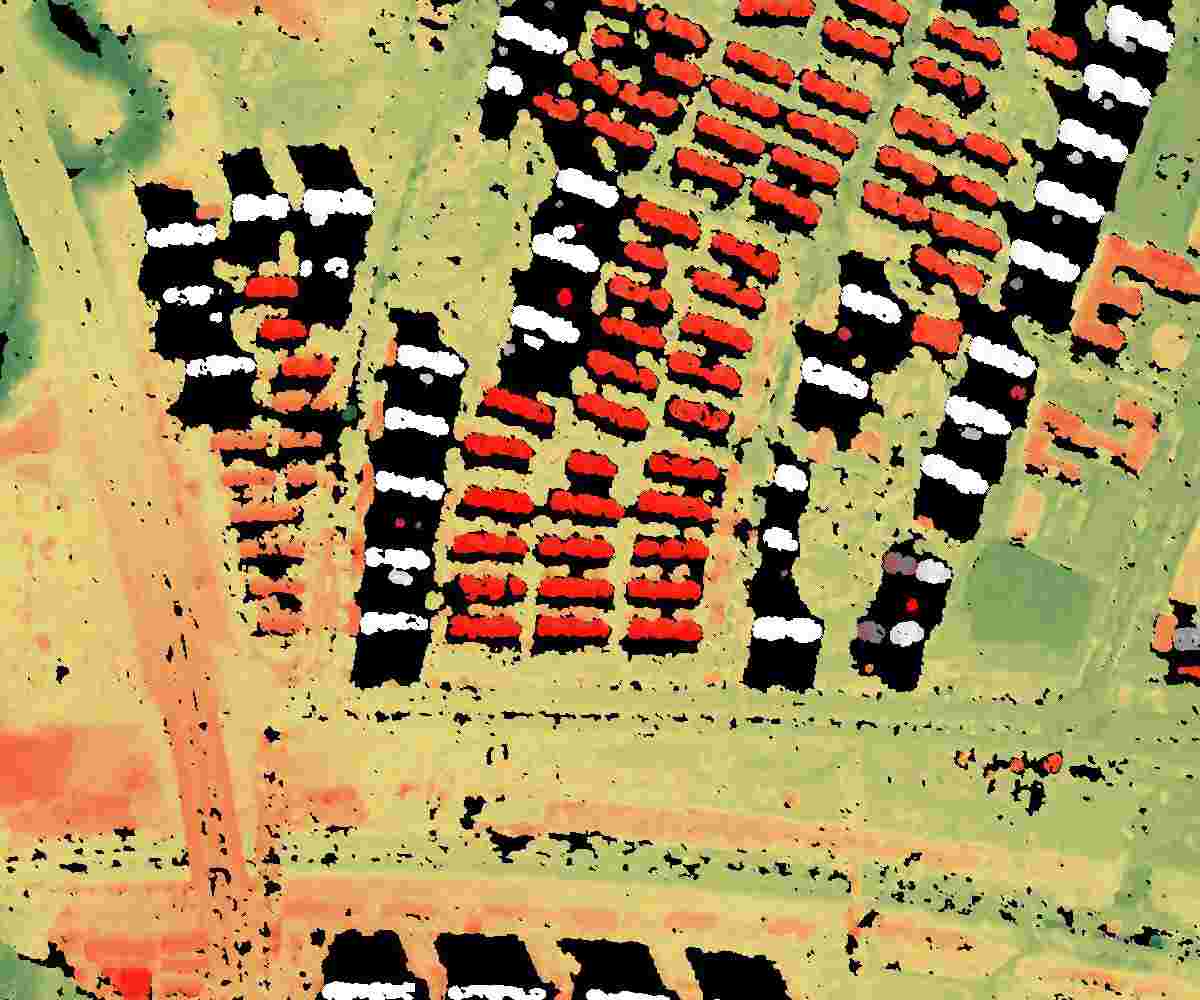}\vspace{1pt}
\includegraphics[width=1\linewidth]{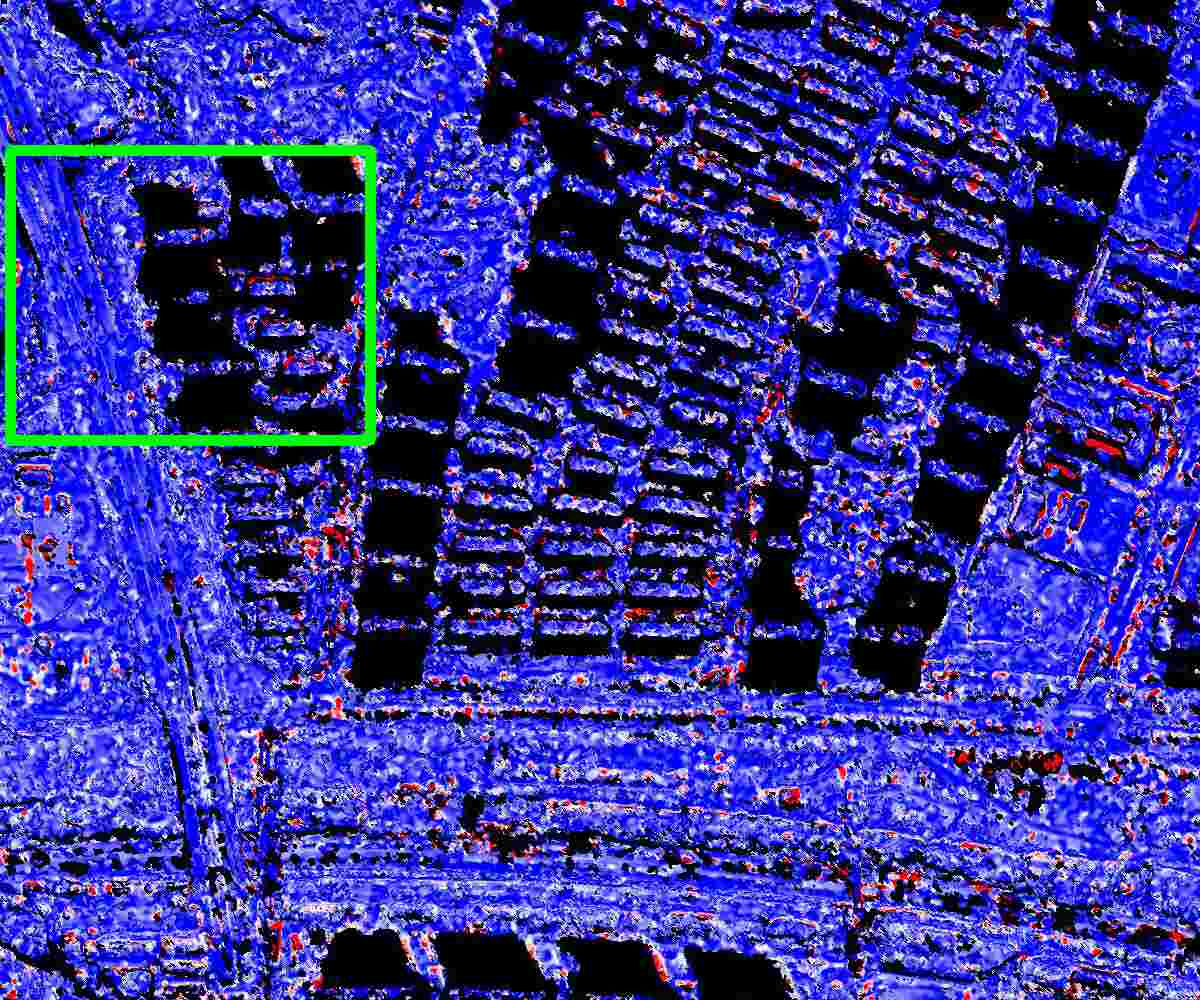}\vspace{1pt}
\end{minipage}}\hspace{-2mm}
\subfigure[REPM+Cor.]{
\begin{minipage}[b]{0.118\textwidth}
\includegraphics[width=1\linewidth]{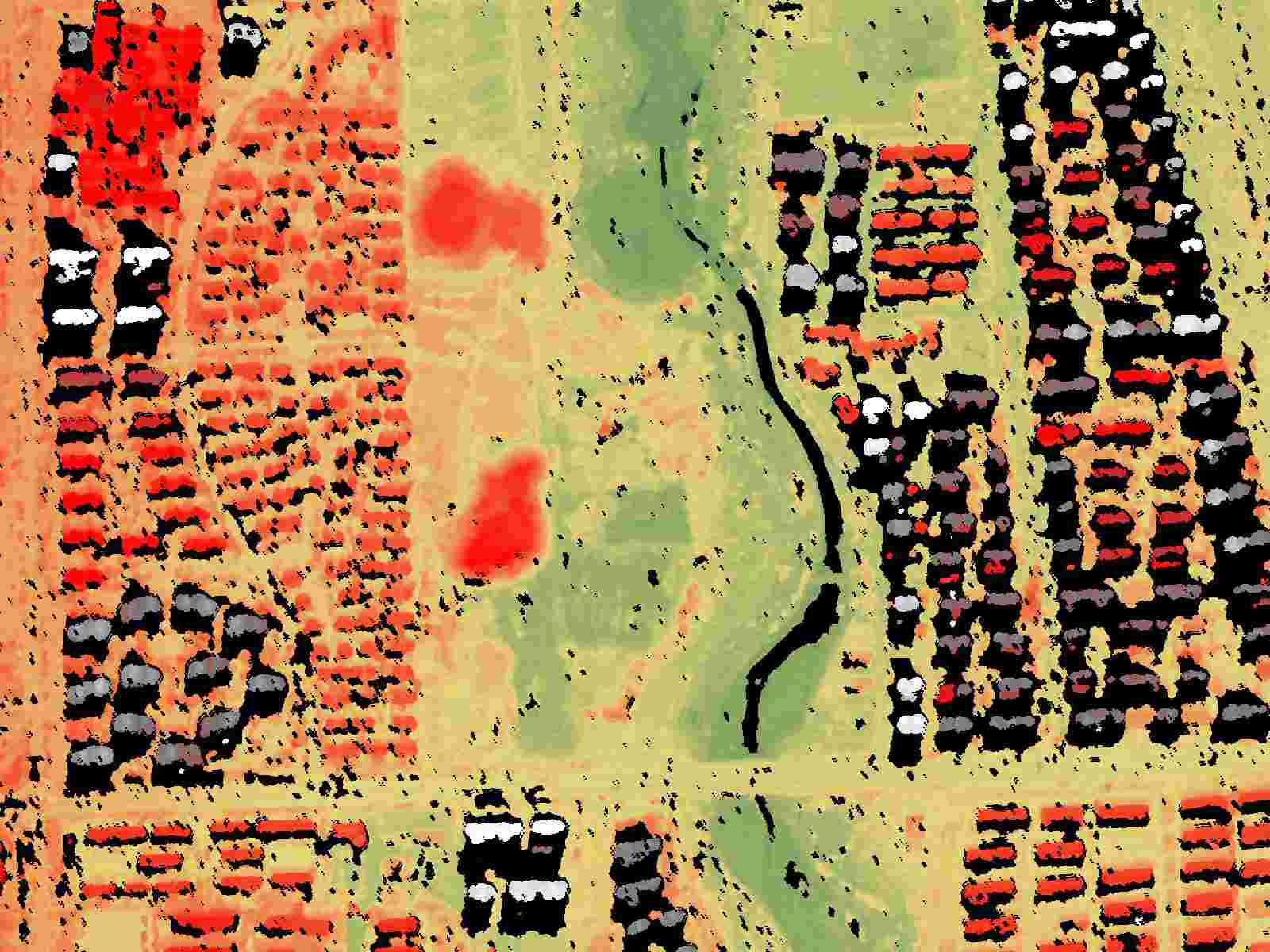}\vspace{1pt}
\includegraphics[width=1\linewidth]{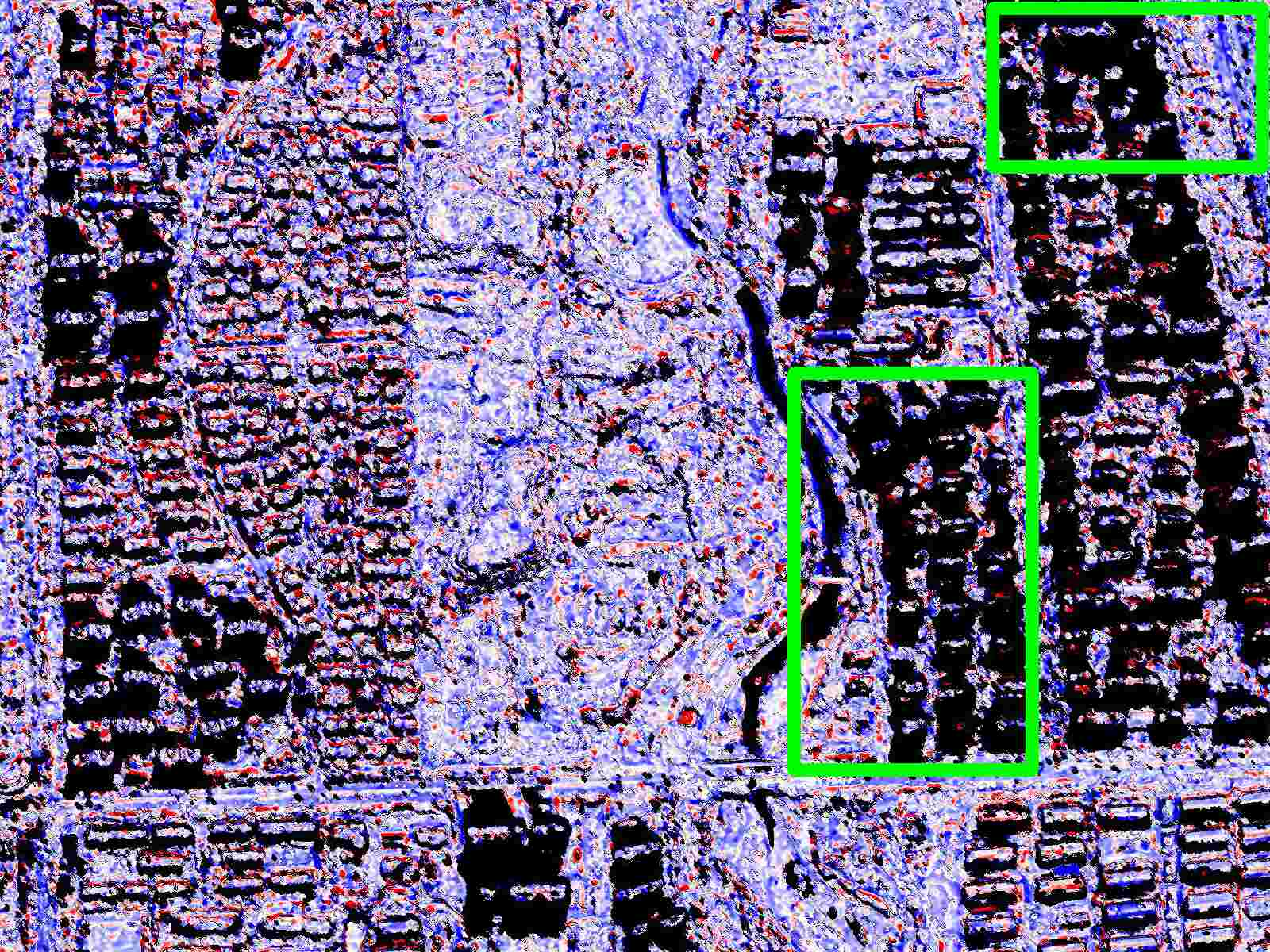}\vspace{1pt}
\includegraphics[width=1\linewidth]{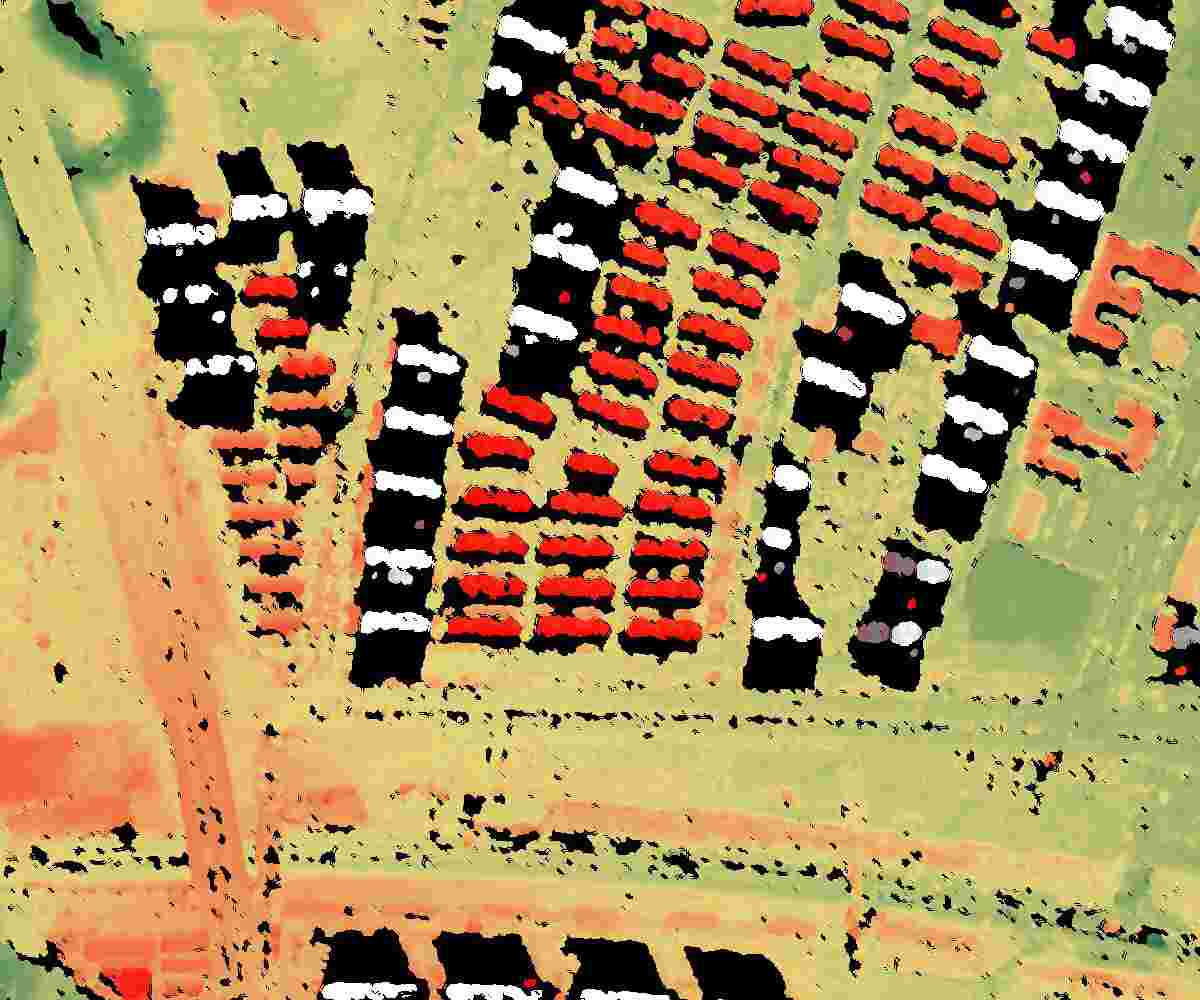}\vspace{1pt}
\includegraphics[width=1\linewidth]{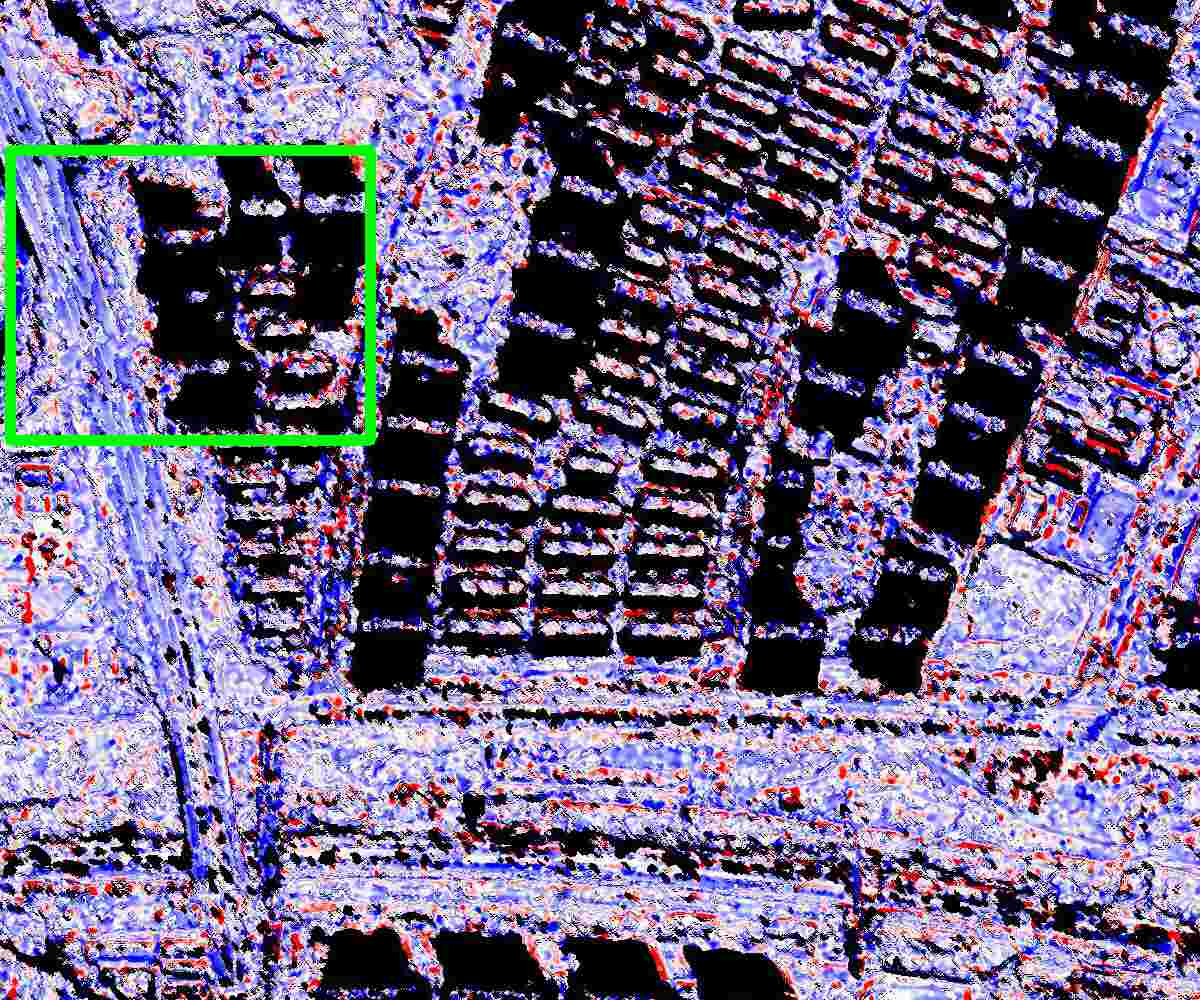}\vspace{1pt}
\end{minipage}}

\subfigure{
\begin{minipage}[b]{0.15\textwidth}
\includegraphics[width=1\linewidth]{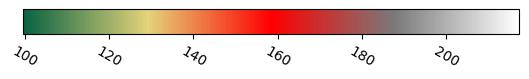}\vspace{4pt}
\end{minipage}
\begin{minipage}[b]{0.15\textwidth}
\includegraphics[width=1\linewidth]{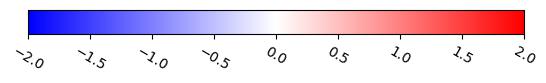}\vspace{4pt}
\end{minipage}}

\caption{Partial DSM results and error plots for the GF7 image data. The color bar is expressed in units of meters.}
\label{fig:GF_benchmark}
\end{figure}

In the experiments with the Gaofen7 (GF7) image of the Zhengzhou region, we used the reference image as a benchmark to calculate the accuracy and completeness of the reconstructed DSM. The quantitative results are presented in Table \ref{tab:GF7}, and Fig. \ref{fig:GF_benchmark} displays the DSM reconstruction results and error maps for partial areas.

\begin{table}[htbp]
\footnotesize
  \caption{\label{tab:GF7}  Quantitative results of DSM reconstruction quality on the GF7 image data. (Bold works best)}
  \centering
  \begin{tabularx}{0.5\textwidth}{lXXXX}
    \toprule
    \textbf{Methods}	& \textbf{\makecell{ME\\(m)\textcolor{red}{$\downarrow$}}}	& \textbf{\makecell{MAE\\(m)\textcolor{red}{$\downarrow$}}}&
    \textbf{\makecell{RMSE\\(m)\textcolor{red}{$\downarrow$}}}	&\textbf{\makecell{Comp$_2$\\(\%)\textcolor{red}{$\uparrow$}}}    \\
    \midrule
    S2P&   \textbf{0.725}  &   \textbf{0.398}   &3.492 &74.810
    \\
    LPS	&2.912	&5.630	&12.037	&32.401
    \\
    REPM &1.280	&0.973 &3.303	& 74.917\\
    REPM+Ref.&	0.939	&0.622	&\textbf{2.804} & \textbf{77.367}\\
    \bottomrule
  \end{tabularx}
\end{table}

As shown in Table \ref{tab:GF7}, our method achieved the highest reconstruction accuracy and completeness, and the incorporation of the image refinement model significantly boosted the accuracy and completeness with a threshold of 2 m. Fig. \ref{fig:GF_benchmark} shows the DSM reconstruction results for local areas. For high-resolution images of urban areas, we mainly compared the reconstruction results of buildings and roads. Our method was optimal in terms of the completeness and accuracy for buildings. In contrast, S2P was unsuccessful in reconstructing tall buildings (white buildings). Furthermore, LPS lost detailed building information, and the overall height estimation accuracy was poor.

% \begin{figure*}[!htbp]
% \centering
% \subfigure[Reference]{
% \begin{minipage}[b]{0.5\textwidth}
% \includegraphics[width=1\linewidth]{figure/GF7/ref.png}\vspace{4pt}
% \end{minipage}}

% \subfigure[S2P]{
% \begin{minipage}[b]{0.48\textwidth}
% \includegraphics[width=1\linewidth]{figure/GF7/s2p.png}\vspace{4pt}
% \end{minipage}}
% \subfigure[LPS]{
% \begin{minipage}[b]{0.48\textwidth}
% \includegraphics[width=1\linewidth]{figure/GF7/erdas.png}\vspace{4pt}
% \end{minipage}}
% \subfigure[REPM@5120]{
% \begin{minipage}[b]{0.48\textwidth}
% \includegraphics[width=1\linewidth]{figure/GF7/epm-full.png}\vspace{4pt}
% \end{minipage}}
% \subfigure[REPM@5120+Cor.]{
% \begin{minipage}[b]{0.48\textwidth}
% \includegraphics[width=1\linewidth]{figure/GF7/cor-full.png}\vspace{4pt}
% \end{minipage}}

% \subfigure{
% \begin{minipage}[b]{0.25\textwidth}
% \includegraphics[width=1\linewidth]{figure/GF7/cor_dsm.cbar.jpg}\vspace{4pt}
% \end{minipage}
% \begin{minipage}[b]{0.25\textwidth}
% \includegraphics[width=1\linewidth]{figure/GF7/error_map.cbar.jpg}\vspace{4pt}
% \end{minipage}
% }

% \caption{DSM results of on the GF7 image data. The units of color bar are meters.}
% \label{fig:GF_benchmark}
% \end{figure*}

\section{Discussion}

\subsection{Summary of the reconstruction results}

In the DSM reconstruction experiments on the four datasets, we can draw the following conclusions.

(1) The proposed REPM pipeline significantly outperforms all the other methods, including both the software and open-source solutions, in terms of the two accuracy metrics of RMSE and Comp. When comparing the reconstructed DSMs and error maps, it can be observed that the pipeline proposed in this paper has yielded the highest accuracy and completeness.

(2) The incorporation of a polynomial image refinement model has proven to further enhance the accuracy and completeness of the DSM reconstruction. Additionally, it is worth noting that for imaging on a large-scale, the DSM reconstruction accuracy and completeness is significantly improved.

\subsection{Influence of the equivalent pinhole model}

%In this section, we discuss the experiments pertaining to the equivalent errors. Specifically, we investigated the influence of image size on the equivalent error within the equivalent model, as well as the impact of the correction model on the equivalent error.

%\subsubsection{Error experiment for the equivalent model}

According to the equivalent error formula presented in Section \ref{sec:3.2}, the factors that affect the equivalent error are directly proportional to the size of the image and the difference in terrain altitude. In the experiment, we primarily explored the effect of the image size on the equivalent error. We defined the equivalent error $L_{E}$ as the distance between the position of pixel $p$ computed using the RPC parameter and the position of pixel $p'$ computed using the KRT parameter for the same object point, as displayed in Eq. (\ref{eq:equerr}). We used Eq. (\ref{eq:RMSE}) to calculate the RMSE accuracy of the equivalent error. 

\begin{equation}
 RMSE = \sqrt{(Samp\ RMSE)^2+(Line\ RMSE)^2}
  \label{eq:equerr}
\end{equation}
 
\begin{equation}
  L_{E} = \sqrt{(x_{p}-x_{p'})^2+(y_{p}-y_{p'})^2} 
  \label{eq:equerr}
\end{equation}

For satellite CCD images within a specific dataset, the error is directly proportional to the image size. For the same image size, as the image resolution increases, the equivalent error increases. The impact of image resolution on the equivalent error can be viewed as follows: as the image resolution increases, the terrain area corresponding to the same image size decreases, which indirectly affects the altitude difference and decreases it. According to Eq. (\ref{eq:equerr}), the equivalent error is proportional to the altitude difference. In summary, the image resolution indirectly affects the equivalent error for the same image size; as the image resolution increases, the equivalent error increases.

\begin{figure}[htbp]
\centering
\subfigure{
\begin{minipage}[b]{0.4\textwidth}
\includegraphics[width=1\linewidth]{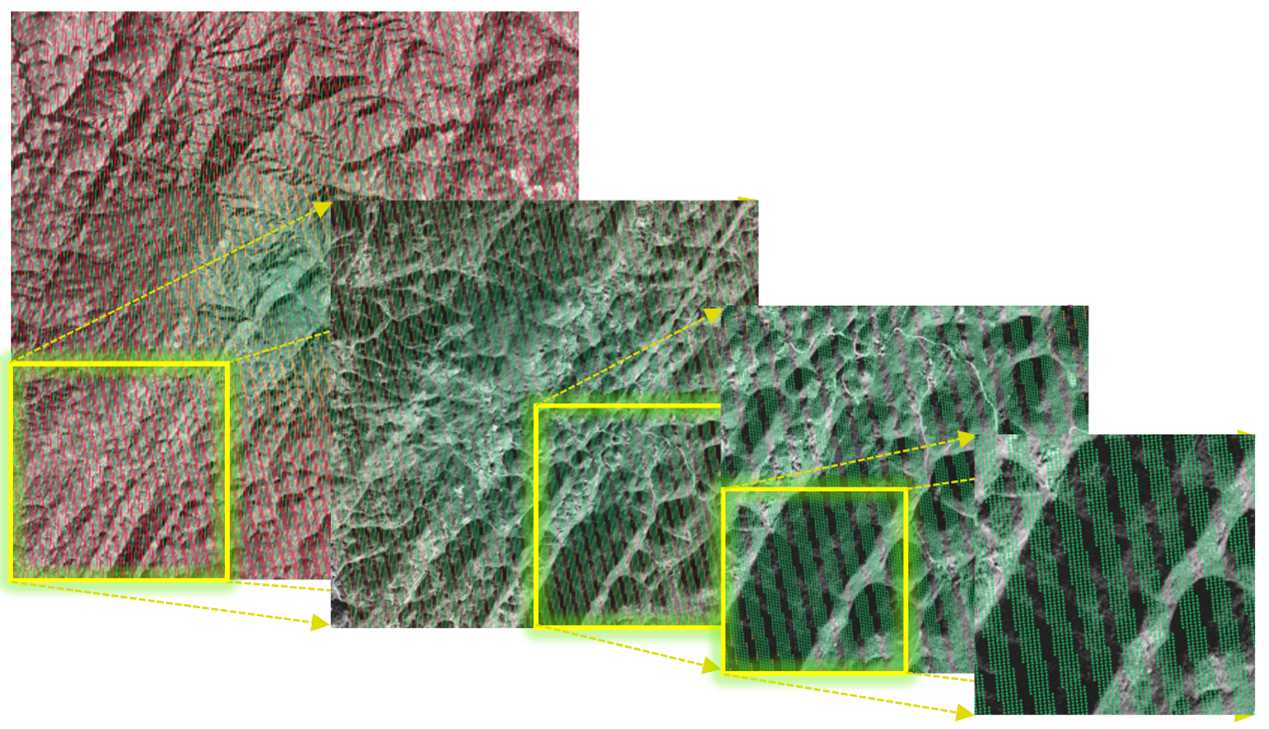}\vspace{4pt}
\end{minipage}}
\vspace{-2mm}
\subfigure{
\begin{minipage}[b]{0.2\textwidth}
\includegraphics[width=1\linewidth] {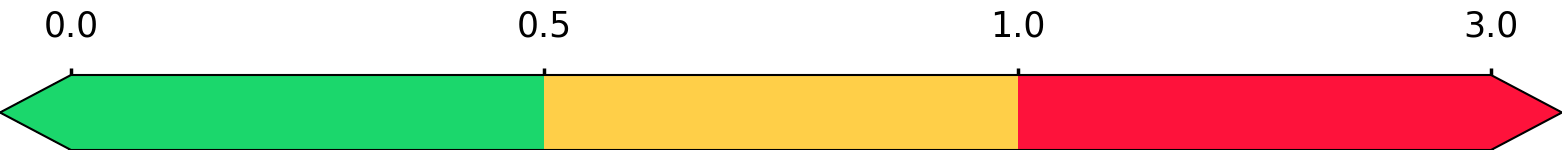}\vspace{4pt}
\end{minipage}}
\caption{Visualizations of the distribution of the equivalent error on the image and the relationship between the equivalent error and the image size. The color bar is expressed units of pixels.}
\label{fig:errordistribution}
\end{figure}

Additionally, Fig. \ref{fig:errordistribution} displays the distribution of pixel-level equivalent errors on the image, providing a visual representation of the impact of image size on the equivalent error. Evidently, the equivalent error decreases as the image size decreases, eventually resulting in subpixel precision.

%(a)Histogram of equivalent error statistics of different image sizes for different satellite images. (b)

% \begin{figure*}[h]
%   \begin{center}
%   \subfigure{
%     \centering
% 		\begin{minipage}[b]{0.5\linewidth}
% 			\centering	\includegraphics[width=1\linewidth]{figs/Figure_1.png}
% 		\end{minipage}
%         \begin{minipage}[b]{0.49\linewidth}
% 			\centering	\includegraphics[width=1\linewidth]{figs/err_visul.png}
% 		\end{minipage}}
%   \vspace{-5mm}
%   \setcounter{subfigure}{0}
%   \subfigure[\label{fig:cartogram}]{
%     \centering
% 		\begin{minipage}[b]{0.5\linewidth}
% 			\centering	\includegraphics[width=1\linewidth]{}
% 		\end{minipage}}
%   \subfigure[\label{fig:errordistribution} ]{
%   \centering
%         \begin{minipage}[b]{0.45\linewidth}
% 			\centering	\includegraphics[width=0.65\linewidth]{figs/Figure_2.png}
% 		\end{minipage}}
%   \end{center}
%   \caption{(a)Histogram of equivalent error statistics of different image sizes for different satellite images. (b)The distribution of the equivalent error on the image and the relationship between the equivalent error and the image size are visualized.}
% \end{figure*}

%Fig.\ref{fig:cartogram} illustrates the comparable error effects of various satellite images more clearly and succinctly through histograms. Additionally, 

%\subsubsection{Influence of image size on reconstruction accuracy}

\begin{figure}[!htbp]
\centering
\subfigure[]{
\begin{minipage}[b]{0.115\textwidth}
\includegraphics[width=1\linewidth]{figure/WHU/5120/1/source_after_align.jpg}\vspace{4pt}
\includegraphics[width=1\linewidth]{figure/WHU/5120/1/error_map.jpg}\vspace{4pt}
\includegraphics[width=1\linewidth]{figure/WHU/5120/2/source_after_align.jpg}\vspace{4pt}
\includegraphics[width=1\linewidth]{figure/WHU/5120/2/error_map.jpg}\vspace{4pt}
\end{minipage}} \hspace{-3mm}
\subfigure[]{
\begin{minipage}[b]{0.115\textwidth}
\includegraphics[width=1\linewidth]{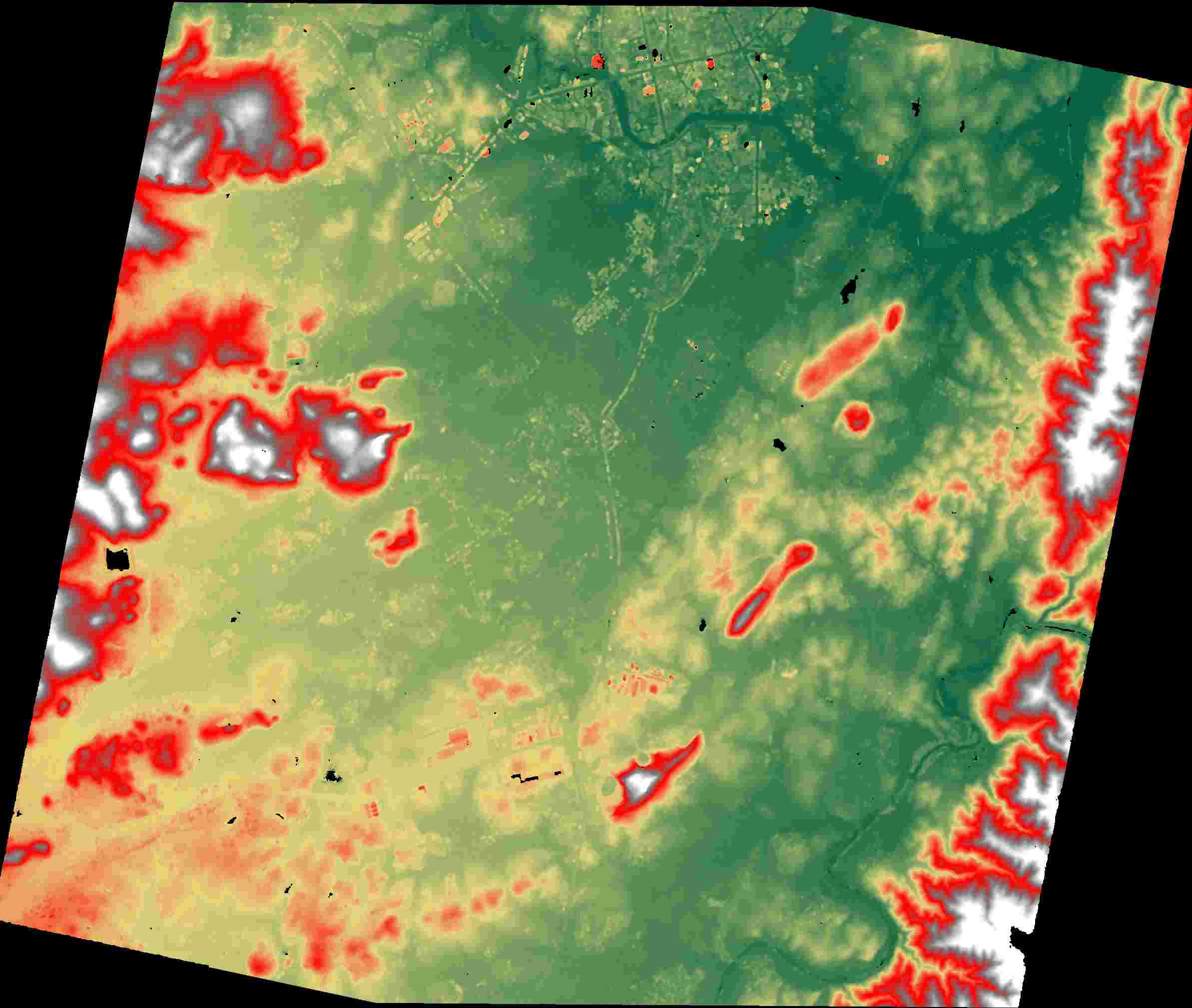}\vspace{4pt}
\includegraphics[width=1\linewidth]{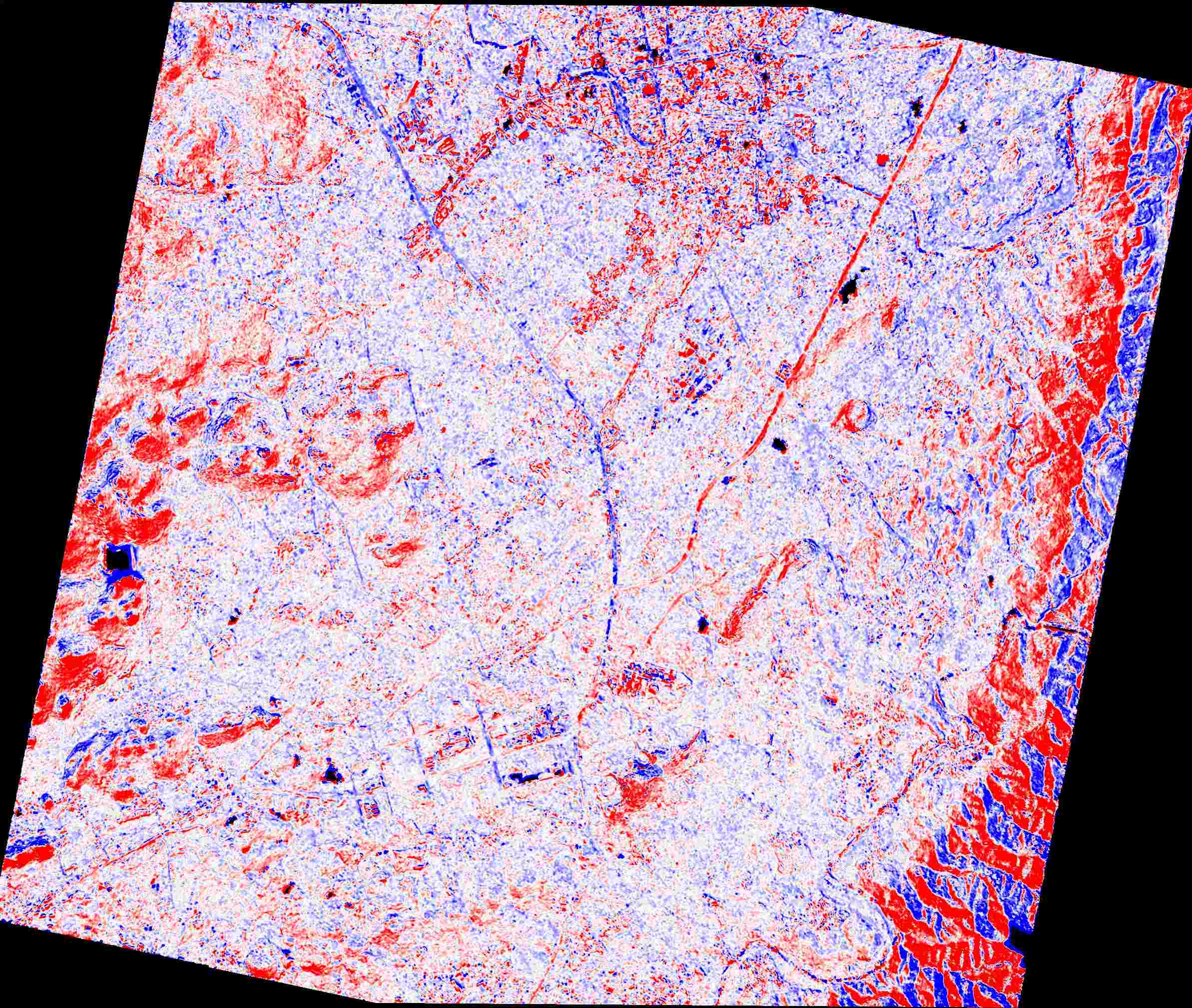}\vspace{4pt}
\includegraphics[width=1\linewidth]{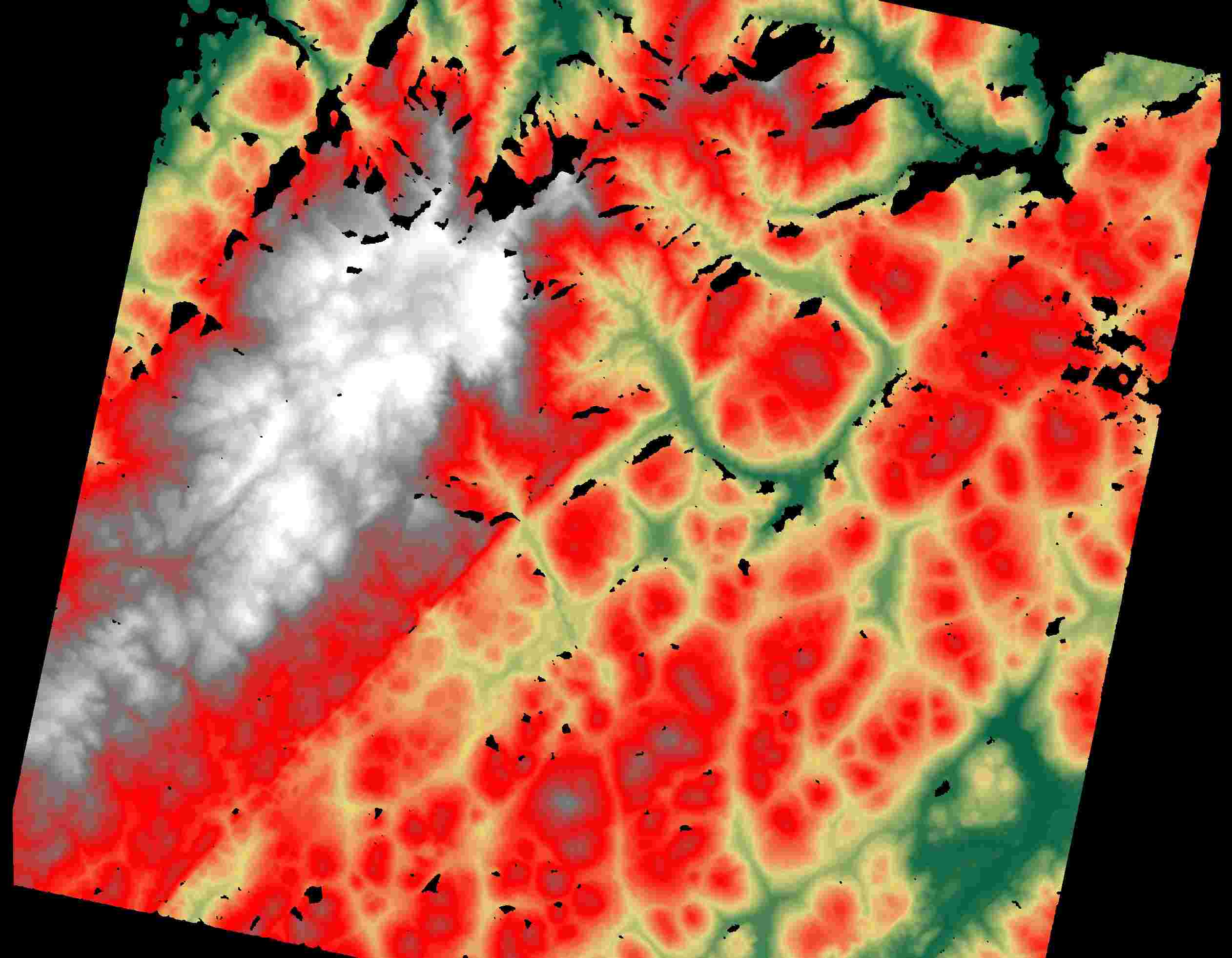}\vspace{4pt}
\includegraphics[width=1\linewidth]{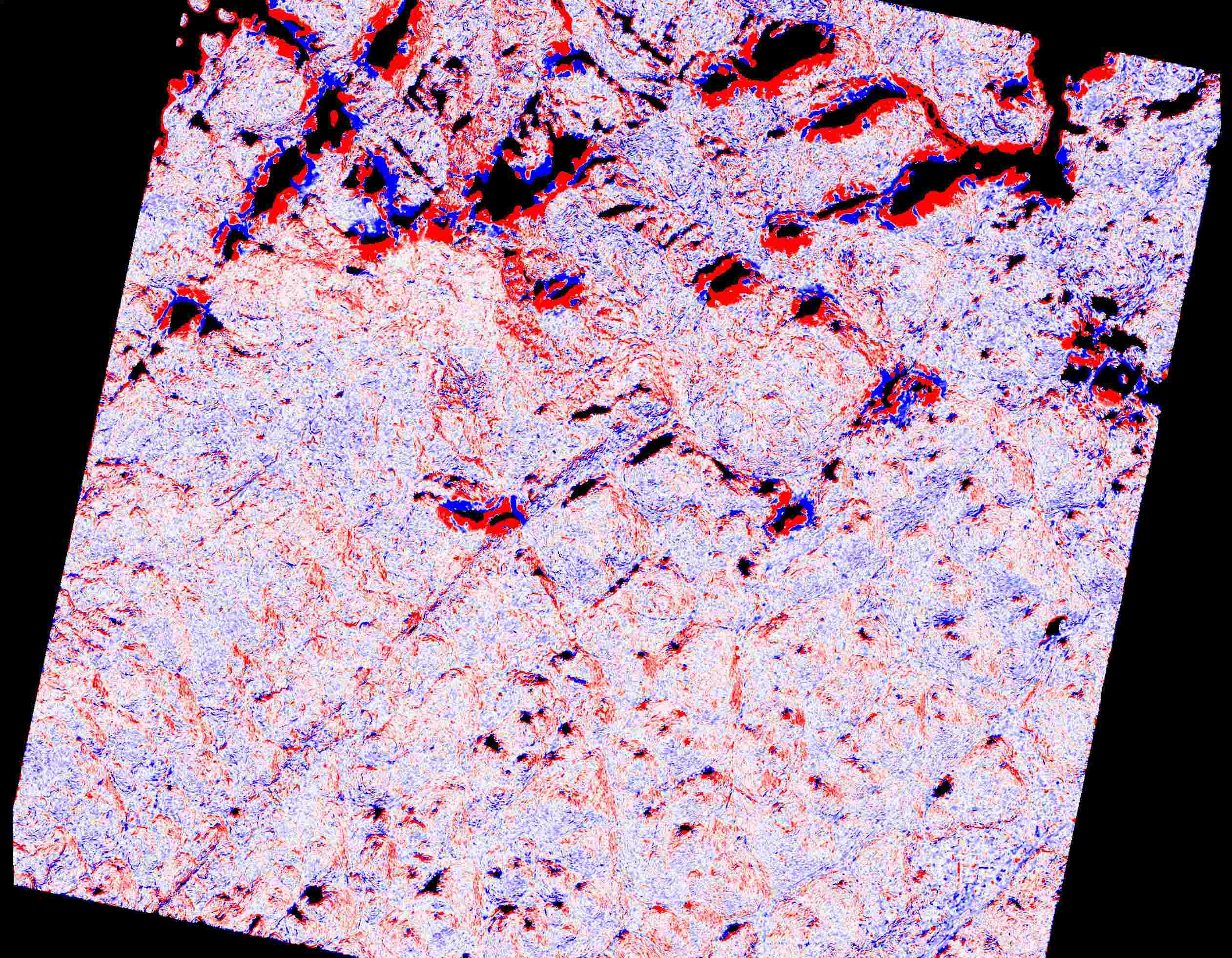}\vspace{4pt}
\end{minipage}}\hspace{-2mm}
\subfigure[]{
\begin{minipage}[b]{0.115\textwidth}
\includegraphics[width=1\linewidth]{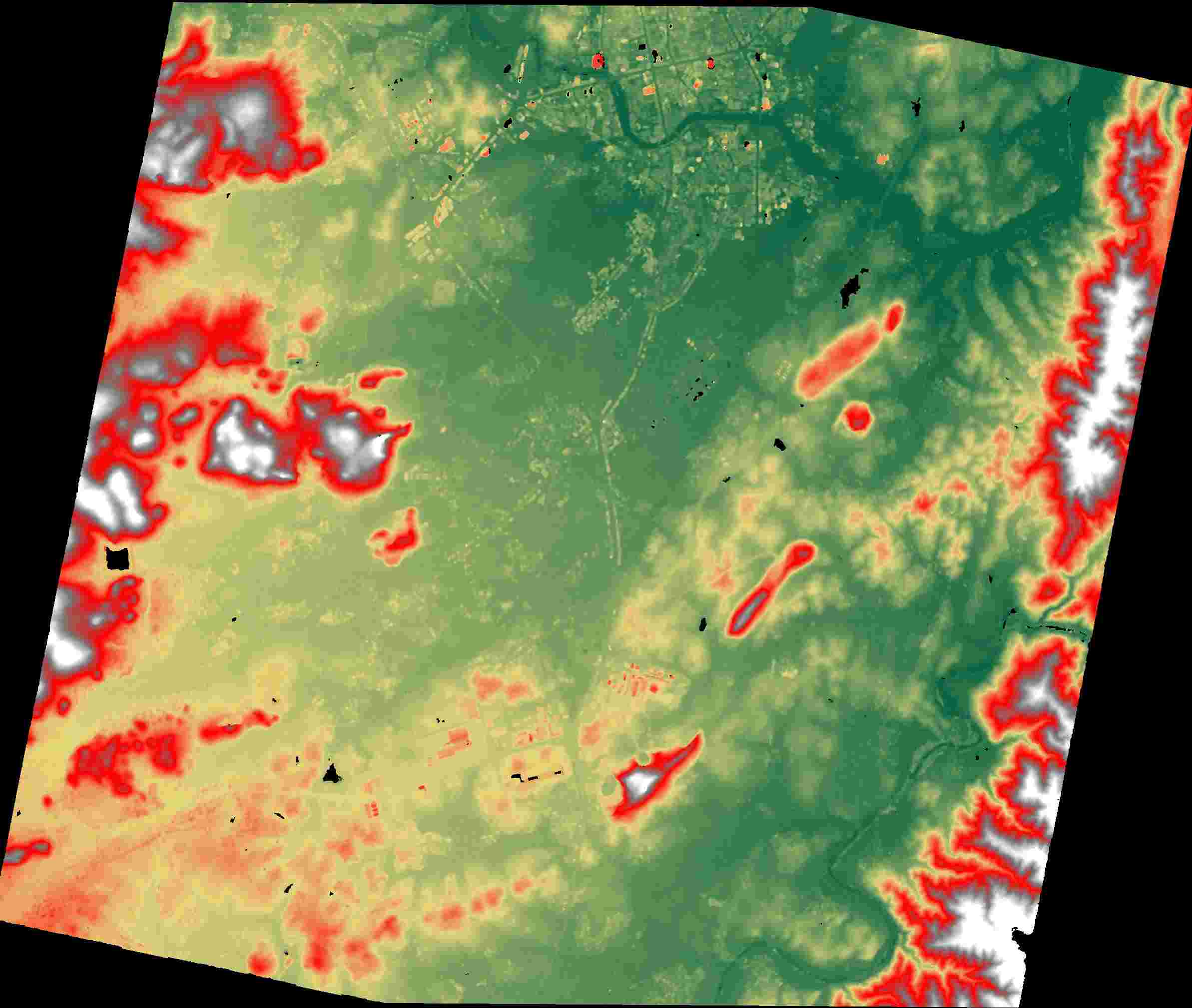}\vspace{4pt}
\includegraphics[width=1\linewidth]{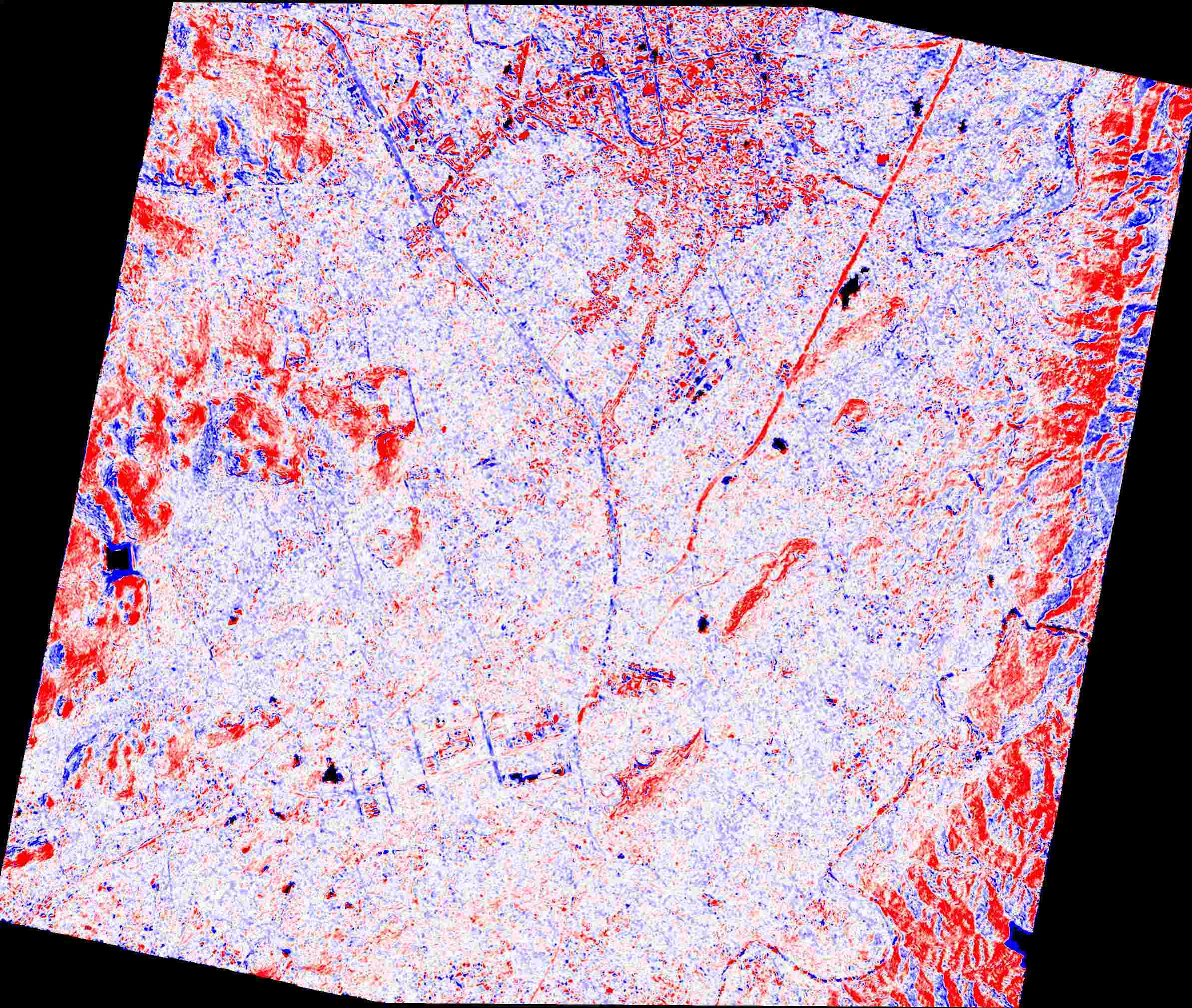}\vspace{4pt}
\includegraphics[width=1\linewidth]{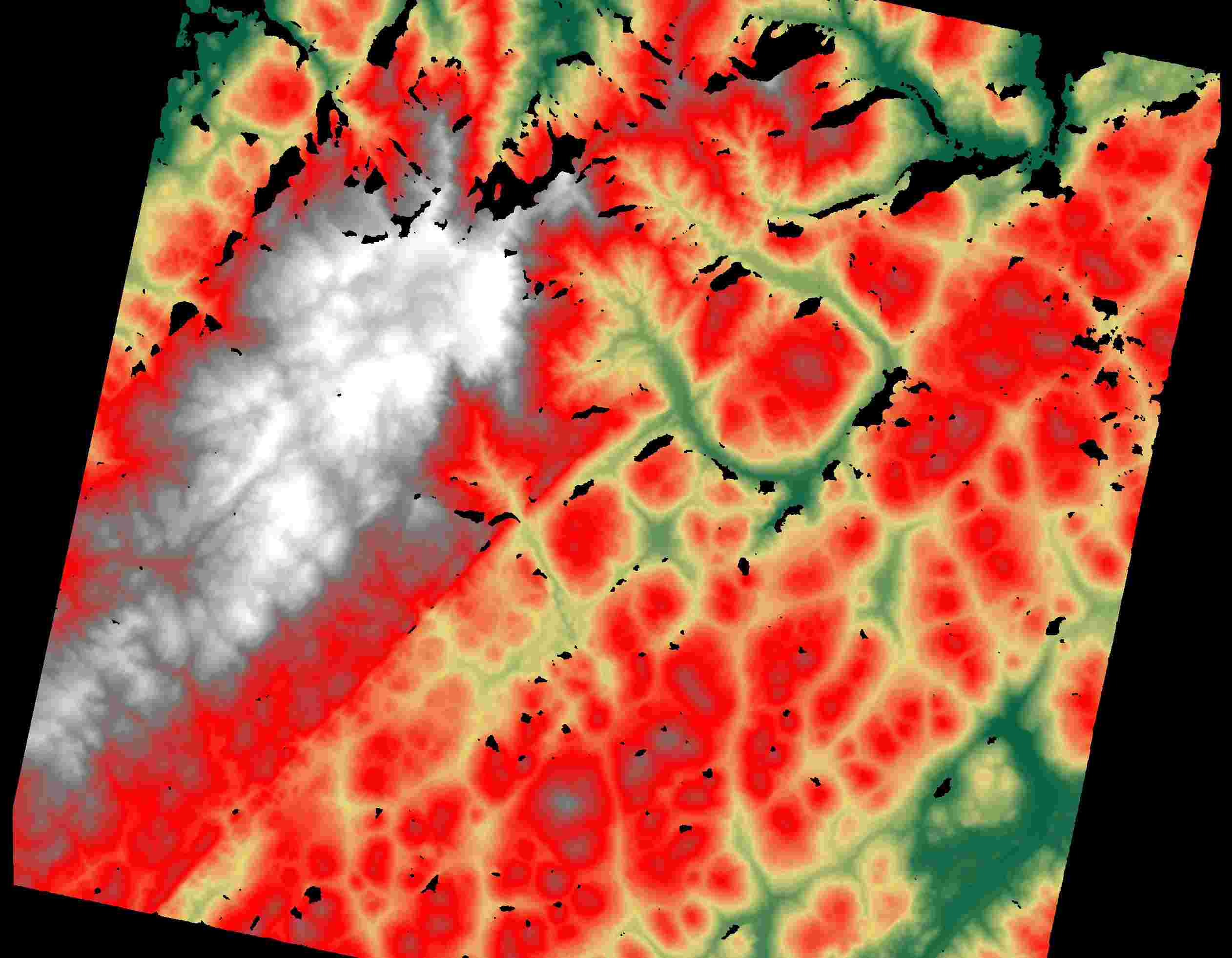}\vspace{4pt}
\includegraphics[width=1\linewidth]{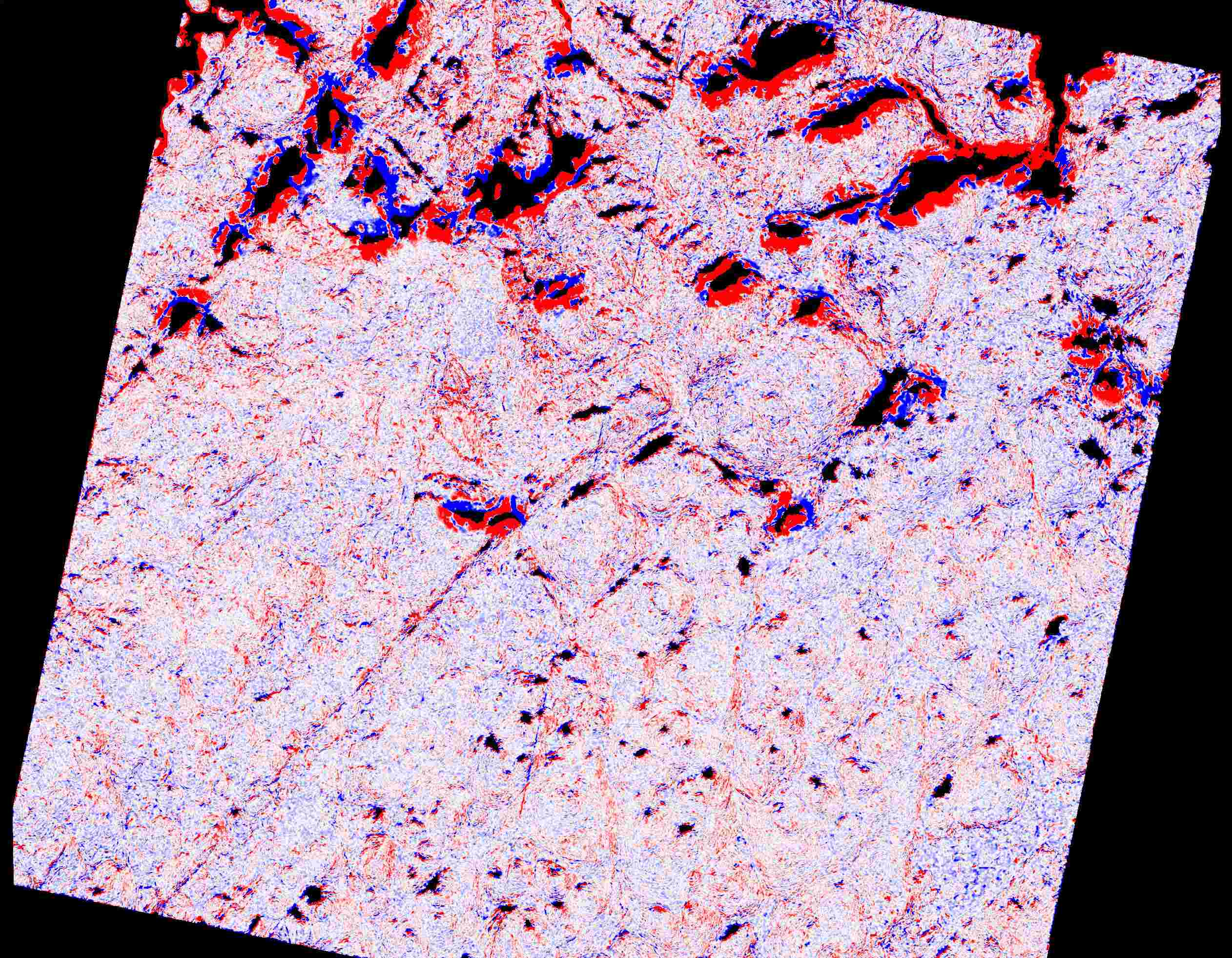}\vspace{4pt}
\end{minipage}} \hspace{-3mm}
\subfigure[]{
\begin{minipage}[b]{0.115\textwidth}
\includegraphics[width=1\linewidth]{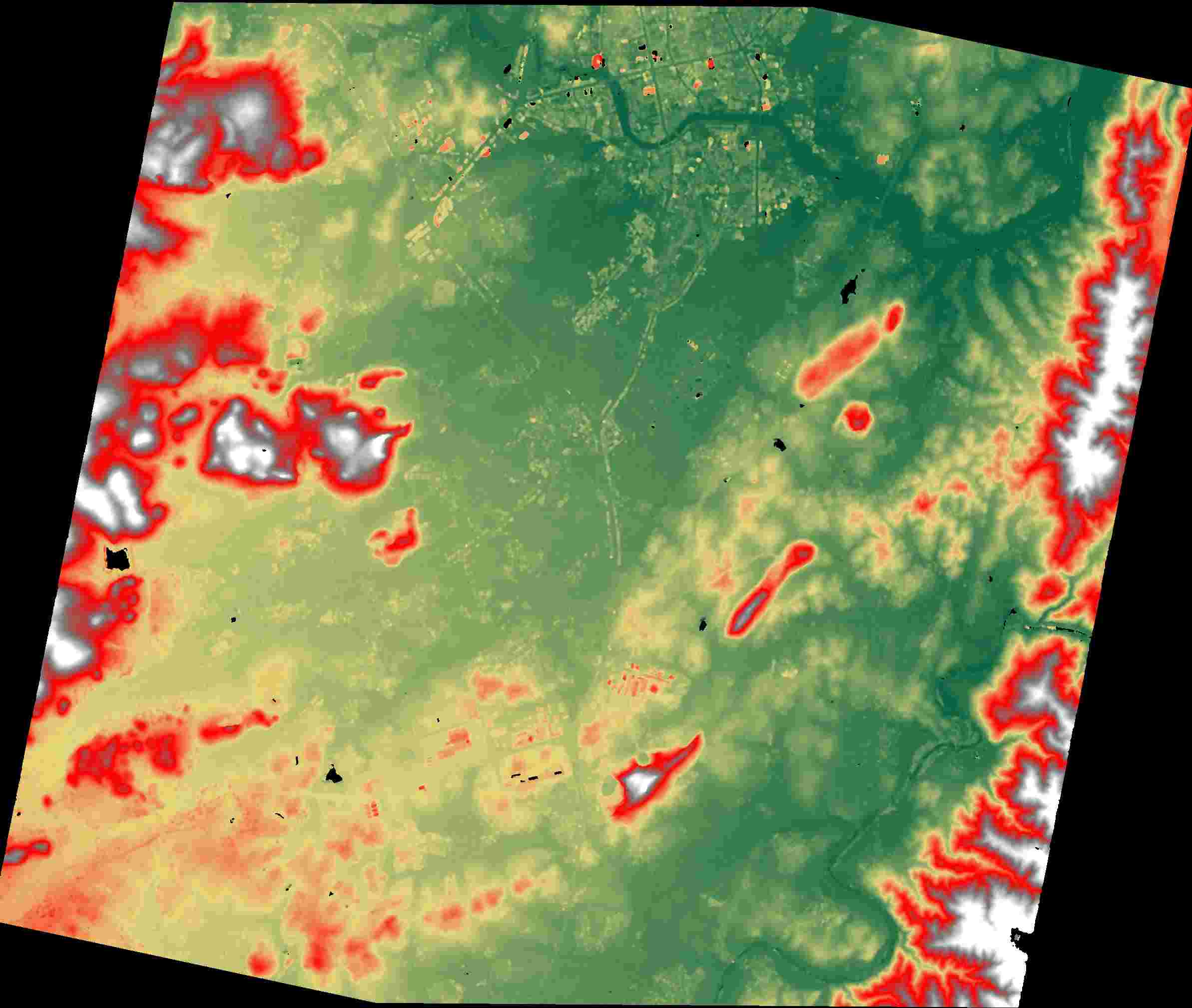}\vspace{4pt}
\includegraphics[width=1\linewidth]{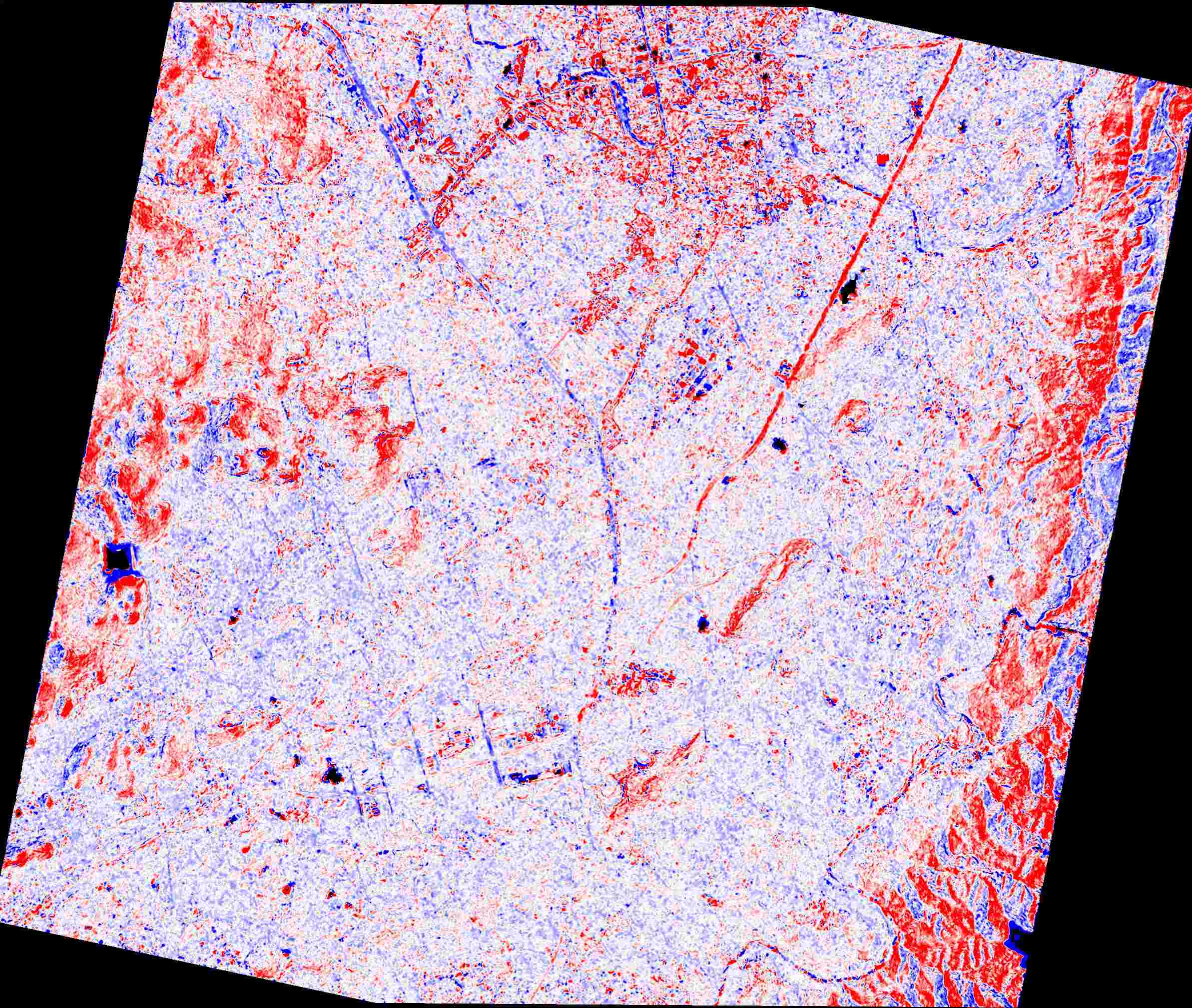}\vspace{4pt}
\includegraphics[width=1\linewidth]{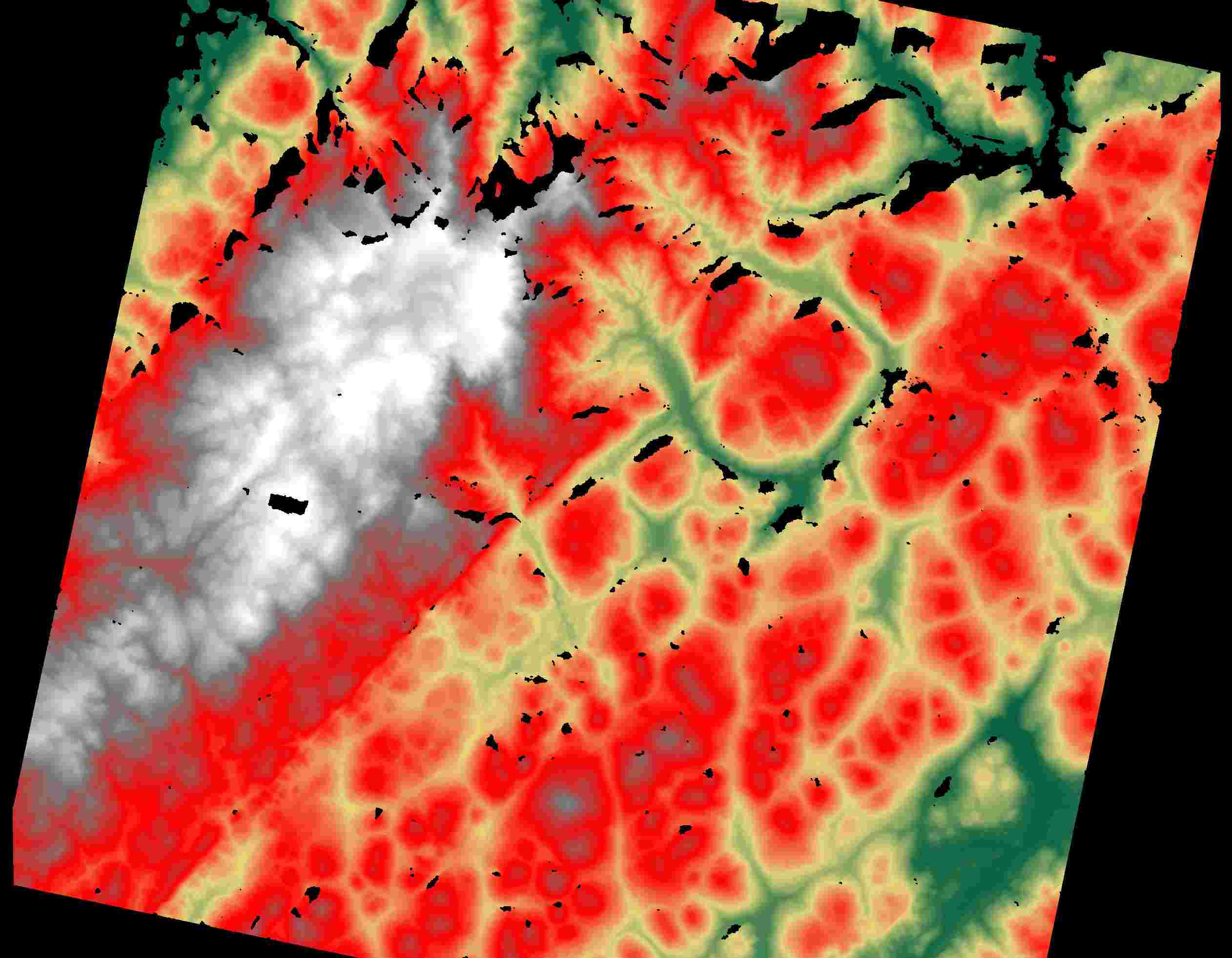}\vspace{4pt}
\includegraphics[width=1\linewidth]{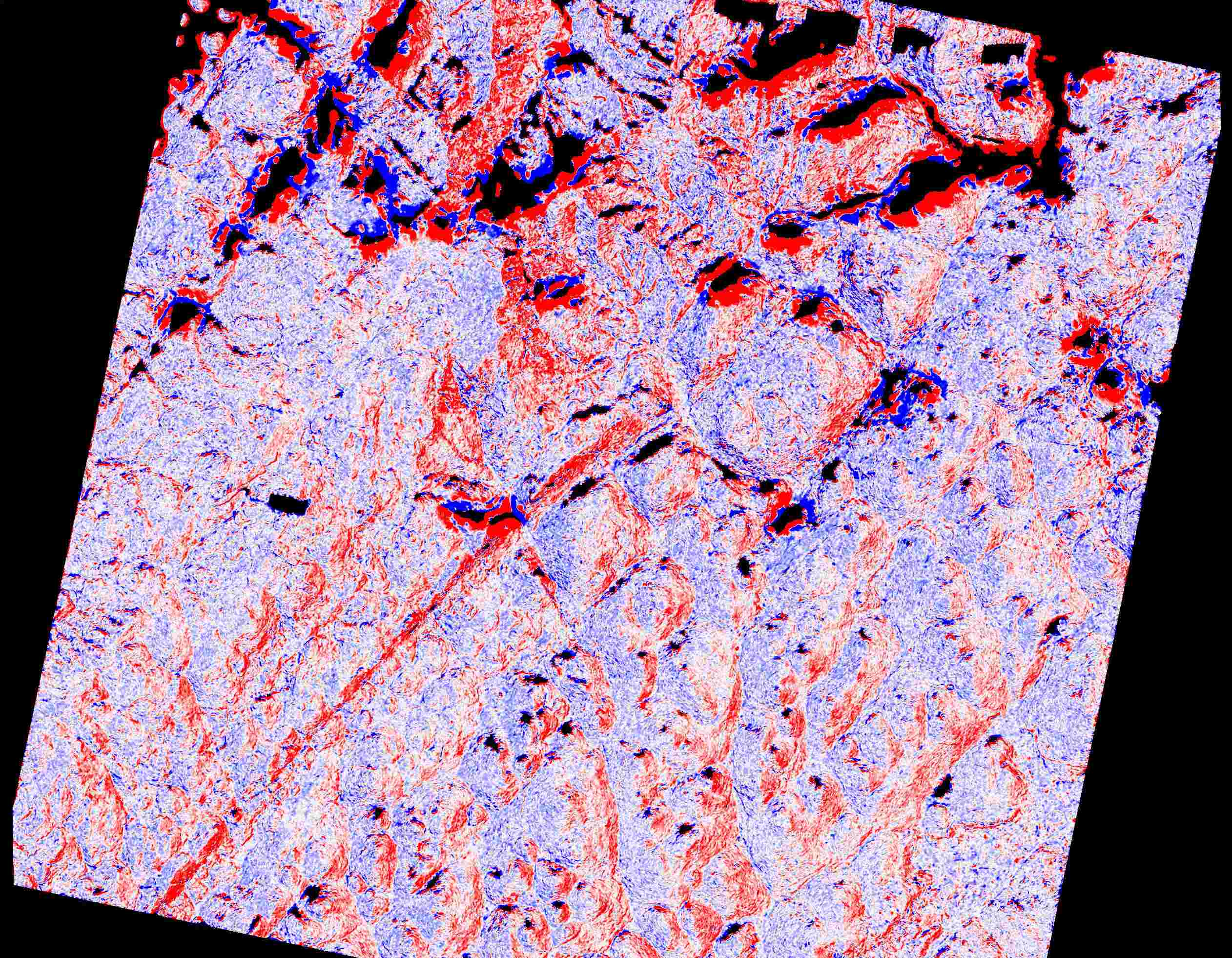}\vspace{4pt}
\end{minipage}}

\subfigure{
\begin{minipage}[b]{0.15\textwidth}
\includegraphics[width=1\linewidth]{figure/WHU/5120/1/source_after_align.cbar.jpg}\vspace{4pt}
\end{minipage}}
\subfigure{
\begin{minipage}[b]{0.15\textwidth}
\includegraphics[width=1\linewidth]{figure/WHU/5120/2/source_after_align.cbar.jpg}\vspace{4pt}
\end{minipage}}
\subfigure{
\begin{minipage}[b]{0.15\textwidth}
\includegraphics[width=1\linewidth]{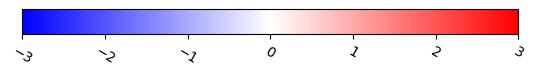}\vspace{4pt}
\end{minipage}}

\caption{DSM results for different image sizes in the WHU-TLC test set: (a) REPM@5120, (b) REPM@2048, (c) REPM@1024, and (d) REPM@512. The color
bar is expressed in units of meters.}
\label{fig:WHU_size}
\end{figure}

Next, we examine the impact of the image size on the reconstructed DSMs in the WHU-TLC dataset, as presented in Table \ref{tab:WHU}. The reduction in image size improved the ME accuracy and DSM completeness, but decreased the RMSE accuracy. Nevertheless, based on the results shown in Fig. \ref{fig:WHU_size}, the altitude error in the error map display decreased as the image size decreased. In the second group of images, which had substantial differences in altitude, the reconstruction precision declined significantly when the image size was reduced from 1024 pixel to 512 pixel. This underscores the importance of selecting an optimal image size. We conclude that reducing the image size leads to an exponential increase in the number of images required, which creates a more complex stereo-matching view-selection problem in the pipeline reconstruction process, resulting in a decrease in the reconstruction accuracy.

\begin{table}[htbp]
\footnotesize
  \caption{\label{tab:WHU} Influence of image size on reconstruction accuracy for same region on WHU-TLC test set.}
  \centering
  \begin{tabularx}{0.5\textwidth}{XXXXX}
    \toprule
    \textbf{Crop Size}	&\textbf{\makecell{ME\\(m)\textcolor{red}{$\downarrow$}}}	&\textbf{\makecell{RMSE\\(m)\textcolor{red}{$\downarrow$}}}	&\textbf{\makecell{Comp$_{2.5}$\\(\%)\textcolor{red}{$\uparrow$}}}	&\textbf{\makecell{Comp$_{5}$\\(\%)	\textcolor{red}{$\uparrow$}}}	\\
    \midrule
    $512$	&\textbf{0.919}	&4.291	&\textbf{79.864}  &\textbf{91.374}	  \\
    $1024$  &0.925	&3.940	&79.547	&91.204	 \\
    $2048$  &0.991	&3.843	&78.827	&90.979	\\
    $5120$  & 1.146	&\textbf{3.334}	&73.926	&88.244 \\	
    \bottomrule
  \end{tabularx}
\end{table}

\subsection{Influence of the polynomial image refinement model}
\label{sec:correctionerror}
%\subsubsection{Error experiment for the correction model} \label{sec:correctionerror}

We validated the effectiveness of the polynomial image refinement model on large scale ISPRS-ZY3 satellite images and GF7 satellite images. We compared the homography correction (H-Cor.) and polynomial correction (P-Cor.) for the equivalent error in images of different sizes. According to Table \ref{tab:re_err}, the polynomial correction was more effective than the homography correction, and the effect of the correction was more significant for large-scale satellite images.

\begin{table*}[htbp]
\footnotesize
  \caption{\label{tab:re_err} Equivalent errors of different scale images after homography correction and polynomial correction. All units in the table are pixels (pix.). The two corrections in the table, H-Cor and P-Cor, which resulted in a decrease in error and the largest decrease are shown in bold.}
  \centering
  \begin{tabular}{cllcccc}
    \toprule
    \textbf{Image Data}& \textbf{Crop Size}& \textbf{Methods}&\textbf{Samp RMSE\textcolor{red}{$\downarrow$}}&\textbf{Line RMSE\textcolor{red}{$\downarrow$}}& \textbf{RMSE\textcolor{red}{$\downarrow$}}& \textbf{Max Error\textcolor{red}{$\downarrow$} }         	\\
    \midrule
    \multirow{12}*{ISPRS-ZY3} &\multirow{3}*{Full size}
    & Uncorrected &2.457 &2.462& 3.479	&15.402 \\~
    & & H-Cor. &2.505 &2.483 &3.528 & 16.855 \\~
    & & P-Cor. &\textbf{1.912}& \textbf{1.897}& \textbf{2.693}	&\textbf{8.156} \\ 
    
    & \multirow{3}*{10000}
    & Uncorrected	&1.302 &1.314 &1.849	&7.622 \\~
    & & H-Cor. &1.315&1.303&1.851	  &7.738  \\~
    & & P-Cor.	&1.166 & 1.166& 1.649	&4.871    \\ 
    
    & \multirow{3}*{5120}
    & Uncorrected	&0.616 &0.617 &0.872	&3.112	       \\~
    & & H-Cor. &0.617&0.619	&0.874  &3.271  \\~
    & & P-Cor.	&0.597& 0.596 &0.844	&2.450    \\ 

    \midrule
    \multirow{9}*{GF7} &\multirow{3}*{Full size}
    & Uncorrected	&1.545 &1.328& 2.038	&8.959 \\~
    & & H-Cor. &1.526  &1.365&2.047& 9.104  \\~
    & & P-Cor.	&\textbf{1.346}	&\textbf{1.206}& \textbf{1.807} &\textbf{5.339}   \\ 
    
    &\multirow{3}*{10000}
    & Uncorrected	&0.342&0.341	&0.483&  1.609    \\~
    & & H-Cor. &0.342&0.341&0.483  & 1.617 \\~
    & & P-Cor.	 &0.338& 0.338&0.478	&1.385    \\ 
    
    & \multirow{3}*{5120}
    & Uncorrected	&0.173 &0.174 &0.246	&0.721	       \\~
    & & H-Cor. &0.174&0.174	&0.246  &0.725  \\~
    & & P-Cor.	&0.173&0.173&0.245	&0.689    \\ 
    \bottomrule
  \end{tabular}
\end{table*}

%& \multirow{3}*{2048}
% & Uncorrected	&0.240&0.241&0.340	&1.038	       \\~
% & & H-Cor. &0.241 &0.241 &0.341  &  1.081 \\~
% & & P-Cor. &0.238 &0.239 &0.337	&0.996    \\ 

In summary, the polynomial image refinement model was found to be more effective based on the metrics of the equivalent error. Subsequently, we evaluated the impact of the polynomial image refinement model on the DSM reconstruction accuracy. The ISPRS-ZY3 data indicated the absolute positioning accuracy calculated by the GCPs, whereas the GF7 data highlighted the accuracy and completeness of the DSM reconstruction. Based on our experimental findings, the addition of a polynomial image refinement model ensured that the reconstruction accuracy did not deteriorate substantially. However, for large images, the reconstruction accuracy and completeness were significantly enhanced.

\subsection{Future and outlook}

The 3D reconstruction method of RFM is equivalent to PCM provides new ideas and possibilities for the 3D reconstruction of linear-array satellite images. In this study, the proposed REPM pipeline exhibited excellent potential. 

However, weakly textured regions (such as bodies of water) remain a significant challenge for the reconstruction method. For instance, the reconstruction results of high-resolution images were more precise in contour. In contrast, the roofs of buildings were prone to larger holes, as shown in Fig. \ref{fig:DFC_benchmark}. Although high-resolution images are favorable for including more details in the reconstruction, this causes certain patches in the images to lack texture, which causes the matching computation and depth estimation to fail in weakly textured regions. Low-resolution images reflect the structural information contained in images more effectively \cite{ACMM}. Consequently, it may be worthwhile to introduce a multi-scale reconstruction framework that enhances the reconstruction effect in weakly textured regions. Furthermore, deep learning could be utilized to address the challenge of reconstructing regions with weak textural features and to enhance the precision and comprehensiveness of DSM reconstruction.

Although we can process VHR images with sizes up to $5120 \times 5120$ pixel, satellite images are generally large. If they are cropped (especially considering the overlap), the number of images, the memory they would occupy, and the time they would take to process would all increase exponentially. The number and size of the images determine the efficiency of the pipeline for a given input. Therefore, it is imperative to enhance the pipeline framework and increase the number of parallel operations to improve the operational efficiency of the pipeline.

In addition, because we determined the optimal image processing size based on the equivalent error, in the future we aim to create an adaptable model for choosing the most suitable image size.

\section{Conclusion}

In this study, we proposed an REPM pipeline for large-scale satellite CCD imagery. The pipeline is equipped with an equivalent pinhole model that converts the RFM to the PCM and a polynomial image refinement model that back-calculates the polynomial correction function to remap the image based on the least squares method. The experimental results showed that the REPM pipeline outperformed the state-of-the-art pipelines on four satellite image datasets. In addition, we assessed the effectiveness of the polynomial image refinement model in enhancing the precision of the DSM reconstruction for large-scale images. This model can be used for 3D reconstruction of large-scale CCD imagery. However, the model has many limitations, such as weak texture region processing methods and low efficiency in handling large-scale satellite images. In the future, we intend to further improve our pipeline by significantly enhancing the stereo matching of low-texture regions and increasing the reconstruction efficiency.

\section*{Acknowledgment}

This study was supported by the National Natural Science Foundation of China (grant numbers: 41971427, 42371459, 42101458, 42130112, and 42201513).

\bibliography{mybibfile}

\end{document}